\documentclass{article}

\usepackage{amsmath,amsfonts,bm}

\def\eqref#1{equation~\ref{#1}}

\def\1{\bm{1}}

\DeclareMathAlphabet{\mathsfit}{\encodingdefault}{\sfdefault}{m}{sl}
\SetMathAlphabet{\mathsfit}{bold}{\encodingdefault}{\sfdefault}{bx}{n}

\usepackage{neurips_2020,times}

\usepackage{amsmath,accents}
\usepackage{esvect}
\usepackage[utf8]{inputenc} 
\usepackage[T1]{fontenc}    
\usepackage[pagebackref=true,breaklinks=true,letterpaper=true,colorlinks,bookmarks=false,citecolor=blue,linkcolor=red]{hyperref}     
\usepackage{url}            
\usepackage{booktabs}       
\usepackage{amsfonts}       
\usepackage{nicefrac}    
\usepackage{amsmath} 
\usepackage{microtype}      
\usepackage[ruled,vlined,noend]{algorithm2e}
\usepackage{multirow}
\usepackage{float}
\usepackage{caption}
\usepackage{graphicx,subcaption}
\usepackage{xcolor}
\usepackage{wrapfig}

\usepackage[normalem]{ulem}

\usepackage{commath}
\title{Conditional Generative Modeling via Learning the Latent Space}

\author{Sameera Ramasinghe\\ 
Australian National University \& CSIRO, Data61 \\
\texttt{sameera.ramasinghe@anu.edu.au} \\
\And
Kanchana Ranasinghe\\ 
University of Moratuwa\\
\texttt{kahnchana@gmail.com} \\
\And
Salman Khan\\ 
Mohamed Bin Zayed University of Artificial Intelligence \\
\texttt{salman.khan@mbzuai.ac.ae} \\
\And
Nick Barnes\\ 
Australian National University \\
\texttt{nick.barnes@anu.edu.au} \\
\AND
Stephen Gould\\ 
Australian National University \\
\texttt{stephen.gould@anu.edu.au} \\
}

\usepackage[compact]{titlesec}
\titlespacing{\section}{0pt}{2ex}{1ex}
\titlespacing{\subsection}{0pt}{1ex}{0ex}
\titlespacing{\subsubsection}{0pt}{0.5ex}{0ex}

\begin{document}

\maketitle

\begin{abstract}
	
Although deep learning has achieved appealing results on several machine learning tasks, most of the models are deterministic at inference, limiting their application to single-modal settings. We propose a novel general-purpose framework for conditional generation in multimodal spaces, that uses latent variables to model generalizable learning patterns while minimizing a family of regression cost functions. At inference, the latent variables are optimized to find optimal solutions corresponding to multiple output modes.  Compared to existing generative solutions,  
our approach demonstrates faster and stable convergence, and can learn better representations for downstream tasks. Importantly, it provides a simple generic model that can beat highly engineered pipelines tailored using domain expertise on a variety of tasks, while generating diverse outputs. Our codes will be released.

\end{abstract}
\section{Introduction}
\label{sec:introduction}
Conditional generative models provide a natural mechanism to jointly learn a data distribution and optimize predictions. In contrast, discriminative models improve predictions by modeling the label distribution. Learning to model the data distribution allows generating novel samples and is considered a preferred way to understand the real world. Existing conditional generative models have generally been explored in single-modal settings, where a one-to-one mapping between input and output domains exists \citep{nalisnick2019deep,fetaya2020understanding}. Here, we investigate continuous multimodal (CMM) spaces for generative modeling, where one-to-many mappings exist between input and output domains. This is critical since many real world situations are inherently multi-modal, e.g., humans can imagine several outcomes for a given occluded image. 
In a discrete setting, this problem becomes relatively easy to tackle using techniques such as maximum-likelihood-estimation, since the output can be predicted as a vector \citep{zhang2016colorful}, which is not possible in continuous domains. Consequently, generative modeling in CMM spaces remains a challenging task. 

To model CMM spaces, a prominent approach in the literature is to use a combination of reconstruction and adversarial losses \citep{isola2017image,zhang2016colorful, contextEncoder}. However, this entails key shortcomings. 1) The goals of adversarial and reconstruction losses are contradictory (Sec.~\ref{adv_dist_compromise}), hence model engineering and numerous regularizers are required to support convergence
\citep{harmonizing,mao2019mode}, thereby resulting in less-generic models tailored for specific applications
\citep{pennet,vitoria2020chromagan}. 2) The adversarial loss based models are notorious for difficult convergence due to the challenge of finding Nash equilibrium of a non-convex min-max game in high-dimensions \citep{barnett2018convergence,chu2020smoothness, kodali2017convergence}. 3) The convergence is heavily dependent on the architecture, hence such models show lack of scalability \citep{thanh2019improving,arora2017gans}.  4) The promise of assisting downstream tasks 
remains challenging, with a large gap in performance between the generative modelling approaches and their discriminative counterparts \citep{secretlyEBM, ssl_survey}. 

In this work, we propose a general-purpose framework for modeling CMM spaces using a set of domain-agnostic regression cost functions instead of the adversarial loss. This improves both the stability and eliminates the incompatibility between the adversarial and reconstruction losses, allowing more precise outputs while maintaining diversity. The underlying notion is to learn the `\emph{behaviour of the latent variables}' in minimizing these cost functions while converging to an optimum mode during the training phase, and mimicking the same at inference.
Despite being a novel direction, the proposed framework showcases promising attributes by: (a) achieving state-of-the-art results on a diverse set of tasks using a generic model, implying 
generalizability, (b) rapid convergence to optimal modes despite architectural changes, 
(c) learning useful features for downstream tasks, 
and (d) producing diverse outputs via traversal through multiple output modes at inference.

\section{Proposed Methodology}
\label{sec:methodology}

We define a family of cost functions   $\{E_{i,j} =  d(y^{g}_{i,j},\mathcal{G}(x_j,w))\} \in \xi$,  where $x_j \sim \chi$ is the input, $y^{g}_{i,j} \sim \Upsilon$ is the $i^{th}$ ground-truth mode for $x_j$, $\mathcal{G}$ is a generator function with weights $w$, and $d(\cdot, \cdot)$ is a distance function. Note that the number of cost functions $E_{(\cdot, j)}$ for a given $x_j$ can vary over $\chi$. Our aim here is to come up with a generator function $ \mathcal{G}(x_j,w)$, that can minimize each $E_{i,j}, \forall i$ as $\mathcal{G}(x_j,w) \rightarrow y^{g}_{i,j}$. However, since $\mathcal{G}$ is a deterministic function ($x$ and $w$ are both fixed at inference), it can only produce a single output. Therefore, we introduce a latent vector $z$ to the generator function, that can be used to converge  $\bar{y}_{i,j} = \mathcal{G}(x_j,w,z_{i,j})$ towards a $y^{g}_{(i,j)}$ at inference, and possibly, to multiple solutions. Formally, the family of cost functions now becomes: $ \{ \hat{E}_{i,j} = d(y^{g}_{i,j},\mathcal{G}(x_j,w,z_{i,j})) \}, \forall z_{i,j} \sim \zeta.$
Then, 
our training objective can be defined as finding a set of optimal $z_{i}^* \in \zeta$ and $w^* \in \omega$ by minimizing $\mathbb{E}_{i \sim I} [\hat{E}_{i}]$,  
where $I$ is the number of possible solutions for $x_j$. Note that $w^*$ is fixed for all $i$ and a different $z_{i}^*$ exists
for each $i$. Considering all the training samples $x_j \sim \chi$, our training objective becomes,
\begin{align}
\label{equ:trainobj}
    \{\{z_{i,j}^*\}, w^*\}= \underset{z_{i,j} \in \zeta , w \in \omega}{\arg\min}\, \mathbb{E}_{i \in I,j\in J}[\hat{E}_{i,j}].
\end{align}
Eq.~\ref{equ:trainobj} can be optimized via Algorithm \ref{alg:train} (proof in App. \ref{app:convergence}).
Intuitively, the goal of 
Eq. \ref{equ:trainobj} is to obtain a family of optimal latent codes \{$z^*_{i,j}\}$, each causing a global minima in the corresponding $\hat{E}_{i,j}$ as $y^{g}_{i,j} = \mathcal{G}(x_j,w, z^*_{i,j})$. Consequently, at inference, we can optimize $\bar{y}_{i,j}$ to converge to an optimal mode in the output space by varying $z$. Therefore, we predict an estimated $\bar{z}_{i,j}$ at inference,
\begin{equation}
\label{equ:testobj}
    \bar{z}_{i,j} \approx  \underset{z}{\min}\, \hat{E}_{i,j},
\end{equation}
for each $y^{g}_{i,j}$, which in turn can be used to obtain the prediction $\mathcal{G}(x_j,\bar{z}_{i,j},w) \approx y^{g}_{i,j}$.
In other words, for a selected $x_j$, let $\bar{y}^t_{i,j}$  be the initial estimate for $\bar{y}_{i,j}$. At inference, $z$ can traverse gradually towards an optimum point $y^g_{i,j}$ in the space, forcing $\bar{y}_{i,j}^{t+n} \rightarrow y^g_{i,j}$, in finite steps ($n$). 

However, still a critical problem exists: Eq. \ref{equ:testobj} depends on $y^g_{i,j}$, which is not available at inference. As a remedy, we enforce Lipschitz constraints on $\mathcal{G}$ over $(x_j,z_{i,j})$, which bounds the gradient norm as,
\begin{align}
\label{equ:bound}
{\scriptstyle \frac{\norm{\mathcal{G}(x_j,w^*,z^*_{i,j}) - \mathcal{G}(x_j,w^*,z_{0})}}{\norm{z^*_{i,j} - z_{0}}} }\leq \int \norm{\nabla_z \mathcal{G}(x_j,w^*,\gamma (t))}dt \leq C,
\end{align}
where $z_0 \sim \zeta$ is an arbitrary random initialization, $C$ is a constant, and $\gamma(\cdot)$ is a straight path from $z_0$ to $z^*_{i,j}$ (proof in App. \ref{app:gradnorm}) . Intuitively, Eq.~\ref{equ:bound} implies that the gradients $\nabla_z \mathcal{G}(x_j,w^*,z_{0}) $ along the path $\gamma(\cdot)$ do not tend to vanish or explode, hence, finding the  path to optimal $z^*_{i,j}$ in the space $\zeta$ becomes a fairly straight forward regression problem. 
Moreover, enforcing the Lipschitz constraint encourages meaningful structuring of the latent space: suppose $z^*_{1,j}$ and $z^*_{2,j}$ are two optimal codes corresponding to two ground truth modes for a particular input. Since $\| z^*_{2,j} - z^*_{1,j} \|$ is lower bounded by $\frac{\norm{\mathcal{G}(x_j,w^*,z^*_{2,j}) - \mathcal{G}(x_j,w^*,z^*_{1,j})}}{L}$, where $L$ is the Lipschitz constant, the minimum distance between the two latent codes is proportional to the difference between the corresponding ground truth modes. In practice, we observed that this encourages the optimum latent codes to be placed sparsely (visual illustration in App. \ref{app:propositions}), which helps a network to learn distinctive paths towards different modes.

\subsection{Convergence at inference}
\label{sec:convergence}
We formulate finding the convergence path of $z$ at inference as a regression problem, i.e., ${z}_{t+1} = r({z}_t,x_j)$.  
 We implement $r(\cdot)$ as a recurrent neural network (RNN). The series of predicted values $\{z_{(t+k)} : k=1,2,..,N\}$ can be modeled as a first-order Markov chain requiring no memory for the RNN. We observe that enforcing Lipschitz continuity on $\mathcal{G}$ over $z$ leads to smooth trajectories even in high dimensional settings, hence, memorizing more than one step
 in to the history is redundant. However, ${z}_{t+1}$ is not a state variable, i.e., the existence of multiple modes for output prediction $\bar{y}$ leads to multiple possible solutions for $z_{t+1}$. On the contrary, $\mathbb{E}[z_{t+1}]$ is a state variable w.r.t. the state $({z}_t,x)$, which can be used as an approximation to reach the optimal ${z}^*$ at inference. Therefore, instead of directly learning $r(\cdot)$, we learn a simplified version $r'(z_t,x) = \mathbb{E}[z_{t+1}]$.
 Intuitively, the whole process can be understood as observing the behavior of $z$ on
 a smooth surface at the training stage, and predicting the movement at inference. A key aspect of $r'(z_t,x)$ is that the model is capable of converging to multiple possible optimum modes at inference based on the initial position of $z$. 
 
 

\subsection{Momentum as a supplementary aid}
\label{sec:kinetic}
Based on Sec.~\ref{sec:convergence}, ${z}$ can now traverse to an optimal position ${z^*}$ during inference. However, there can exist rare symmetrical positions in the $\zeta$ 
where $\mathbb{E}[{z}_{t+1}] - {z}_t \approx 0$, although far away from $\{{z^*}\}$, forcing ${z}_{t+1} \approx {z}_t$. Simply, the above phenomenon can occur if  some ${z}_{t+1}$ has traveled in many non-orthogonal directions, so the vector addition of ${z}_{t+1} \approx 0$. This can \emph{fool} the system to falsely identify convergence points, forming \emph{phantom} optimum point distributions amongst the true distribution (see Fig.~\ref{fig:toy_example_plots}). To avoid such behavior, we consider $\vec{v}(z_t,x_j) = ({z}_{t+1} - {z_t})_{x_j}$. Then, we learn the expected momentum $\mathbb{E}[\rho(z_t,x_j)] = \alpha \mathbb{E}[|\vec{v}(z_t,x_j)|]$ at each $({z}_{t}, x_j)$ during the training phase, where $\alpha$ is an empirically chosen scalar. In practice, $\mathbb{E}[\rho(z_t,x_j)] \rightarrow 0$ as ${z}_{t+1},{z}_t \rightarrow \{{z}^* \}$. Thus, to avoid \emph{phantom} distributions, we improve the $z$ update as,
\begin{align}
\label{eq:momentum}
    z_{t+1} = z_{t} + \mathbb{E}[\rho(z_t,x_j)]  \left[ { \frac{r'(z_t,x_j) - {z}_t}{\norm {r'(z_t,x_j) - {z}_t }} } \right].
\end{align}
Since both $\mathbb{E}[\rho(z_t,x_j)]$ and $r'(z_t,x_j)$ are functions on $(z_t,x_j)$, we jointly learn these two functions using a single network $\mathcal{Z}(z_t,x_j)$. Note that coefficient $\mathbb{E}[\rho(z_t,x_j)]$  serves two practical purposes: 1) slows down the movement of $z$ near true distributions, 2) pushes $z$ out of the phantom distributions. 

\section{Overall Design}\label{sec:imp}
\begin{figure}
\centering
\captionsetup{size=small}
\begin{subfigure}{0.4\textwidth}
\includegraphics[width=\linewidth]{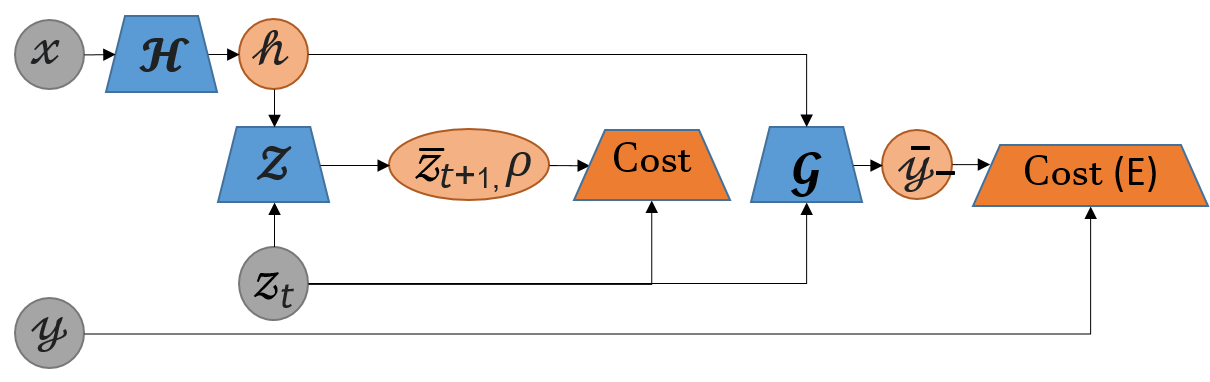}\vspace{-0.2em}
\caption{Training}
\end{subfigure}
\hspace{10pt}
\begin{subfigure}{0.4\textwidth}
\includegraphics[width=\linewidth]{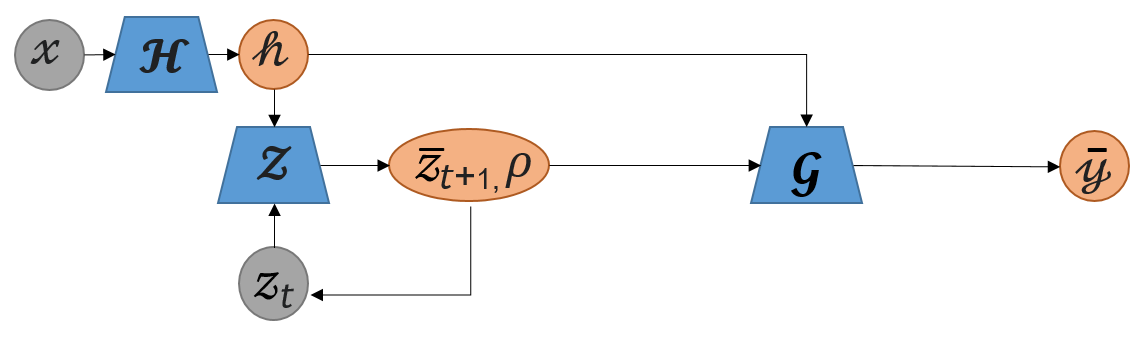}\vspace{-0.2em}
\caption{Inference}
\end{subfigure}
\vspace{0.2em}
\caption{Training and inference process. Refer to Algorithm \ref{alg:train} for the training process. At inference, $z$ is iteratively updated using the predictions of $\mathcal{Z}$ and fed to $\mathcal{G}$ to obtain increasingly fine-tuned outputs (see Sec.~\ref{sec:imp}).}  \label{fig:deisgn}
\vspace{0.5em}
\end{figure}
\begin{algorithm}[t]
\setlength{\textfloatsep}{0pt}
\SetAlgoLined
\small
    sample inputs $\{ x_1,x_2, ..., x_J \} \in \chi $; sample outputs $\{ y_1,y_2, ..., y_J \} \in \Upsilon $ ; \\
\For{$k$ epochs}{
\For{$x$ in $\chi$ }{
  \For{$l$ steps }{
     update $z = \{ z_1,z_2, ..., z_J \}$:
     $\nabla_{z}\hat{E} $ \hfill $\rhd$ Freeze $\mathcal{H},\mathcal{G},\mathcal{Z}$ and update $z$ \\
     update $\mathcal{Z}$: $\nabla_{w}L_1[(z_{t+1},\rho), \mathcal{Z}(z_t,\mathcal{H}(x))]  $ \hfill $\rhd$ Freeze $\mathcal{H},\mathcal{G},z$ and update $\mathcal{Z}$
  }
  update $\mathcal{G},\mathcal{H}$: $\nabla_{w}\hat{E}$ \hfill $\rhd$ Freeze $\mathcal{Z},z$ and update $\mathcal{H},\mathcal{G}$
 }
 }
\caption{\small Training algorithm }
\label{alg:train}
\end{algorithm}

The proposed model consists of three major blocks as shown in Fig.~\ref{fig:deisgn}: an encoder $\mathcal{H}$, a generator $\mathcal{G}$, and  $\mathcal{Z}$. Note that for derivations in Sec.~\ref{sec:methodology}, we used $x$ instead of $h = \mathcal{H}(x)$, as $h$ is a high-level representation of $x$. The training process is illustrated in Algorithm \ref{alg:train}. At each optimization $z_{t+1} = z_t - \beta \nabla_{z_t} [ \hat{E}_{i,j} ]$, $\mathcal{Z}$ is trained separately to approximate $(z_{t+1}, \rho)$. At inference, $x$ is fed to $\mathcal{H}$, and then the $\mathcal{Z}$ optimizes the output $\bar{y}$ by updating $z$ for a pre-defined number of iterations of Eq.~\ref{eq:momentum}. 
For $\hat{E}(\cdot, \cdot)$, we use $L_1$ loss. Furthermore, it is important to limit the search space for $z_{t+1}$, to improve the performance of $\mathcal{Z}$. To this end, we sample $z$ from the surface of the $n$-dimensional sphere ($\mathbb{S}^n$). Moreover, to ensure faster convergence of the model, we force the Lipschitz continuity on both $\mathcal{Z}$ and the $\mathcal{G}$ (App. \ref{app:lipschitz}) . For hyper-parameters and training details, see App. \ref{sec:architecture}. 



\section{Motivation}
\label{adv_dist_compromise}
Here, we explain the drawbacks of conditional GAN methods and illustrate our idea via a toy example. 

\noindent \textbf{Incompatibility of adversarial and reconstruction losses:} cGANs use a combination of adversarial and reconstruction losses. We note that this combination is suboptimal to model CMM spaces. \\
\textbf{\textit{Remark:}} \textit{Consider a generator ${G}(x,z)$ and a discriminator $D(x,z)$, 
where $x$ and $z$ are the input and the noise vector, respectively. Then, consider an arbitrary input $x_j$ and the corresponding set of ground-truths $\{y^g_{i,j}\}, i=1,2,..N$. Further, let us define the optimal generator $ G^*(x_j, z) = \hat{y}, \hat{y} \in \{y^g_{i,j}\}$, $L_{GAN} = \mathbb{E}_{i} [\log D(y^g_{i,j})] + \mathbb{E}_{z}[\log(1-D(G(x_j,z))]$ and  $L_{\ell} = \mathbb{E}_{i,z}[|y^g_{i,j} - G(x_j,z)|]$. Then, $G^* \neq \hat{G}^*$ where $\hat{G}^* = \arg\underaccent{G}{\min}\underaccent{D}{\max} L_{GAN} + \lambda L_{\ell}$,  $\forall \lambda \neq 0$.} (Proof in App. \ref{app:remark}).


\noindent \textbf{Generalizability:} The incompatibility of above mentioned loss functions demands domain specific design choices from models that target high realism in CMM settings. 
This hinders the generalizability  across different tasks \citep{vitoria2020chromagan, pennet}. 
We further argue that due to this discrepancy, cGANs learn sub-optimal features which are less useful for downstream tasks (Sec.~\ref{sec:completion}).

\noindent \textbf{Convergence and the sensitivity to the architecture:} The difficulty of converging GANs to the Nash equilibrium of a non-convex min-max game in high-dimensional spaces is well explored \citep{barnett2018convergence, chu2020smoothness, kodali2017convergence}. \citet{goodfellow2014generative} underlines \emph{if the discriminator has enough capacity, and is optimal at every step of the GAN algorithm, then the generated distribution converges to the real distribution}; that cannot be guaranteed in a practical scenario. In fact, \citet{arora2018gans} confirmed that the adversarial objective can easily approach to an equilibrium even if the generated distribution has very low support, and further, the number of training samples required to avoid mode collapse can be in order of $\exp(d)$ ($d$ is the data dimension).

\noindent \textbf{Multimodality:}
The ability to generate diverse outputs, i.e., convergence to multiple modes in the output space, is an important requirement. 
Despite the typical noise input, cGANs generally lack the ability to generate diverse outputs \citep{harmonizing}. \citet{contextEncoder} and \citet{iizuka2016let} even state that better results are obtained when the noise is completely removed. Further, variants of cGAN that target diversity often face a trafe-off between the realism and diversity \citep{he2018diverse}, as they have to compromise between the reconstruction and adversarial losses. 

\begin{figure}[b]
\captionsetup{size=small}
\centering 
\begin{minipage}[h]{.70\textwidth} \vspace{-1.2em}
\begin{subfigure}{0.015\textwidth}
\rotatebox[origin=c]{90}{{$\scriptstyle y_1 = 4x$}}
\end{subfigure}
\begin{subfigure}{0.19\textwidth}
\includegraphics[width=\linewidth]{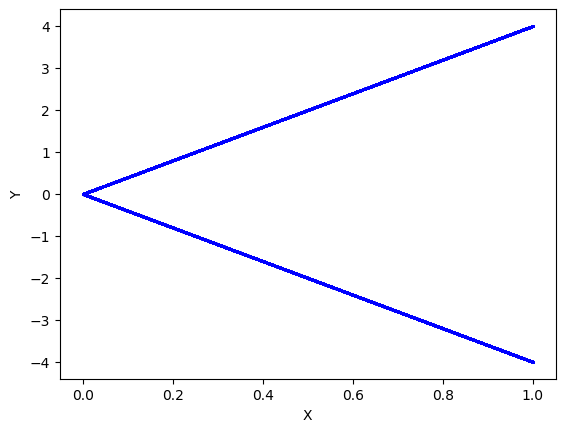} 
\end{subfigure} 
\begin{subfigure}{0.19\textwidth}
\includegraphics[width=\linewidth]{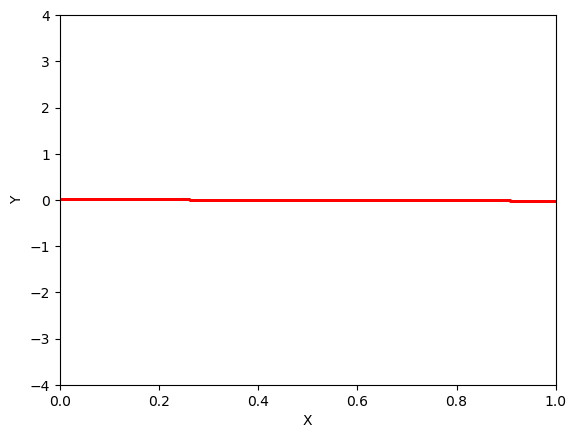} 
\end{subfigure} 
\begin{subfigure}{0.19\textwidth}
\includegraphics[width=\linewidth]{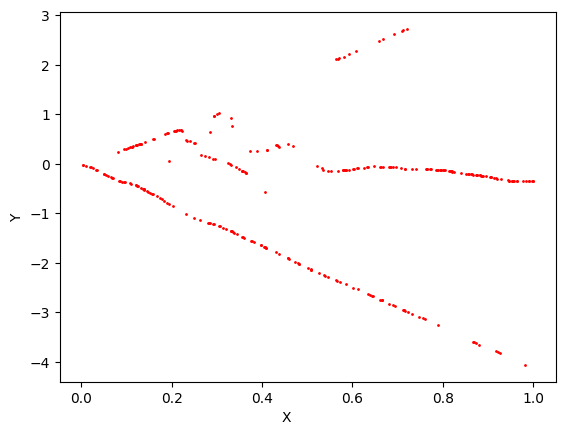} 
\end{subfigure} 
\begin{subfigure}{0.19\textwidth}
\includegraphics[width=\linewidth]{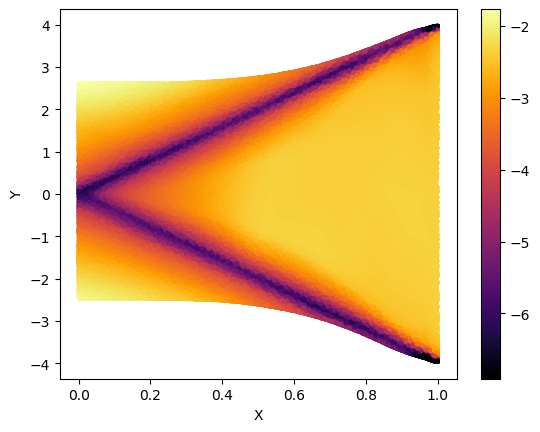}
\end{subfigure}
\begin{subfigure}{0.19\textwidth}
\includegraphics[width=\linewidth]{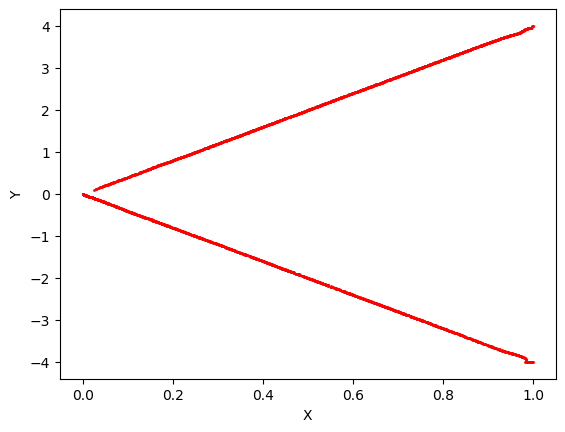} 
\end{subfigure} 

\begin{subfigure}{0.015\textwidth}\hspace{-2mm}
\rotatebox[origin=c]{90}{{$\scriptstyle y_2 = 4x^2$}}
\end{subfigure}
\begin{subfigure}{0.19\textwidth}
\includegraphics[width=\linewidth]{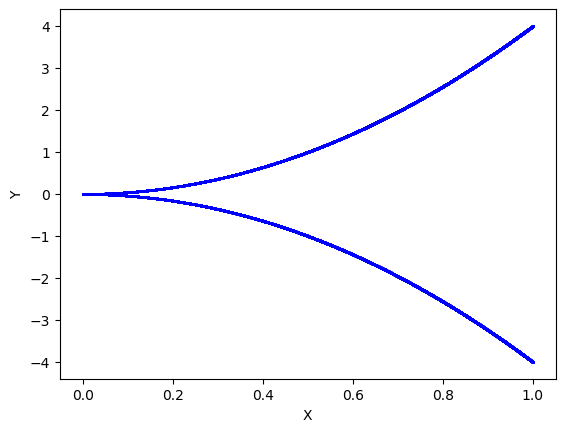}
\end{subfigure}
\begin{subfigure}{0.19\textwidth}
\includegraphics[width=\linewidth]{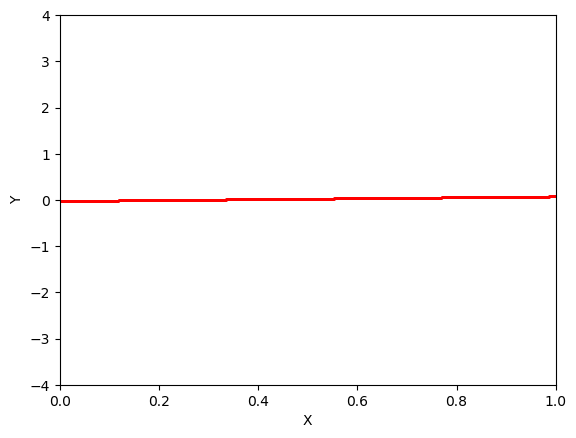}
\end{subfigure}
\begin{subfigure}{0.19\textwidth}
\includegraphics[width=\linewidth]{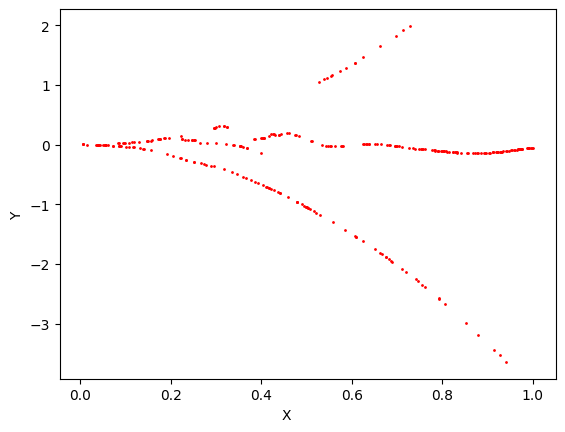}
\end{subfigure}
\begin{subfigure}{0.19\textwidth}
\includegraphics[width=\linewidth]{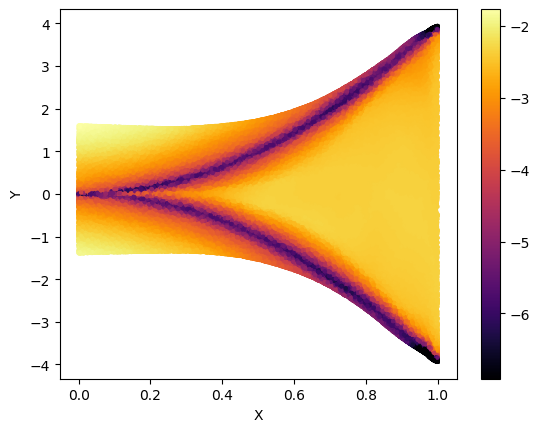}
\end{subfigure}
\begin{subfigure}{0.19\textwidth}
\includegraphics[width=\linewidth]{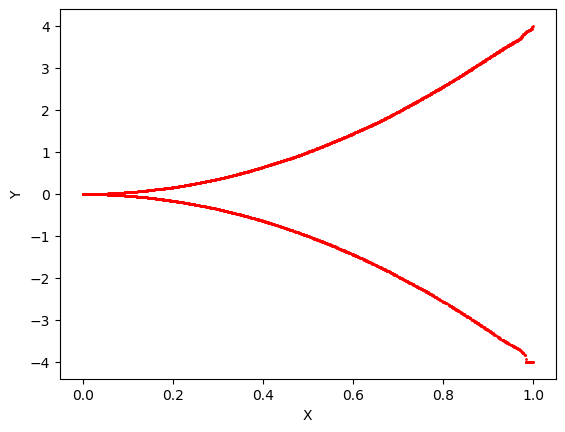}
\end{subfigure}

\begin{subfigure}{0.015\textwidth}\hspace{-2mm}
\rotatebox[origin=c]{90}{{$\scriptstyle y_3 = 4x^3$}}
\end{subfigure}
\begin{subfigure}{0.19\textwidth}
\includegraphics[width=\linewidth]{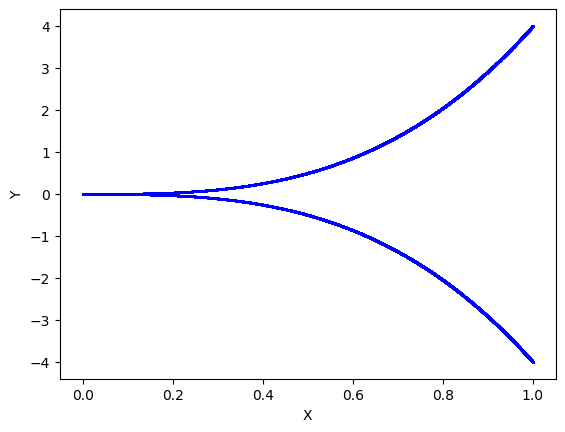}\vspace{-0.5em}
\caption{\tiny{GT}}
\end{subfigure}
\begin{subfigure}{0.19\textwidth}
\includegraphics[width=\linewidth]{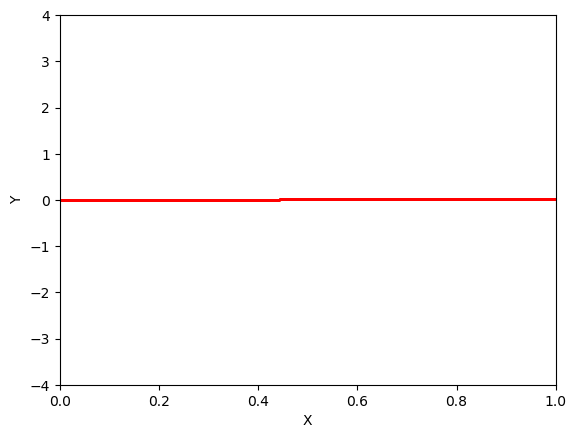}\vspace{-0.5em}
\caption{\tiny{$L_1$}}
\end{subfigure}
\begin{subfigure}{0.19\textwidth}
\includegraphics[width=\linewidth]{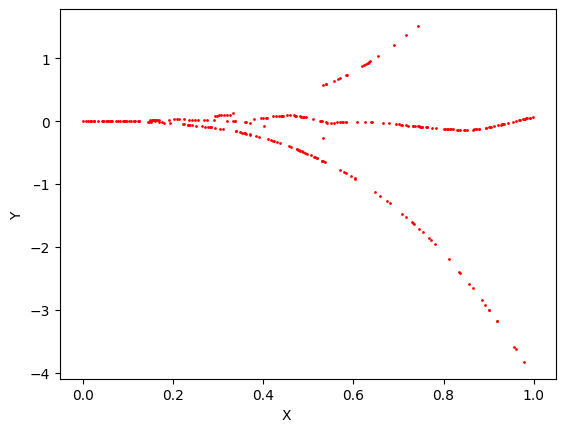}\vspace{-0.5em}
\caption{\tiny{Ours w/o $\rho$}}
\end{subfigure}
\begin{subfigure}{0.19\textwidth}
\includegraphics[width=\linewidth]{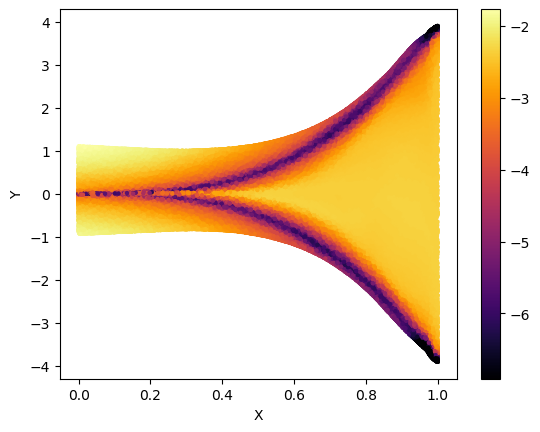}\vspace{-0.5em}
\caption{\tiny{$\mathbb{E}[\rho]$ heatmap}}
\end{subfigure}
\begin{subfigure}{0.19\textwidth}
\includegraphics[width=\linewidth]{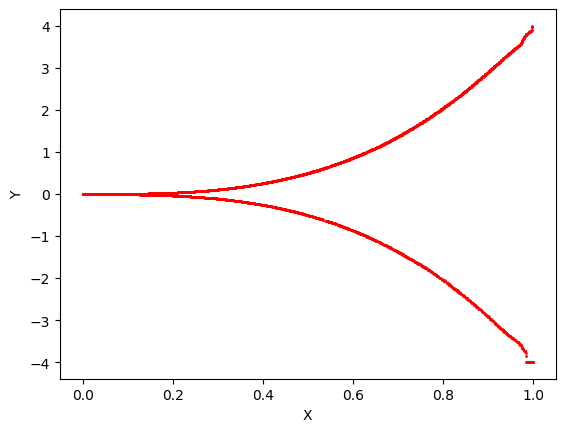}\vspace{-0.5em}
\caption{\tiny{Ours}}
\end{subfigure}
\end{minipage}\hfill
\begin{minipage}[h]{.29\textwidth}
\captionof{figure}{\emph{Toy Example:} Plots generated for each dimension of the CMM space $\Upsilon$. (a) Ground-truth distributions. (b) Model outputs for $L_1$ loss. (c) Output when trained with the proposed objective (without $\rho$ correction). Note the \emph{phantom distribution} identified by the model. (d) $\mathbb{E}[\rho]$ as a heatmap on $(x,y)$. $\mathbb{E}[\rho]$ is lower near the true distribution and higher otherwise. (e) Model outputs after $\rho$ correction.}
\label{fig:toy_example_plots}
\end{minipage}
\end{figure}

\textbf{A toy example:}
Here, we experiment with the formulations in Sec.~\ref{sec:methodology}. Consider a 3D CMM space $y = \pm 4(x, x^2, x^3)$.
Then, we construct three layer multi-layer perceptrons (MLP) to represent each of the functions, $\mathcal{H}$, $\mathcal{G}$, and $\mathcal{Z}$, and compare the proposed method against the $L_1$ loss. Figure \ref{fig:toy_example_plots} illustrates the results. As expected, $L_1$ loss generates the line $y=0$, and is inadequate to model the multimodal space.  As explained in Sec.~\ref{sec:kinetic}, without
momentum correction, the network is fooled by a phantom distribution where $\mathbb{E}[{z}_{t+1}] \approx 0$ at training time. However, the \emph{push} of 
momentum removes the phantom distribution and refines the output to closely resemble the input distribution. As implied in Sec.~\ref{sec:kinetic}, the momentum is maximized near the true distribution and minimized otherwise.

\section{Experiments and discussions}
The distribution of natural images lies on a high dimensional manifold, making the task of modelling it extremely challenging. Moreover, conditional image generation poses an additional challenge with their constrained multimodal output space (a single input may correspond to multiple outputs while not all of them are available for training). In this section, we experiment on several such tasks. For a fair comparison with a similar capacity GAN, 
we use the encoder and decoder architectures used in \citet{contextEncoder} for $\mathcal{H}$ and $\mathcal{G}$ respectively. We make two minor modifications: the channel-wise fully connected (FC) layers are removed and U-Net style skip connections are added (see App. \ref{sec:architecture}). 
We train the existing models for a maximum of $200$ epochs where pretrained weights are not provided, and demonstrate the generalizability of our theoretical framework in diverse practical settings by using a generic network for all the experiments. Models used for comparisons are denoted as follows: PN \citep{pennet}, CA \citep{yu2018generative}, DSGAN \citep{yang2019diversity}, CIC~\citep{zhang2016colorful}, Chroma~\citep{vitoria2020chromagan}, P2P \citep{isola2017image}, Izuka \citep{iizuka2016let}, CE \citep{contextEncoder}, CRN \citep{chen2017photographic}, and B-GAN \citep{zhu2017toward}.


\begin{figure}[b]
	\vspace{-5mm}
\captionsetup{size=small}
\begin{minipage}[h]{.53\textwidth}
\centering

\scalebox{0.95}{
\begin{subfigure}{0.015\textwidth}
\rotatebox[origin=c]{90}{{\tiny GT}}\hspace{-2mm}
\end{subfigure}
\begin{subfigure}{0.08\textwidth}
\includegraphics[width=\linewidth]{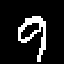} 
\end{subfigure} 
\begin{subfigure}{0.08\textwidth}
\includegraphics[width=\linewidth]{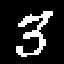}
\end{subfigure}
\begin{subfigure}{0.08\textwidth}
\includegraphics[width=\linewidth]{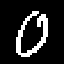}
\end{subfigure}
\begin{subfigure}{0.08\textwidth}
\includegraphics[width=\linewidth]{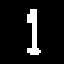}
\end{subfigure}
\begin{subfigure}{0.08\textwidth}
\includegraphics[width=\linewidth]{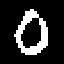}
\end{subfigure}
\begin{subfigure}{0.08\textwidth}
\includegraphics[width=\linewidth]{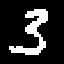}
\end{subfigure}
\begin{subfigure}{0.08\textwidth}
\includegraphics[width=\linewidth]{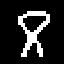}
\end{subfigure}
\begin{subfigure}{0.08\textwidth}
\includegraphics[width=\linewidth]{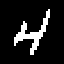}
\end{subfigure}
\begin{subfigure}{0.08\textwidth}
\includegraphics[width=\linewidth]{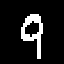}
\end{subfigure}
\begin{subfigure}{0.08\textwidth}
\includegraphics[width=\linewidth]{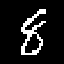}
\end{subfigure}
\begin{subfigure}{0.08\textwidth}
\includegraphics[width=\linewidth]{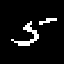}
\end{subfigure}
}

\scalebox{0.95}{
\begin{subfigure}{0.015\textwidth}
\rotatebox[origin=c]{90}{{\tiny Input}}\hspace{-2mm}
\end{subfigure}
\begin{subfigure}{0.08\textwidth}
\includegraphics[width=\linewidth]{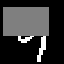} 
\end{subfigure} 
\begin{subfigure}{0.08\textwidth}
\includegraphics[width=\linewidth]{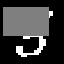}
\end{subfigure}
\begin{subfigure}{0.08\textwidth}
\includegraphics[width=\linewidth]{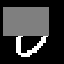}
\end{subfigure}
\begin{subfigure}{0.08\textwidth}
\includegraphics[width=\linewidth]{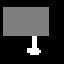}
\end{subfigure}
\begin{subfigure}{0.08\textwidth}
\includegraphics[width=\linewidth]{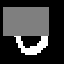}
\end{subfigure}
\begin{subfigure}{0.08\textwidth}
\includegraphics[width=\linewidth]{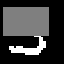}
\end{subfigure}
\begin{subfigure}{0.08\textwidth}
\includegraphics[width=\linewidth]{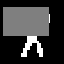}
\end{subfigure}
\begin{subfigure}{0.08\textwidth}
\includegraphics[width=\linewidth]{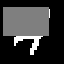}
\end{subfigure}
\begin{subfigure}{0.08\textwidth}
\includegraphics[width=\linewidth]{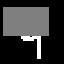}
\end{subfigure}
\begin{subfigure}{0.08\textwidth}
\includegraphics[width=\linewidth]{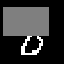}
\end{subfigure}
\begin{subfigure}{0.08\textwidth}
\includegraphics[width=\linewidth]{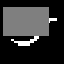}
\end{subfigure}
}

\scalebox{0.95}{
\begin{subfigure}{0.015\textwidth}
\rotatebox[origin=c]{90}{{\tiny $L_1$}}\hspace{-2mm}
\end{subfigure}
\begin{subfigure}{0.08\textwidth}
\includegraphics[width=\linewidth]{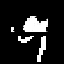} 
\end{subfigure} 
\begin{subfigure}{0.08\textwidth}
\includegraphics[width=\linewidth]{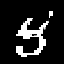}
\end{subfigure}
\begin{subfigure}{0.08\textwidth}
\includegraphics[width=\linewidth]{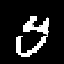}
\end{subfigure}
\begin{subfigure}{0.08\textwidth}
\includegraphics[width=\linewidth]{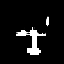}
\end{subfigure}
\begin{subfigure}{0.08\textwidth}
\includegraphics[width=\linewidth]{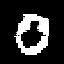}
\end{subfigure}
\begin{subfigure}{0.08\textwidth}
\includegraphics[width=\linewidth]{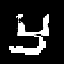}
\end{subfigure}
\begin{subfigure}{0.08\textwidth}
\includegraphics[width=\linewidth]{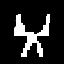}
\end{subfigure}
\begin{subfigure}{0.08\textwidth}
\includegraphics[width=\linewidth]{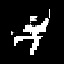}
\end{subfigure}
\begin{subfigure}{0.08\textwidth}
\includegraphics[width=\linewidth]{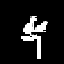}
\end{subfigure}
\begin{subfigure}{0.08\textwidth}
\includegraphics[width=\linewidth]{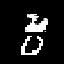}
\end{subfigure}
\begin{subfigure}{0.08\textwidth}
\includegraphics[width=\linewidth]{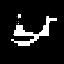}
\end{subfigure}
}

\scalebox{0.95}{
\begin{subfigure}{0.015\textwidth}
\rotatebox[origin=c]{90}{{\tiny CE}}\hspace{-2mm}
\end{subfigure}
\begin{subfigure}{0.08\textwidth}
\includegraphics[width=\linewidth]{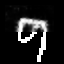} 
\end{subfigure} 
\begin{subfigure}{0.08\textwidth}
\includegraphics[width=\linewidth]{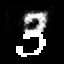}
\end{subfigure}
\begin{subfigure}{0.08\textwidth}
\includegraphics[width=\linewidth]{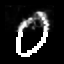}
\end{subfigure}
\begin{subfigure}{0.08\textwidth}
\includegraphics[width=\linewidth]{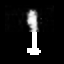}
\end{subfigure}
\begin{subfigure}{0.08\textwidth}
\includegraphics[width=\linewidth]{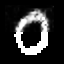}
\end{subfigure}
\begin{subfigure}{0.08\textwidth}
\includegraphics[width=\linewidth]{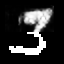}
\end{subfigure}
\begin{subfigure}{0.08\textwidth}
\includegraphics[width=\linewidth]{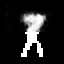}
\end{subfigure}
\begin{subfigure}{0.08\textwidth}
\includegraphics[width=\linewidth]{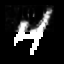}
\end{subfigure}
\begin{subfigure}{0.08\textwidth}
\includegraphics[width=\linewidth]{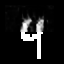}
\end{subfigure}
\begin{subfigure}{0.08\textwidth}
\includegraphics[width=\linewidth]{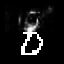}
\end{subfigure}
\begin{subfigure}{0.08\textwidth}
\includegraphics[width=\linewidth]{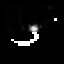}
\end{subfigure}
}

\scalebox{0.95}{
\begin{subfigure}{0.015\textwidth}
\rotatebox[origin=c]{90}{{\tiny Ours}}\hspace{-2mm}
\end{subfigure}
\begin{subfigure}{0.08\textwidth}
\includegraphics[width=\linewidth]{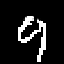} 
\end{subfigure} 
\begin{subfigure}{0.08\textwidth}
\includegraphics[width=\linewidth]{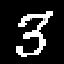}
\end{subfigure}
\begin{subfigure}{0.08\textwidth}
\includegraphics[width=\linewidth]{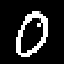}
\end{subfigure}
\begin{subfigure}{0.08\textwidth}
\includegraphics[width=\linewidth]{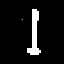}
\end{subfigure}
\begin{subfigure}{0.08\textwidth}
\includegraphics[width=\linewidth]{figures/mnist2/predz_9.png}
\end{subfigure}
\begin{subfigure}{0.08\textwidth}
\includegraphics[width=\linewidth]{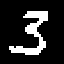}
\end{subfigure}
\begin{subfigure}{0.08\textwidth}
\includegraphics[width=\linewidth]{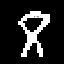}
\end{subfigure}
\begin{subfigure}{0.08\textwidth}
\includegraphics[width=\linewidth]{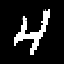}
\end{subfigure}
\begin{subfigure}{0.08\textwidth}
\includegraphics[width=\linewidth]{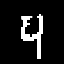}
\end{subfigure}
\begin{subfigure}{0.08\textwidth}
\includegraphics[width=\linewidth]{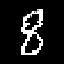}
\end{subfigure}
\begin{subfigure}{0.08\textwidth}
\includegraphics[width=\linewidth]{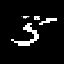}
\end{subfigure}
}\vspace{-0.5em}
\captionof{figure}{Performance with 20\% corrupted data. Our model demonstrates better convergence compared to  $L_1$ loss and a similar capacity GAN \citep{contextEncoder}.} \label{fig:mnist_example_01}
\end{minipage}
\hfill
\begin{minipage}[h]{.20\textwidth}
\captionsetup{size=small}
\scalebox{0.8}{
\begin{subfigure}{0.21\textwidth}
\includegraphics[width=\linewidth]{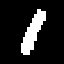} 
\end{subfigure} 
\begin{subfigure}{0.21\textwidth}
\includegraphics[width=\linewidth]{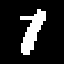}
\end{subfigure}
\begin{subfigure}{0.21\textwidth}
\includegraphics[width=\linewidth]{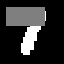}
\end{subfigure}
\begin{subfigure}{0.21\textwidth}
\includegraphics[width=\linewidth]{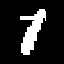}
\end{subfigure}
\begin{subfigure}{0.21\textwidth}
\includegraphics[width=\linewidth]{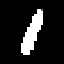}
\end{subfigure}
}

\scalebox{0.8}{
\begin{subfigure}{0.21\textwidth}
\includegraphics[width=\linewidth]{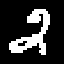} 
\end{subfigure} 
\begin{subfigure}{0.21\textwidth}
\includegraphics[width=\linewidth]{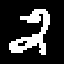}
\end{subfigure}
\begin{subfigure}{0.21\textwidth}
\includegraphics[width=\linewidth]{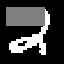}
\end{subfigure}
\begin{subfigure}{0.21\textwidth}
\includegraphics[width=\linewidth]{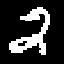}
\end{subfigure}
\begin{subfigure}{0.21\textwidth}
\includegraphics[width=\linewidth]{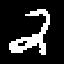}
\end{subfigure}
}

\scalebox{0.8}{
\begin{subfigure}{0.21\textwidth}
\includegraphics[width=\linewidth]{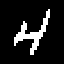} 
\end{subfigure} 
\begin{subfigure}{0.21\textwidth}
\includegraphics[width=\linewidth]{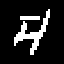}
\end{subfigure}
\begin{subfigure}{0.21\textwidth}
\includegraphics[width=\linewidth]{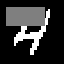}
\end{subfigure}
\begin{subfigure}{0.21\textwidth}
\includegraphics[width=\linewidth]{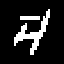}
\end{subfigure}
\begin{subfigure}{0.21\textwidth}
\includegraphics[width=\linewidth]{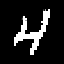}
\end{subfigure}
}

\scalebox{0.8}{
\begin{subfigure}{0.21\textwidth}
\includegraphics[width=\linewidth]{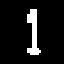} 
\end{subfigure} 
\begin{subfigure}{0.21\textwidth}
\includegraphics[width=\linewidth]{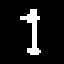}
\end{subfigure}
\begin{subfigure}{0.21\textwidth}
\includegraphics[width=\linewidth]{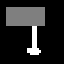}
\end{subfigure}
\begin{subfigure}{0.21\textwidth}
\includegraphics[width=\linewidth]{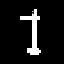}
\end{subfigure}
\begin{subfigure}{0.21\textwidth}
\includegraphics[width=\linewidth]{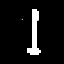}
\end{subfigure}
}

\scalebox{0.8}{
\begin{subfigure}{0.21\textwidth}
\centering
\tiny{GT 1 (70\%)} 
\end{subfigure} 
\begin{subfigure}{0.21\textwidth}
\centering
\tiny{GT 2 (30\%)} 
\end{subfigure}
\begin{subfigure}{0.21\textwidth}
\centering
\tiny{Input} 
\end{subfigure}
\begin{subfigure}{0.21\textwidth}
\centering
\tiny{Output 1} 
\end{subfigure}
\begin{subfigure}{0.21\textwidth}
\centering
\tiny{Output 2} 
\end{subfigure}
}

\captionof{figure}{With >$30\%$ alternate mode data, our model can converge to both the input modes (cols 4-5).} \label{fig:mnist_example_02}
\end{minipage}
\hfill
\begin{minipage}[h]{.25\textwidth}
\captionsetup{size=small}
\scalebox{0.85}{
\begin{subfigure}{0.21\textwidth}
\includegraphics[width=\linewidth]{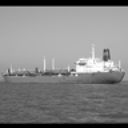}
\end{subfigure} 
\begin{subfigure}{0.21\textwidth}
\includegraphics[width=\linewidth]{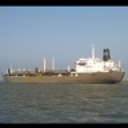}
\end{subfigure}
\begin{subfigure}{0.21\textwidth}
\includegraphics[width=\linewidth]{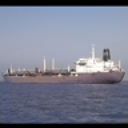}
\end{subfigure}
\begin{subfigure}{0.21\textwidth}
\includegraphics[width=\linewidth]{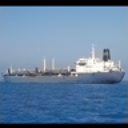}
\end{subfigure}
\begin{subfigure}{0.21\textwidth}
\includegraphics[width=\linewidth]{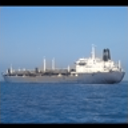}
\end{subfigure}}

\scalebox{0.85}{
\begin{subfigure}{0.21\textwidth}
\includegraphics[width=\linewidth]{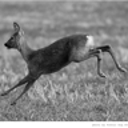}
\end{subfigure} 
\begin{subfigure}{0.21\textwidth}
\includegraphics[width=\linewidth]{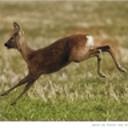}
\end{subfigure}
\begin{subfigure}{0.21\textwidth}
\includegraphics[width=\linewidth]{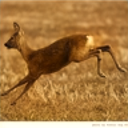}
\end{subfigure}
\begin{subfigure}{0.21\textwidth}
\includegraphics[width=\linewidth]{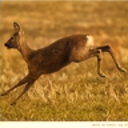}
\end{subfigure}
\begin{subfigure}{0.21\textwidth}
\includegraphics[width=\linewidth]{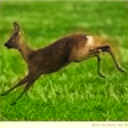}
\end{subfigure}}

\scalebox{0.85}{
\begin{subfigure}{0.21\textwidth}
\includegraphics[width=\linewidth]{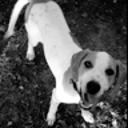}
\end{subfigure} 
\begin{subfigure}{0.21\textwidth}
\includegraphics[width=\linewidth]{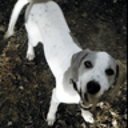}
\end{subfigure}
\begin{subfigure}{0.21\textwidth}
\includegraphics[width=\linewidth]{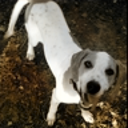}
\end{subfigure}
\begin{subfigure}{0.21\textwidth}
\includegraphics[width=\linewidth]{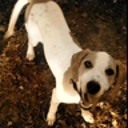}
\end{subfigure}
\begin{subfigure}{0.21\textwidth}
\includegraphics[width=\linewidth]{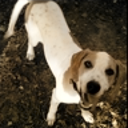}
\end{subfigure}
}

\scalebox{0.85}{
\begin{subfigure}{0.21\textwidth}
\includegraphics[width=\linewidth]{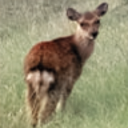}
\end{subfigure} 
\begin{subfigure}{0.21\textwidth}
\includegraphics[width=\linewidth]{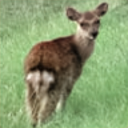}
\end{subfigure}
\begin{subfigure}{0.21\textwidth}
\includegraphics[width=\linewidth]{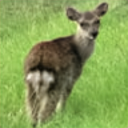}
\end{subfigure}
\begin{subfigure}{0.21\textwidth}
\includegraphics[width=\linewidth]{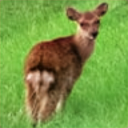}
\end{subfigure}
\begin{subfigure}{0.21\textwidth}
\includegraphics[width=\linewidth]{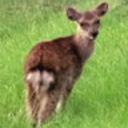}
\end{subfigure}
}

\scalebox{0.85}{
\begin{subfigure}{0.21\textwidth}
\centering
\tiny{itr 0} 
\end{subfigure} 
\begin{subfigure}{0.21\textwidth}
\centering
\tiny{itr 5} 
\end{subfigure}
\begin{subfigure}{0.21\textwidth}
\centering
\tiny{itr 10} 
\end{subfigure}
\begin{subfigure}{0.21\textwidth}
\centering
\tiny{itr 15} 
\end{subfigure}
\begin{subfigure}{0.21\textwidth}
\centering
\tiny{itr 20} 
\end{subfigure}
}
\captionof{figure}{The prediction quality increases as the $z$ traverses to an optimum position at the inference.} \label{fig:colorzoptimiazation}
\end{minipage}
\vspace{2mm}
\end{figure}

\begin{figure}[!hb]
    \begin{minipage}{0.33\textwidth}
        \centering
        \setlength{\tabcolsep}{2pt}
    \scalebox{0.66}{
    \begin{tabular}{|l|c|c||c|}
  \hline
      \multirow{2}{*}{Method} & \multicolumn{2}{c||}{User study} & Turing test   \\
      \cline{2-4}
      & STL & ImageNet & ImageNet \\
      \hline
    Izuka  &      21.89 &         32.28   &  -\\
    Chroma & 32.40 &         31.67   &  - \\ \hline
    Ours &                          \textbf{45.71} & \textbf{36.05} & 31.66\\
      \hline
      \end{tabular}}\vspace{-0.5em}
      \captionof{table}{\small Colorization: Psychophysical study and Turing test results. All performances are in $\%$.}
  \label{tab:userstudy}
    \end{minipage}
    \hfill
    \begin{minipage}{0.65\textwidth}
        \centering \setlength{\tabcolsep}{3pt}
    \scalebox{0.65}{
     \begin{tabular}{|l|c|c|c|c||c|c|c|c|}
  \hline
      \multirow{2}{*}{Method} & \multicolumn{4}{c||}{STL}  & \multicolumn{4}{c|}{ImageNet} \\
      \cline{2-9}
     & LPIP $\downarrow$ & PieAPP $\downarrow$ & SSIM $\uparrow$ & PSNR $\uparrow$ & LPIP $\downarrow$  & PieAPP $\downarrow$  & SSIM $\uparrow$ & PSNR $\uparrow$\\
      \hline
      Izuka  &0.18 & 2.37 & 0.81 & 24.30 &0.17& 2.47 & 0.87 &  18.43\\
      P2P &1.21&2.69&0.73 & 17.80 & 2.01 & 2.80 & 0.87 & 18.43\\
      CIC & 0.18 &  2.81 & 0.71& 22.04 & 0.19 & 2.56 & 0.71 & 19.11\\
      Chroma &0.16& 2.06&0.91&25.57& \textbf{0.16} & 2.13 & 0.90 & 23.33 \\
      \hline
      Ours&\textbf{0.12}&\textbf{1.47}& \textbf{0.95}&\textbf{ 27.03}& \textbf{0.16 }& \textbf{2.04 } & \textbf{0.92} & \textbf{24.51} \\
       Ours (w/o $\rho$) & 0.16 & 1.90 & 0.89&  25.02 & 0.20 & 2.11 & 0.88 & 23.21 \\
      \hline
      \end{tabular}}\vspace{-0.5em}
      \captionof{table}{\small Colorization: Quantitative analysis of our method against the state-of-the-art. Ours perform better on a variety of metrics.}
  \label{tab:colorcomplexity}
    \end{minipage}
\end{figure}

\subsection{Corrupted Image Recovery}
\label{sec:mnist}


We design this task as image completion, i.e., given a masked image as input, our goal is to recover the masked area. Interestingly, we observed that the MNIST dataset, in its original form, does not have a multimodal behaviour, i.e., a fraction of the input image only maps to a single output. Therefore, we modify the training data as follows: first, we overlap the top half of an input image with the top half of another randomly sampled image. We carry out this corruption for $20\%$ of the training data. Corrupted samples are not fixed across epochs. Then, we apply a random sized mask to the top half, and ask the network to predict the missing pixels. We choose two competitive baselines here: our network with the $L_1$ loss and CE.
Fig.~\ref{fig:mnist_example_01} illustrates the predictions. As shown, our model converges to the most probable non-corrupted mode without any ambiguity, while other baselines give sub-optimal results.  In the next experiment, we add a small white box to the top part of the ground-truth images at different rates. At inference, our model was able to converge to both the modes (Fig.~\ref{fig:mnist_example_02}), depending on the initial position of $z$, as the probability of the alternate mode reaches $0.3$.

\subsection{Automatic image colorization}
\label{sec:colorization}
Deep models have tackled this problem using semantic priors \citep{iizuka2016let,vitoria2020chromagan}, adversarial and $L_1$ losses \citep{isola2017image,bicycleGAN,harmonizing},
or by conversion to a discrete form through binning of color values \citep{zhang2016colorful}. Although these methods provide compelling results, several inherent limitations exist: (a) use of semantic priors results in complex models, (b) adversarial loss suffers from drawbacks (see Sec.~\ref{adv_dist_compromise}), and (c) discretization reduces the precision. In contrast, we achieve better results using a simpler model.

\begin{figure}
\captionsetup{size=small}
\centering
\begin{minipage}[h]{1.0\textwidth}
\begin{subfigure}{0.015\textwidth}\hspace{-2mm}
\rotatebox[origin=c]{90}{\scriptsize{GT}}
\end{subfigure}
\begin{subfigure}{0.09\textwidth}
\includegraphics[width=\linewidth]{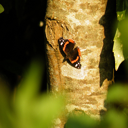}
\end{subfigure} 
\begin{subfigure}{0.09\textwidth}
\includegraphics[width=\linewidth]{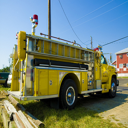}
\end{subfigure}
\begin{subfigure}{0.09\textwidth}
\includegraphics[width=\linewidth]{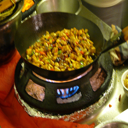}
\end{subfigure} 
\begin{subfigure}{0.09\textwidth}
\includegraphics[width=\linewidth]{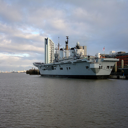}
\end{subfigure} 
\begin{subfigure}{0.09\textwidth}
\includegraphics[width=\linewidth]{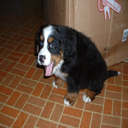}
\end{subfigure} 
\hspace{5pt}
\begin{subfigure}{0.09\textwidth}
\includegraphics[width=\linewidth]{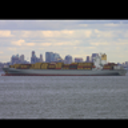}
\end{subfigure}
\begin{subfigure}{0.09\textwidth}
\includegraphics[width=\linewidth]{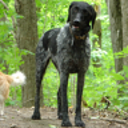}
\end{subfigure}
\begin{subfigure}{0.09\textwidth}
\includegraphics[width=\linewidth]{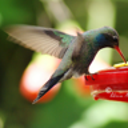}
\end{subfigure}
\begin{subfigure}{0.09\textwidth}
\includegraphics[width=\linewidth]{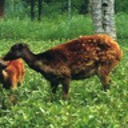}
\end{subfigure}
\begin{subfigure}{0.09\textwidth}
\includegraphics[width=\linewidth]{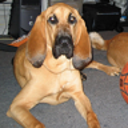}
\end{subfigure}

\begin{subfigure}{0.015\textwidth}\hspace{-2mm}
\rotatebox[origin=c]{90}{\scriptsize{Izuka} }
\end{subfigure}
\begin{subfigure}{0.09\textwidth}
\includegraphics[width=\linewidth]{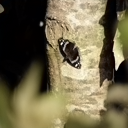}
\end{subfigure} 
\begin{subfigure}{0.09\textwidth}
\includegraphics[width=\linewidth]{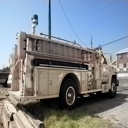}
\end{subfigure} 
\begin{subfigure}{0.09\textwidth}
\includegraphics[width=\linewidth]{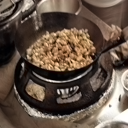}
\end{subfigure} 
\begin{subfigure}{0.09\textwidth}
\includegraphics[width=\linewidth]{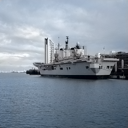}
\end{subfigure} 
\begin{subfigure}{0.09\textwidth}
\includegraphics[width=\linewidth]{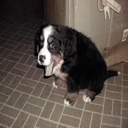}
\end{subfigure} 
\hspace{5pt}
\begin{subfigure}{0.09\textwidth}
\includegraphics[width=\linewidth]{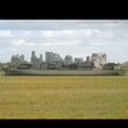}
\end{subfigure}
\begin{subfigure}{0.09\textwidth}
\includegraphics[width=\linewidth]{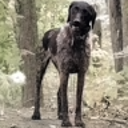}
\end{subfigure}
\begin{subfigure}{0.09\textwidth}
\includegraphics[width=\linewidth]{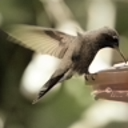}
\end{subfigure}
\begin{subfigure}{0.09\textwidth}
\includegraphics[width=\linewidth]{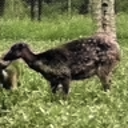}
\end{subfigure}
\begin{subfigure}{0.09\textwidth}
\includegraphics[width=\linewidth]{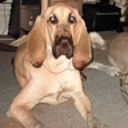}
\end{subfigure}

\begin{subfigure}{0.015\textwidth}\hspace{-2mm}
\rotatebox[origin=c]{90}{\scriptsize{P2P}  }
\end{subfigure}
\begin{subfigure}{0.09\textwidth}
\includegraphics[width=\linewidth]{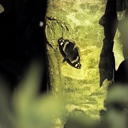}
\end{subfigure} 
\begin{subfigure}{0.09\textwidth}
\includegraphics[width=\linewidth]{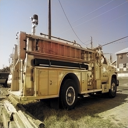}
\end{subfigure} 
\begin{subfigure}{0.09\textwidth}
\includegraphics[width=\linewidth]{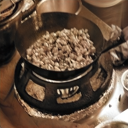}
\end{subfigure} 
\begin{subfigure}{0.09\textwidth}
\includegraphics[width=\linewidth]{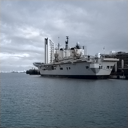}
\end{subfigure} 
\begin{subfigure}{0.09\textwidth}
\includegraphics[width=\linewidth]{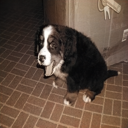}
\end{subfigure} 
\hspace{5pt}
\begin{subfigure}{0.09\textwidth}
\includegraphics[width=\linewidth]{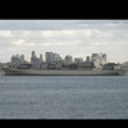}
\end{subfigure}
\begin{subfigure}{0.09\textwidth}
\includegraphics[width=\linewidth]{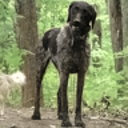}
\end{subfigure}
\begin{subfigure}{0.09\textwidth}
\includegraphics[width=\linewidth]{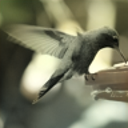}
\end{subfigure}
\begin{subfigure}{0.09\textwidth}
\includegraphics[width=\linewidth]{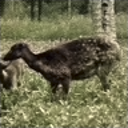}
\end{subfigure}
\begin{subfigure}{0.09\textwidth}
\includegraphics[width=\linewidth]{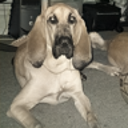}
\end{subfigure}

\begin{subfigure}{0.015\textwidth}\hspace{-2mm}
\rotatebox[origin=c]{90}{\scriptsize{Chroma}  }
\end{subfigure}
\begin{subfigure}{0.09\textwidth}
\includegraphics[width=\linewidth]{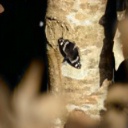}
\end{subfigure} 
\begin{subfigure}{0.09\textwidth}
\includegraphics[width=\linewidth]{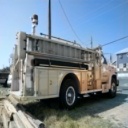}
\end{subfigure}
\begin{subfigure}{0.09\textwidth}
\includegraphics[width=\linewidth]{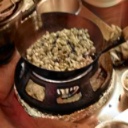}
\end{subfigure} 
\begin{subfigure}{0.09\textwidth}
\includegraphics[width=\linewidth]{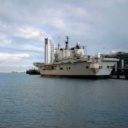}
\end{subfigure} 
\begin{subfigure}{0.09\textwidth}
\includegraphics[width=\linewidth]{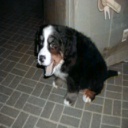}
\end{subfigure} 
\hspace{5pt}
\begin{subfigure}{0.09\textwidth}
\includegraphics[width=\linewidth]{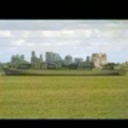}
\end{subfigure}
\begin{subfigure}{0.09\textwidth}
\includegraphics[width=\linewidth]{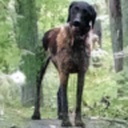}
\end{subfigure}
\begin{subfigure}{0.09\textwidth}
\includegraphics[width=\linewidth]{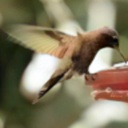}
\end{subfigure}
\begin{subfigure}{0.09\textwidth}
\includegraphics[width=\linewidth]{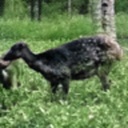}
\end{subfigure}
\begin{subfigure}{0.09\textwidth}
\includegraphics[width=\linewidth]{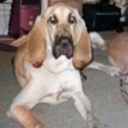}
\end{subfigure}

\begin{subfigure}{0.015\textwidth}\hspace{-2mm}
\rotatebox[origin=c]{90}{\scriptsize{CIC}  }
\end{subfigure}
\begin{subfigure}{0.09\textwidth}
\includegraphics[width=\linewidth]{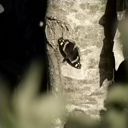}
\end{subfigure} 
\begin{subfigure}{0.09\textwidth}
\includegraphics[width=\linewidth]{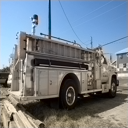}
\end{subfigure} 
\begin{subfigure}{0.09\textwidth}
\includegraphics[width=\linewidth]{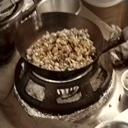}
\end{subfigure} 
\begin{subfigure}{0.09\textwidth}
\includegraphics[width=\linewidth]{figures/imgnet/out27d.png}
\end{subfigure} 
\begin{subfigure}{0.09\textwidth}
\includegraphics[width=\linewidth]{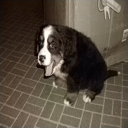}
\end{subfigure} 
\hspace{5pt}
\begin{subfigure}{0.09\textwidth}
\includegraphics[width=\linewidth]{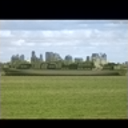}
\end{subfigure}
\begin{subfigure}{0.09\textwidth}
\includegraphics[width=\linewidth]{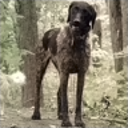}
\end{subfigure}
\begin{subfigure}{0.09\textwidth}
\includegraphics[width=\linewidth]{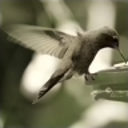}
\end{subfigure}
\begin{subfigure}{0.09\textwidth}
\includegraphics[width=\linewidth]{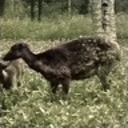}
\end{subfigure}
\begin{subfigure}{0.09\textwidth}
\includegraphics[width=\linewidth]{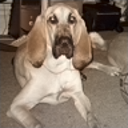}
\end{subfigure}

\begin{subfigure}{0.015\textwidth}\hspace{-2mm}
\rotatebox[origin=c]{90}{\scriptsize{Ours}}
\end{subfigure}
\begin{subfigure}{0.09\textwidth}
\includegraphics[width=\linewidth]{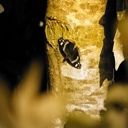}
\end{subfigure} 
\begin{subfigure}{0.09\textwidth}
\includegraphics[width=\linewidth]{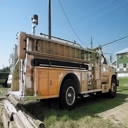}
\end{subfigure} 
\begin{subfigure}{0.09\textwidth}
\includegraphics[width=\linewidth]{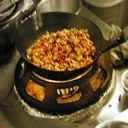}
\end{subfigure}
\begin{subfigure}{0.09\textwidth}
\includegraphics[width=\linewidth]{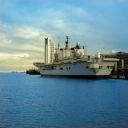}
\end{subfigure}
\begin{subfigure}{0.09\textwidth}
\includegraphics[width=\linewidth]{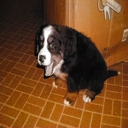}
\end{subfigure}
\hspace{5pt}
\begin{subfigure}{0.09\textwidth}
\includegraphics[width=\linewidth]{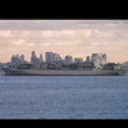}
\end{subfigure}
\begin{subfigure}{0.09\textwidth}
\includegraphics[width=\linewidth]{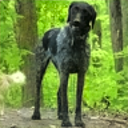}
\end{subfigure}
\begin{subfigure}{0.09\textwidth}
\includegraphics[width=\linewidth]{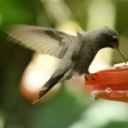}
\end{subfigure}
\begin{subfigure}{0.09\textwidth}
\includegraphics[width=\linewidth]{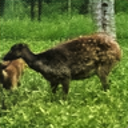}
\end{subfigure}
\begin{subfigure}{0.09\textwidth}
\includegraphics[width=\linewidth]{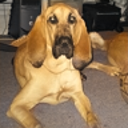}
\end{subfigure}
\vspace{-0.5em}
\caption{Qualitative comparison against the state-of-the-art on ImageNet (left 5 columns) and STL (right 5 columns) datasets. Our model generally produces more vibrant and balanced color distributions.} \label{fig:color}
\end{minipage}
\end{figure}

\begin{figure}
\begin{minipage}[h]{0.54\textwidth}
\captionsetup{size=small}
\centering
\begin{subfigure}{0.15\textwidth}
\includegraphics[width=\linewidth]{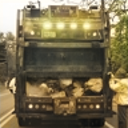}
\end{subfigure} 
\begin{subfigure}{0.15\textwidth}
\includegraphics[width=\linewidth]{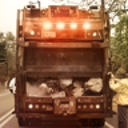}
\end{subfigure}
\begin{subfigure}{0.15\textwidth}
\includegraphics[width=\linewidth]{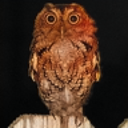}
\end{subfigure}
\begin{subfigure}{0.15\textwidth}
\includegraphics[width=\linewidth]{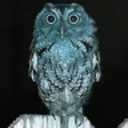}
\end{subfigure}
\begin{subfigure}{0.15\textwidth}
\includegraphics[width=\linewidth]{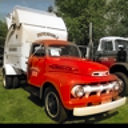}
\end{subfigure}
\begin{subfigure}{0.15\textwidth}
\includegraphics[width=\linewidth]{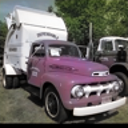}
\end{subfigure}

\begin{subfigure}{0.15\textwidth}
\includegraphics[width=\linewidth]{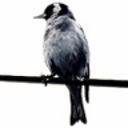}
\end{subfigure} 
\begin{subfigure}{0.15\textwidth}
\includegraphics[width=\linewidth]{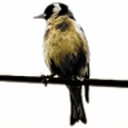}
\end{subfigure}
\begin{subfigure}{0.15\textwidth}
\includegraphics[width=\linewidth]{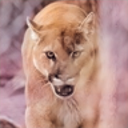}
\end{subfigure}
\begin{subfigure}{0.15\textwidth}
\includegraphics[width=\linewidth]{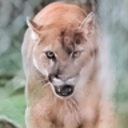}
\end{subfigure}
\begin{subfigure}{0.15\textwidth}
\includegraphics[width=\linewidth]{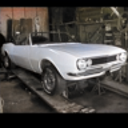}
\end{subfigure}
\begin{subfigure}{0.15\textwidth}
\includegraphics[width=\linewidth]{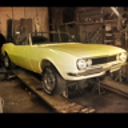}
\end{subfigure}

\begin{subfigure}{0.15\textwidth}
\includegraphics[width=\linewidth]{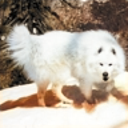}
\end{subfigure} 
\begin{subfigure}{0.15\textwidth}
\includegraphics[width=\linewidth]{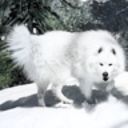}
\end{subfigure}
\begin{subfigure}{0.15\textwidth}
\includegraphics[width=\linewidth]{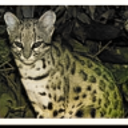}
\end{subfigure}
\begin{subfigure}{0.15\textwidth}
\includegraphics[width=\linewidth]{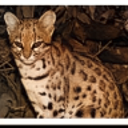}
\end{subfigure}
\begin{subfigure}{0.15\textwidth}
\includegraphics[width=\linewidth]{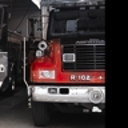}
\end{subfigure}
\begin{subfigure}{0.15\textwidth}
\includegraphics[width=\linewidth]{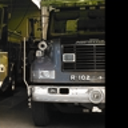}
\end{subfigure}

\begin{subfigure}{0.15\textwidth}
\includegraphics[width=\linewidth]{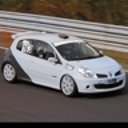}
\end{subfigure} 
\begin{subfigure}{0.15\textwidth}
\includegraphics[width=\linewidth]{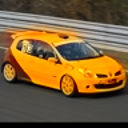}
\end{subfigure}
\begin{subfigure}{0.15\textwidth}
\includegraphics[width=\linewidth]{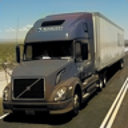}
\end{subfigure}
\begin{subfigure}{0.15\textwidth}
\includegraphics[width=\linewidth]{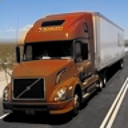}
\end{subfigure}
\begin{subfigure}{0.15\textwidth}
\includegraphics[width=\linewidth]{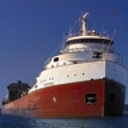}
\end{subfigure}
\begin{subfigure}{0.15\textwidth}
\includegraphics[width=\linewidth]{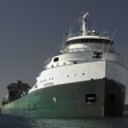}
\end{subfigure}

\begin{subfigure}{0.15\textwidth}
\centering
{\tiny Mode 1}
\end{subfigure}
\begin{subfigure}{0.15\textwidth}
\centering
{\tiny Mode 2}
\end{subfigure}
\begin{subfigure}{0.15\textwidth}
\centering
{\tiny Mode 1}
\end{subfigure}
\begin{subfigure}{0.15\textwidth}
\centering
{\tiny Mode 2}
\end{subfigure}
\begin{subfigure}{0.15\textwidth}
\centering
{\tiny Mode 1}
\end{subfigure}
\begin{subfigure}{0.15\textwidth}
\centering
{\tiny Mode 2}
\end{subfigure}
\caption{Multiple colorization modes predicted by our model for a single input. \emph{(Best viewed in color).}} \label{fig:colorization_vis_01}
\end{minipage}
\hfill
\begin{minipage}{0.445\textwidth}
\begin{subfigure}{0.98\textwidth}
\includegraphics[width=\linewidth]{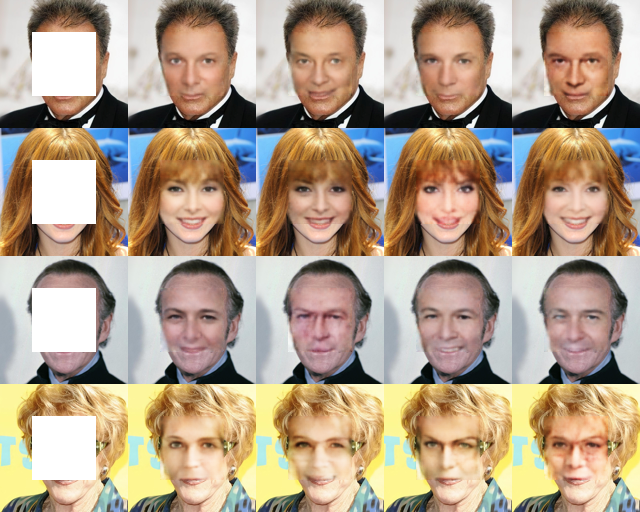}
\end{subfigure}
\captionof{figure}{\small Multi-modality of our predictions on Celeb-HQ dataset. \emph{(Best viewed with zoom)} }  \label{fig:celeb_multy}
\end{minipage}
\end{figure}

The input and the output of the network are $l$ and $(a,b)$ planes respectively (LAB color space). However, since the color distributions of $a$ and $b$ spaces are highly imbalanced over a natural dataset \citep{zhang2016colorful}, we add another constraint to the cost function $E$ to push the predicted $a$ and $b$ colors towards a uniform distribution: $E= \| a_{gt} - a \| + \| b_{gt} - b \| + \lambda (loss_{kl,a} + loss_{kl,b})$, where $loss_{kl,\cdot} = \mathrm{KL}(\cdot || u(0,1))$. Here, $\mathrm{KL}(\cdot||\cdot)$ is the $\mathrm{KL}$ divergence and $u(0,1)$ is a uniform distribution (see App.~\ref{app:colordist}).
Fig.~\ref{fig:color} and Table \ref{tab:colorcomplexity} depict our qualitative and quantitative results, respectively. We demonstrate the superior performance of our method against four metrics: LPIP, PieAPP, SSIM and PSNR (App. \ref{app:evaluation}). Fig.~\ref{fig:colorization_vis_01} depicts examples of multimodality captured by our model (more examples in App. \ref{sec:multimodality}). 
Fig.~\ref{fig:colorzoptimiazation} shows colorization behaviour as the $z$ converges during inference.

\textbf{User study:} We also conduct two user studies to further validate the quality of generated samples (Table~\ref{tab:userstudy}). \textbf{a)} In the \textsc{Psychophysical study}, we present volunteers with batches of 3 images, each generated with a different method. A batch is displayed for 5 secs and the user has to pick the most realistic image. After 5 secs, the next image batch is displayed. \textbf{b)} We conduct a \textsc{Turing test} to validate our output quality against the ground-truth, following the setting proposed by \citet{zhang2016colorful}. The volunteers are presented with a series of paired images (ground-truth and our output). The images are visible for 1 sec, and then the user has an unlimited time to pick the realistic image.

\begin{figure}
\vspace{-1mm}
\begin{minipage}{0.48\textwidth}
\captionsetup{size=small}
\centering
\begin{subfigure}{0.015\textwidth}
\rotatebox[origin=c]{90}{{\tiny GT}}\hspace{-2mm}
\end{subfigure}
\begin{subfigure}{0.14\textwidth}
\includegraphics[width=\linewidth]{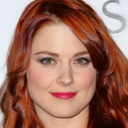}
\end{subfigure}
\begin{subfigure}{0.14\textwidth}
\includegraphics[width=\linewidth]{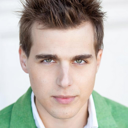}
\end{subfigure}
\begin{subfigure}{0.14\textwidth}
\includegraphics[width=\linewidth]{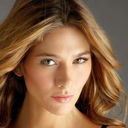}
\end{subfigure}
\hspace{1pt}
\begin{subfigure}{0.14\textwidth}
\includegraphics[width=\linewidth]{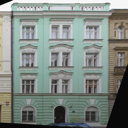}
\end{subfigure}
\begin{subfigure}{0.14\textwidth}
\includegraphics[width=\linewidth]{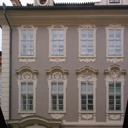}
\end{subfigure}
\begin{subfigure}{0.14\textwidth}
\includegraphics[width=\linewidth]{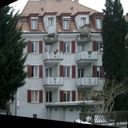}
\end{subfigure}

\begin{subfigure}{0.015\textwidth}
\rotatebox[origin=c]{90}{{\tiny Input}}\hspace{-2mm}
\end{subfigure}
\begin{subfigure}{0.14\textwidth}
\includegraphics[width=\linewidth]{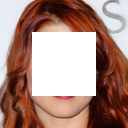}
\end{subfigure}
\begin{subfigure}{0.14\textwidth}
\includegraphics[width=\linewidth]{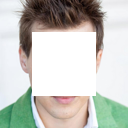}
\end{subfigure}
\begin{subfigure}{0.14\textwidth}
\includegraphics[width=\linewidth]{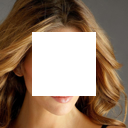}
\end{subfigure}
\hspace{1pt}
\begin{subfigure}{0.14\textwidth}
\includegraphics[width=\linewidth]{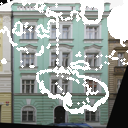}
\end{subfigure}
\begin{subfigure}{0.14\textwidth}
\includegraphics[width=\linewidth]{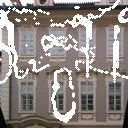}
\end{subfigure}
\begin{subfigure}{0.14\textwidth}
\includegraphics[width=\linewidth]{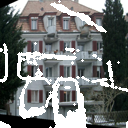}
\end{subfigure}

\begin{subfigure}{0.015\textwidth}
\rotatebox[origin=c]{90}{{\tiny Ours}}\hspace{-2mm}
\end{subfigure}
\begin{subfigure}{0.14\textwidth}
\includegraphics[width=\linewidth]{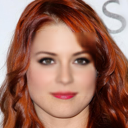}
\end{subfigure}
\begin{subfigure}{0.14\textwidth}
\includegraphics[width=\linewidth]{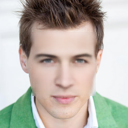}
\end{subfigure}
\begin{subfigure}{0.14\textwidth}
\includegraphics[width=\linewidth]{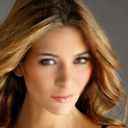}
\end{subfigure}
\hspace{1pt}
\begin{subfigure}{0.14\textwidth}
\includegraphics[width=\linewidth]{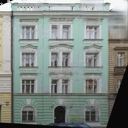}
\end{subfigure}
\begin{subfigure}{0.14\textwidth}
\includegraphics[width=\linewidth]{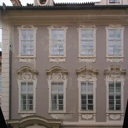}
\end{subfigure}
\begin{subfigure}{0.14\textwidth}
\includegraphics[width=\linewidth]{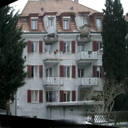}
\end{subfigure}\vspace{-0.5em}
\captionof{figure}{Image completion on Celeb-HQ (left) and Facade (right) datasets. We used fixed center masks and random irregular masks \citep{liu2018image} for Celeb-HQ and Facades datasets, respectively.} \label{fig:celeb}
\end{minipage}
\hfill
\begin{minipage}{0.495\textwidth}
\centering
\captionsetup{size=small}







\begin{subfigure}{0.14\textwidth}
\includegraphics[width=\linewidth]{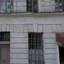}
\end{subfigure}
\begin{subfigure}{0.14\textwidth}
\includegraphics[width=\linewidth]{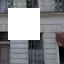}
\end{subfigure}
\begin{subfigure}{0.14\textwidth}
\includegraphics[width=\linewidth]{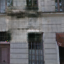}
\end{subfigure}
\begin{subfigure}{0.14\textwidth}
\includegraphics[width=\linewidth]{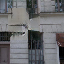}
\end{subfigure}
\begin{subfigure}{0.14\textwidth}
\includegraphics[width=\linewidth]{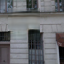}
\end{subfigure}
\begin{subfigure}{0.14\textwidth}
\includegraphics[width=\linewidth]{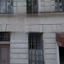}
\end{subfigure}


\begin{subfigure}{0.14\textwidth}
\includegraphics[width=\linewidth]{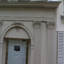}
\end{subfigure}
\begin{subfigure}{0.14\textwidth}
\includegraphics[width=\linewidth]{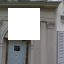}
\end{subfigure}
\begin{subfigure}{0.14\textwidth}
\includegraphics[width=\linewidth]{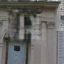}
\end{subfigure}
\begin{subfigure}{0.14\textwidth}
\includegraphics[width=\linewidth]{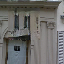}
\end{subfigure}
\begin{subfigure}{0.14\textwidth}
\includegraphics[width=\linewidth]{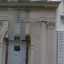}
\end{subfigure}
\begin{subfigure}{0.14\textwidth}
\includegraphics[width=\linewidth]{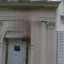}
\end{subfigure}

\begin{subfigure}{0.14\textwidth}
\includegraphics[width=\linewidth]{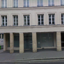}
\end{subfigure}
\begin{subfigure}{0.14\textwidth}
\includegraphics[width=\linewidth]{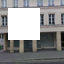}
\end{subfigure}
\begin{subfigure}{0.14\textwidth}
\includegraphics[width=\linewidth]{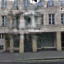}
\end{subfigure}
\begin{subfigure}{0.14\textwidth}
\includegraphics[width=\linewidth]{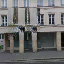}
\end{subfigure}
\begin{subfigure}{0.14\textwidth}
\includegraphics[width=\linewidth]{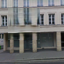}
\end{subfigure}
\begin{subfigure}{0.14\textwidth}
\includegraphics[width=\linewidth]{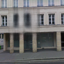}
\end{subfigure}

\begin{subfigure}{0.14\textwidth}
\centering
{\scriptsize GT} \label{fig:4picsd}
\end{subfigure}
\begin{subfigure}{0.14\textwidth}
\centering
{\scriptsize Input} \label{fig:4picsd}
\end{subfigure}
\begin{subfigure}{0.14\textwidth}
\centering
{\scriptsize P2P}
\end{subfigure}
\begin{subfigure}{0.14\textwidth}
\centering
{\scriptsize CA}
\end{subfigure}
\begin{subfigure}{0.14\textwidth}
\centering
{\scriptsize PN} 
\end{subfigure}
\begin{subfigure}{0.14\textwidth}
\centering
{\scriptsize Ours} 
\end{subfigure}
\vspace{-0.5em}
\captionof{figure}{Qualitative comparison for image completion with 25\% missing data (models trained with random sized square masks). 
}  \label{fig:inpainting_example_01}
\end{minipage}
\vspace{-1mm}
\end{figure}

\begin{table}[t]
  \scriptsize
      \centering
      \setlength{\tabcolsep}{3pt}
      \scalebox{1}{
  \begin{tabular}{|c|c|c|c|c|c|c|c|c|c|c|c|c|}
  \hline
      \multirow{2}{*}{Method} & \multicolumn{4}{c|}{10\% corruption} & \multicolumn{4}{c|}{15\% corruption} & \multicolumn{4}{c|}{25\% corruption}\\
      \cline{2-13}
      & LPIP $\downarrow$ & PieAPP $\downarrow$  &  PSNR $\uparrow$ & SSIM $\uparrow$ & LPIP $\downarrow$ & PieAPP $\downarrow$ & PSNR $\uparrow$ & SSIM $\uparrow$ & LPIP $\downarrow$ & PieAPP $\downarrow$  & PSNR $\uparrow$ & SSIM $\uparrow$  \\
       \hline
       DSGAN  & 0.101 & 1.577 & 20.13 & 0.67 & 0.189 & 2.970 & 18.45 & 0.55 & 0.213 & 3.54 & 16.44 & 0.49  \\
      PN   & \textbf{0.045}  &\textbf{0.639} & 27.11 & 0.88  & 0.084 & \textbf{0.680} & 20.50 & 0.71 & 0.147 & 0.764 & 19.41 & 0.63 
      \\
      CE  & 0.092 & 1.134 & 22.34 & 0.71 & 0.134 & 2.134 & 19.11 & 0.63 & 0.189 & 2.717 & 17.44 & 0.51\\
      P2P  & 0.074 & 0.942 & 22.33 & 0.79 & 0.101 & 1.971 & 19.34 & 0.70 & 0.185 & 2.378 & 17.81 & 0.57\\
      CA  & 0.048 & 0.731 & 26.45 & 0.83 & 0.091 & 0.933 & 20.12 & 0.72 & 0.166 & 0.822 & 21.43 & 0.72  \\
      \hline
           Ours (w/o $\rho$) & 0.053 & 0.799 & 27.77 & 0.83 & 0.085 & 0.844 & 23.22 & 0.76 &  0.141 &  0.812 & 22.31 & 0.74\\
      Ours  & 0.051 & 0.727 & \textbf{27.83} & \textbf{0.89} & \textbf{0.080} & 0.740 & \textbf{26.43} & \textbf{0.80} &  \textbf{0.129} &  \textbf{0.760} & \textbf{24.16} & \textbf{0.77}\\
      \hline
      \end{tabular}}\vspace{-0.8em}
      \caption{\small \emph{Image completion:} Quantitative analysis of our method against state-of-the-art on a variety of metrics. 
      }
  \label{tab:inpainting_comparison}

\end{table}

  
      



\begin{figure}
\vspace{-2mm}
\begin{minipage}{0.39\textwidth}
\captionsetup{size=small}
\centering
\begin{subfigure}{0.14\textwidth}
\includegraphics[width=\linewidth]{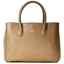}
\end{subfigure}
\begin{subfigure}{0.14\textwidth}
\includegraphics[width=\linewidth]{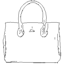}
\end{subfigure}
\begin{subfigure}{0.14\textwidth}
\includegraphics[width=\linewidth]{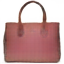}
\end{subfigure}
\begin{subfigure}{0.14\textwidth}
\includegraphics[width=\linewidth]{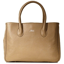}
\end{subfigure}
\begin{subfigure}{0.14\textwidth}
\includegraphics[width=\linewidth]{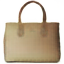}
\end{subfigure}
\begin{subfigure}{0.14\textwidth}
\includegraphics[width=\linewidth]{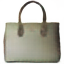}
\end{subfigure}

\begin{subfigure}{0.14\textwidth}
\includegraphics[width=\linewidth]{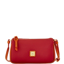}
\end{subfigure}
\begin{subfigure}{0.14\textwidth}
\includegraphics[width=\linewidth]{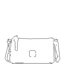}
\end{subfigure}
\begin{subfigure}{0.14\textwidth}
\includegraphics[width=\linewidth]{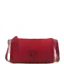}
\end{subfigure}
\begin{subfigure}{0.14\textwidth}
\includegraphics[width=\linewidth]{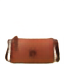}
\end{subfigure}
\begin{subfigure}{0.14\textwidth}
\includegraphics[width=\linewidth]{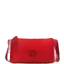}
\end{subfigure}
\begin{subfigure}{0.14\textwidth}
\includegraphics[width=\linewidth]{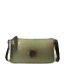}
\end{subfigure}

\begin{subfigure}{0.14\textwidth}
\includegraphics[width=\linewidth]{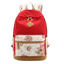}
\end{subfigure}
\begin{subfigure}{0.14\textwidth}
\includegraphics[width=\linewidth]{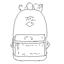}
\end{subfigure}
\begin{subfigure}{0.14\textwidth}
\includegraphics[width=\linewidth]{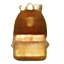}
\end{subfigure}
\begin{subfigure}{0.14\textwidth}
\includegraphics[width=\linewidth]{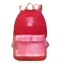}
\end{subfigure}
\begin{subfigure}{0.14\textwidth}
\includegraphics[width=\linewidth]{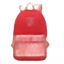}
\end{subfigure}
\begin{subfigure}{0.14\textwidth}
\includegraphics[width=\linewidth]{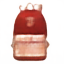}
\end{subfigure}

\begin{subfigure}{0.14\textwidth}
\includegraphics[width=\linewidth]{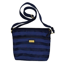}
\end{subfigure}
\begin{subfigure}{0.14\textwidth}
\includegraphics[width=\linewidth]{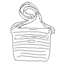}
\end{subfigure}
\begin{subfigure}{0.14\textwidth}
\includegraphics[width=\linewidth]{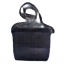}
\end{subfigure}
\begin{subfigure}{0.14\textwidth}
\includegraphics[width=\linewidth]{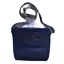}
\end{subfigure}
\begin{subfigure}{0.14\textwidth}
\includegraphics[width=\linewidth]{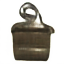}
\end{subfigure}
\begin{subfigure}{0.14\textwidth}
\includegraphics[width=\linewidth]{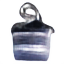}
\end{subfigure}
\captionof{figure}{Translation from hand bag sketches to images.} \label{fig:bag}
\end{minipage}
\hfill
\begin{minipage}{0.39\textwidth}
\centering
\captionsetup{size=small}
\begin{subfigure}{0.14\textwidth}
\includegraphics[width=\linewidth]{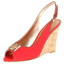}
\end{subfigure}
\begin{subfigure}{0.14\textwidth}
\includegraphics[width=\linewidth]{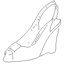}
\end{subfigure}
\begin{subfigure}{0.14\textwidth}
\includegraphics[width=\linewidth]{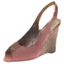}
\end{subfigure}
\begin{subfigure}{0.14\textwidth}
\includegraphics[width=\linewidth]{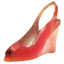}
\end{subfigure}
\begin{subfigure}{0.14\textwidth}
\includegraphics[width=\linewidth]{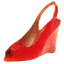}
\end{subfigure}
\begin{subfigure}{0.14\textwidth}
\includegraphics[width=\linewidth]{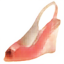}
\end{subfigure}

\begin{subfigure}{0.14\textwidth}
\includegraphics[width=\linewidth]{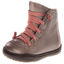}
\end{subfigure}
\begin{subfigure}{0.14\textwidth}
\includegraphics[width=\linewidth]{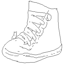}
\end{subfigure}
\begin{subfigure}{0.14\textwidth}
\includegraphics[width=\linewidth]{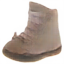}
\end{subfigure}
\begin{subfigure}{0.14\textwidth}
\includegraphics[width=\linewidth]{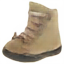}
\end{subfigure}
\begin{subfigure}{0.14\textwidth}
\includegraphics[width=\linewidth]{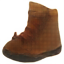}
\end{subfigure}
\begin{subfigure}{0.14\textwidth}
\includegraphics[width=\linewidth]{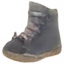}
\end{subfigure}

\begin{subfigure}{0.14\textwidth}
\includegraphics[width=\linewidth]{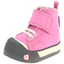}
\end{subfigure}
\begin{subfigure}{0.14\textwidth}
\includegraphics[width=\linewidth]{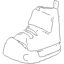}
\end{subfigure}
\begin{subfigure}{0.14\textwidth}
\includegraphics[width=\linewidth]{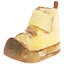}
\end{subfigure}
\begin{subfigure}{0.14\textwidth}
\includegraphics[width=\linewidth]{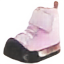}
\end{subfigure}
\begin{subfigure}{0.14\textwidth}
\includegraphics[width=\linewidth]{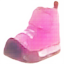}
\end{subfigure}
\begin{subfigure}{0.14\textwidth}
\includegraphics[width=\linewidth]{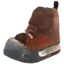}
\end{subfigure}

\begin{subfigure}{0.14\textwidth}
\includegraphics[width=\linewidth]{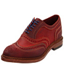}
\end{subfigure}
\begin{subfigure}{0.14\textwidth}
\includegraphics[width=\linewidth]{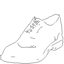}
\end{subfigure}
\begin{subfigure}{0.14\textwidth}
\includegraphics[width=\linewidth]{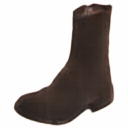}
\end{subfigure}
\begin{subfigure}{0.14\textwidth}
\includegraphics[width=\linewidth]{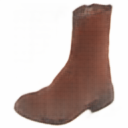}
\end{subfigure}
\begin{subfigure}{0.14\textwidth}
\includegraphics[width=\linewidth]{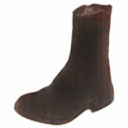}
\end{subfigure}
\begin{subfigure}{0.14\textwidth}
\includegraphics[width=\linewidth]{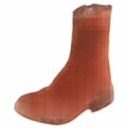}
\end{subfigure}

\captionof{figure}{Translation from hand shoe sketches to images.}  \label{fig:shoes}
\end{minipage}
\hfill
\begin{minipage}{0.18\textwidth}
\centering
\captionsetup{size=small}

\begin{subfigure}{0.3\textwidth}
\includegraphics[width=\linewidth]{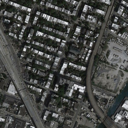}
\end{subfigure} 
\begin{subfigure}{0.3\textwidth}
\includegraphics[width=\linewidth]{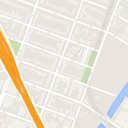}
\end{subfigure} 
\begin{subfigure}{0.3\textwidth}
\includegraphics[width=\linewidth]{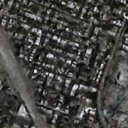}
\end{subfigure} 

\begin{subfigure}{0.3\textwidth}
\includegraphics[width=\linewidth]{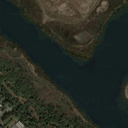}
\end{subfigure} 
\begin{subfigure}{0.3\textwidth}
\includegraphics[width=\linewidth]{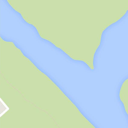}
\end{subfigure} 
\begin{subfigure}{0.3\textwidth}
\includegraphics[width=\linewidth]{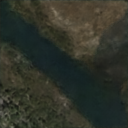}
\end{subfigure}

\begin{subfigure}{0.3\textwidth}
\includegraphics[width=\linewidth]{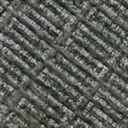}
\end{subfigure} 
\begin{subfigure}{0.3\textwidth}
\includegraphics[width=\linewidth]{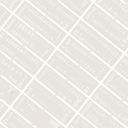}
\end{subfigure} 
\begin{subfigure}{0.3\textwidth}
\includegraphics[width=\linewidth]{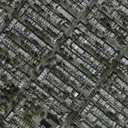}
\end{subfigure}
\vspace{-0.6em}
\captionof{figure}{Map to aerial image translation. \emph{From left}: GT, Input and Output. Also see App.~
\ref{sec:maptophoto}.} \label{fig:map}
\end{minipage}
\end{figure}

\subsection{Image completion}
\label{sec:completion}
In this case, we show that our generic model outperforms a similar capacity GAN (CE) as well as task-specific GANs. In contrast to task-specific models, we do not use any 
domain-specific modifications to make our outputs perceptually pleasing.
We observe that with random irregular and fixed-sized masks, all the models perform well, and we were not able to visually observe a considerable difference (Fig.~\ref{fig:celeb}, see App. \ref{app:imagecomplete} for more results). Therefore, we presented models with a more challenging task: train with random sized square-shaped masks and evaluate the performance against masks of varying sizes. Fig.~\ref{fig:inpainting_example_01} illustrates qualitative results of the models with 25\% masked data. 
As evident, our model recovers details more accurately compared to the state-of-the-art. Notably, all models produce comparable results when trained with a fixed sized center mask, but find this setting more challenging. Table \ref{tab:inpainting_comparison} includes a  quantitative comparison. Observe that in the case of smaller sized masks, PN performs slightly better than ours, but worse otherwise. We also evaluate the learned features of the models against a downstream classification task (Table~\ref{tab:downstream}). First, we train all the models on Facades \citep{facade_dataset} against random masks, and then apply the trained models on CIFAR10~\citep{cifar} to extract bottleneck features, and finally pass them through a FC layer for classification (App. \ref{app:ssl}).  We compare PN and ours against an oracle (AlexNet features pre-trained on ImageNet) and show our model performs closer to the oracle.  

\begin{figure}[t]
\begin{minipage}{0.45\textwidth}
\captionsetup{size=small}
\includegraphics[width=\linewidth]{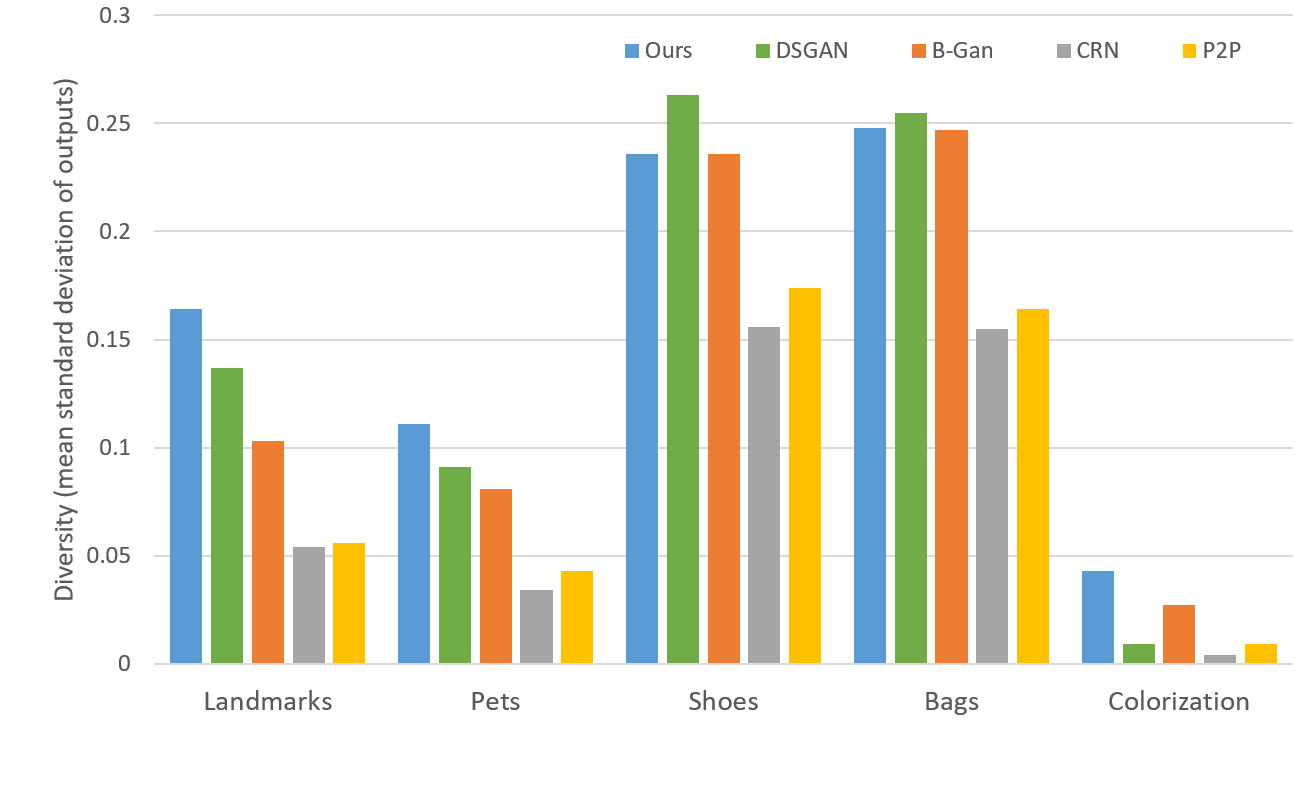} \vspace{-2.5em}
\captionof{figure}{Diversity: Quantitative comparisons.}. \label{fig:diversity}
\end{minipage}
\hfill
\begin{minipage}{0.26\textwidth}
\captionsetup{size=small}
\includegraphics[width=\linewidth]{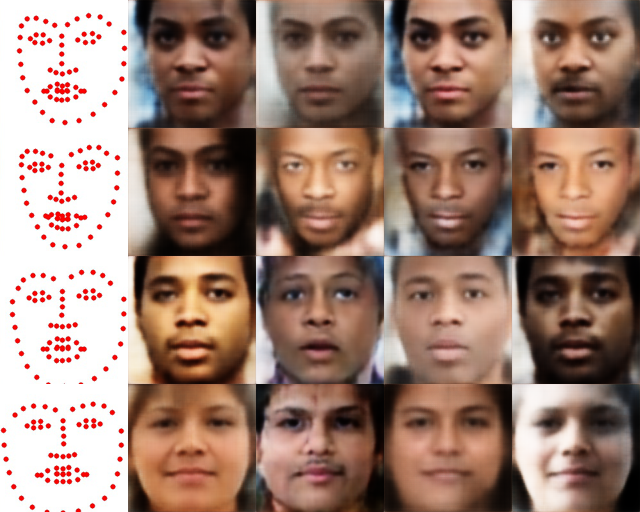} 
\captionof{figure}{{Translation from facial landmarks to faces.}} \label{fig:landmarks}
\end{minipage}
\hfill
\begin{minipage}{0.26\textwidth}
\captionsetup{size=small}
\includegraphics[width=\linewidth]{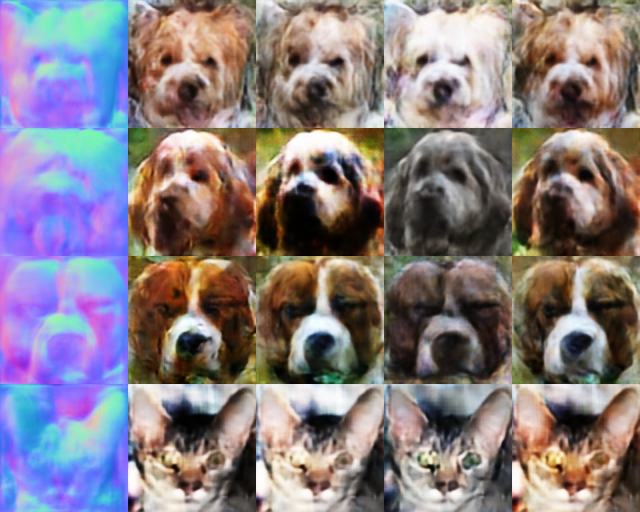} 
\captionof{figure}{{Translation from surface-normals to pet faces.}} \label{fig:pets}

\end{minipage}
\end{figure}

\begin{figure}[t]
\begin{minipage}{0.58\textwidth}
\captionsetup{size=small}
\centering
\begin{subfigure}{0.18\textwidth}
\includegraphics[width=\linewidth]{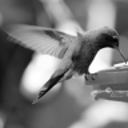}
\end{subfigure} 
\begin{subfigure}{0.18\textwidth}
\includegraphics[width=\linewidth]{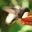}
\end{subfigure} 
\begin{subfigure}{0.18\textwidth}
\includegraphics[width=\linewidth]{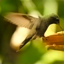}
\end{subfigure} 
\begin{subfigure}{0.18\textwidth}
\includegraphics[width=\linewidth]{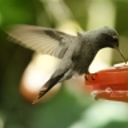}
\end{subfigure} 
\begin{subfigure}{0.18\textwidth}
\includegraphics[width=\linewidth]{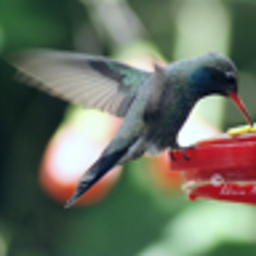}
\end{subfigure}

\begin{subfigure}{0.18\textwidth}
\includegraphics[width=\linewidth]{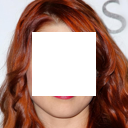}
\end{subfigure} 
\begin{subfigure}{0.18\textwidth}
\includegraphics[width=\linewidth]{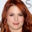}
\end{subfigure} 
\begin{subfigure}{0.18\textwidth}
\includegraphics[width=\linewidth]{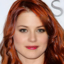}
\end{subfigure} 
\begin{subfigure}{0.18\textwidth}
\includegraphics[width=\linewidth]{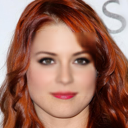}
\end{subfigure} 
\begin{subfigure}{0.18\textwidth}
\includegraphics[width=\linewidth]{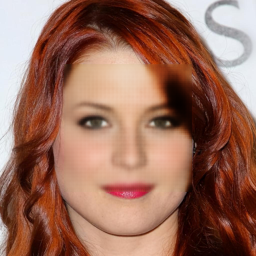}
\end{subfigure}

\begin{subfigure}{0.18\textwidth}
\centering
\tiny{Input}
\end{subfigure}
\begin{subfigure}{0.18\textwidth}
\centering
\tiny{$32 \times 32$}
\end{subfigure}
\begin{subfigure}{0.18\textwidth}
\centering
\tiny{$64 \times 64$}
\end{subfigure}
\begin{subfigure}{0.18\textwidth}
\centering
\tiny{$128 \times 128$}
\end{subfigure}
\begin{subfigure}{0.18\textwidth}
\centering
\tiny{$256 \times 256$}
\end{subfigure}

\captionof{figure}{Scalability:  we subsequently add layers to the architecture to be trained on increasingly high-resolution inputs} \label{fig:scalability}
\end{minipage}
\hfill
\begin{minipage}{0.4\textwidth}
\captionsetup{size=small}
\centering
\scalebox{0.9}{
\begin{subfigure}{0.18\textwidth}
\includegraphics[width=\linewidth]{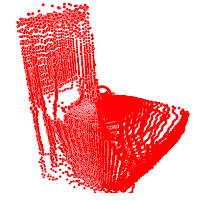} 
\end{subfigure} 
\begin{subfigure}{0.18\textwidth}
\includegraphics[width=\linewidth]{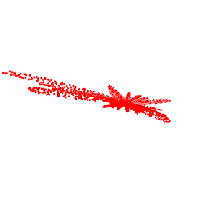} 
\end{subfigure} 
\begin{subfigure}{0.18\textwidth}
\includegraphics[width=\linewidth]{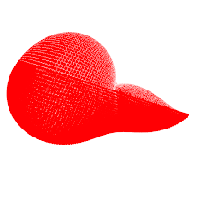} 
\end{subfigure} 
\begin{subfigure}{0.18\textwidth}
\includegraphics[width=\linewidth]{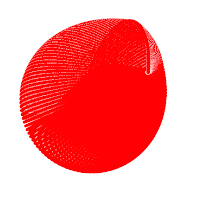} 
\end{subfigure} 
\begin{subfigure}{0.18\textwidth}
\includegraphics[width=\linewidth]{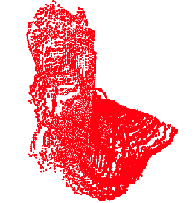} 
\end{subfigure} 
}

\scalebox{0.9}{
\begin{subfigure}{0.18\textwidth}
\includegraphics[width=\linewidth]{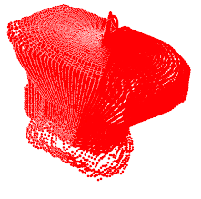} 
\end{subfigure} 
\begin{subfigure}{0.18\textwidth}
\includegraphics[width=\linewidth]{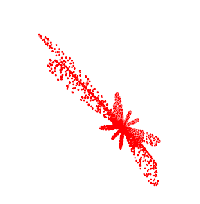} 
\end{subfigure} 
\begin{subfigure}{0.18\textwidth}
\includegraphics[width=\linewidth]{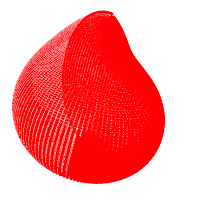} 
\end{subfigure} 
\begin{subfigure}{0.18\textwidth}
\includegraphics[width=\linewidth]{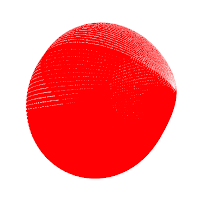} 
\end{subfigure} 
\begin{subfigure}{0.18\textwidth}
\includegraphics[width=\linewidth]{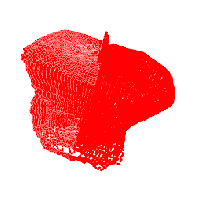} 
\end{subfigure} 
}

\scalebox{0.9}{
\begin{subfigure}{0.18\textwidth}
\includegraphics[width=\linewidth]{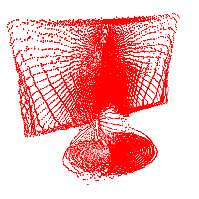} 
\end{subfigure} 
\begin{subfigure}{0.18\textwidth}
\includegraphics[width=\linewidth]{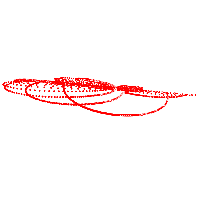} 
\end{subfigure} 
\begin{subfigure}{0.18\textwidth}
\includegraphics[width=\linewidth]{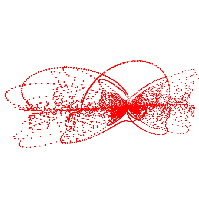} 
\end{subfigure} 
\begin{subfigure}{0.18\textwidth}
\includegraphics[width=\linewidth]{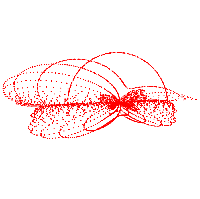} 
\end{subfigure} 
\begin{subfigure}{0.18\textwidth}
\includegraphics[width=\linewidth]{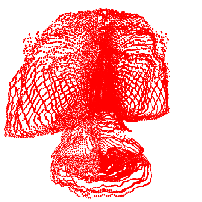} 
\end{subfigure}
}

\scalebox{0.9}{
\begin{subfigure}{0.18\textwidth}
\centering
\tiny{GT}
\end{subfigure}
\begin{subfigure}{0.18\textwidth}
\centering
\tiny{Input}
\end{subfigure}
\begin{subfigure}{0.18\textwidth}
\centering
\tiny{CE}
\end{subfigure}
\begin{subfigure}{0.18\textwidth}
\centering
\tiny{cVAE}
\end{subfigure}
\begin{subfigure}{0.18\textwidth}
\centering
\tiny{Ours}
\end{subfigure}
}
\vspace{-0.5em}
\captionof{figure}{Qualitative comparison of 3D spectral denoising. The results are converted to the spatial domain for a clear visualization.} \label{fig:3ddenoise}

\end{minipage}
\end{figure}

\subsubsection{Diversity and other compelling attributes}
\begin{figure}
\begin{minipage}{0.28\textwidth}
    \captionsetup{size=small}
    \includegraphics[width=1\linewidth]{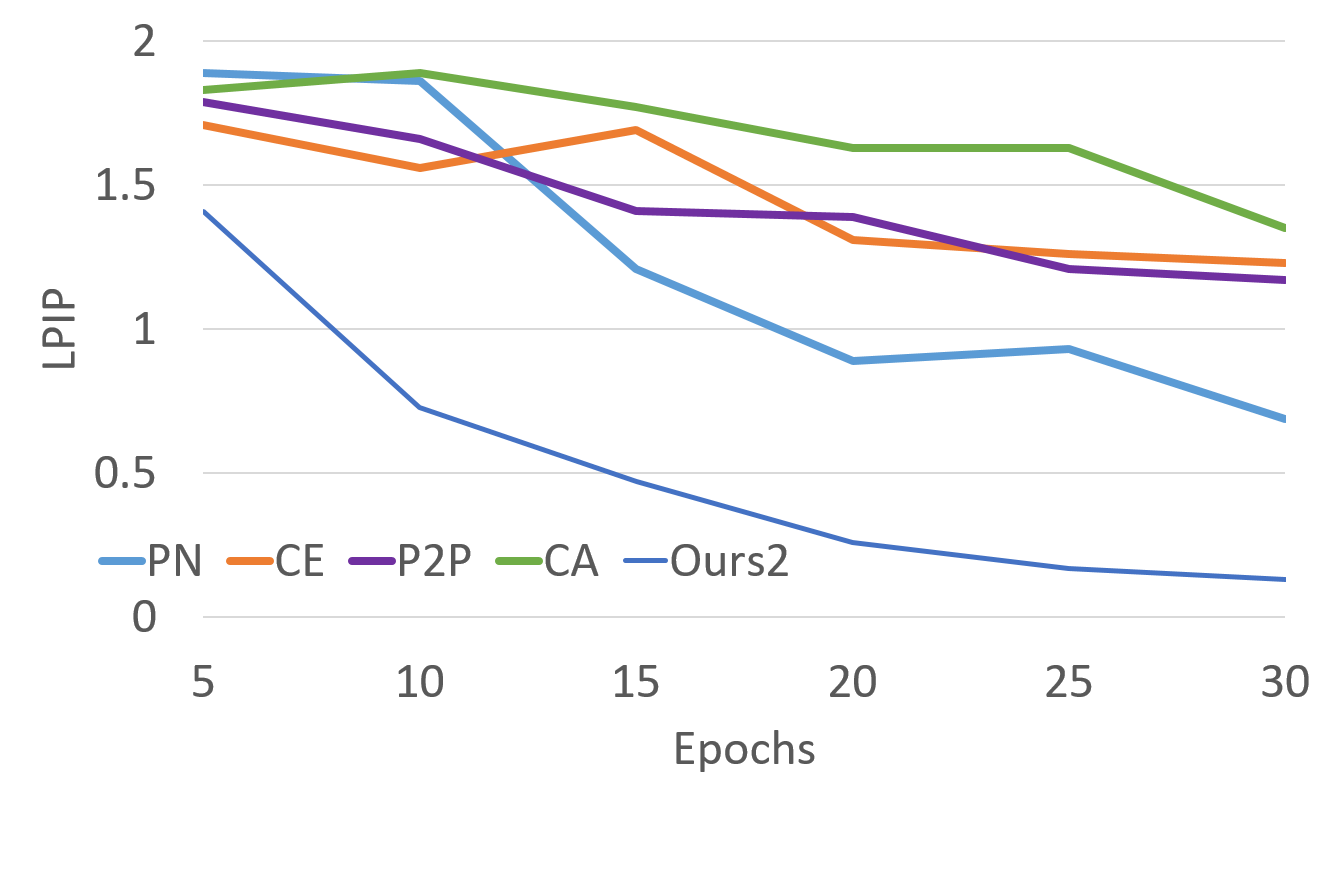}\vspace{-1.5em}
    \captionof{figure}{Convergence on image completion (Paris view). Our model exhibits rapid and stable convergence compared to state-of-the-art (PN, CE, P2P, CA).} \label{fig:convergence}
\end{minipage}
\hfill
\begin{minipage}{0.38\textwidth}
\scriptsize \setlength{\tabcolsep}{6pt}
  \centering
  \begin{tabular}{|l|c|c|c|c|}
  \hline
      Method & M10 & M40  \\
      \hline
       \cite{sharma2016vconv} & 80.5\% & 75.5\% \\
      \cite{han2019view} & 92.2\% & 90.2\% \\
     \cite{achlioptas2017representation} & \textbf{95.3\%} & 85.7\% \\
       \cite{yang2018foldingnet} & 94.4\% & 88.4\% \\
       \cite{sauder2019self} & 94.5\% & 90.6\%   \\
      \cite{ramasinghe2019spectral} & 93.1\% & - \\
      \cite{khan2019unsupervised} & 92.2\% & - \\
      \hline
      Ours & 92.4\% & \textbf{90.9\%} \\
      \hline
      \end{tabular}
      \captionof{table}{\small Downstream 3D object classification results on ModelNet10 and ModelNet40 using features learned in an unsupervised manner. All results in \% accuracy.}
  \label{tab:3dssl}
\end{minipage}
\hfill
\begin{minipage}{0.29\textwidth}
\scriptsize
\begin{minipage}{1\textwidth}
    \centering \setlength{\tabcolsep}{4pt}
  \begin{tabular}{|l|c|c|}
  \hline
      Method & Pretext & Acc. (\%)   \\
      \hline
      ResNet$^*$  & ImageNet Cls. & 74.2 \\
      PN  & Im. Completion & 40.3\\
      Ours  & Im. Completion & \textbf{62.5} \\
      \hline
      \end{tabular}\vspace{-0.9em}
      \captionof{table}{\small Comparison on downstream task (CIFAR10 cls). ($^*$) denotes the oracle case.  
      }
  \label{tab:downstream}\vspace{6pt}
\end{minipage}
\begin{minipage}{1\textwidth}
 \setlength{\tabcolsep}{10pt}
    \centering
    \begin{tabular}{|l|c|c|}
  \hline
      Method & M10 & M40 \\
      \hline
      CE  & 10.3 & 4.6  \\
      cVAE & 8.7 & 4.2 \\
      \hline
      Ours & \textbf{84.2 } & \textbf{79.4} \\
      \hline
    \end{tabular}\vspace{-0.7em}
    \captionof{table}{\small Reconstruction mAP of 3d spectral denoising.}
  \label{tab:3dreconstruct}
\end{minipage}
\end{minipage}
\end{figure}

We also experiment on a diverse set of image translation tasks to demonstrate our generalizability. Fig.~\ref{fig:bag}, \ref{fig:shoes}, \ref{fig:map}, \ref{fig:landmarks} and \ref{fig:pets} illustrate the qualitative results of \textit{sketch-to-handbag}, \textit{sketch-to-shoes}, \textit{map-to-arial}, \textit{lanmarks-to-faces} and \textit{surface-normals-to-pets} tasks. 
Fig.~\ref{fig:colorization_vis_01}, \ref{fig:celeb_multy}, \ref{fig:bag}, \ref{fig:shoes}, \ref{fig:landmarks} and \ref{fig:pets} show the ability of our model to converge to  multiple modes, depending on the $z$  initialization. Fig.~\ref{fig:diversity} demonstrates the quantitative comparison against other models.  
See App.~\ref{sec:multimodality} for further details on experiments.
Another appealing feature of our model is its strong convergence properties irrespective of the architecture, hence, \emph{scalability} to different input sizes. Fig.~\ref{fig:scalability} shows examples from image completion and colorization for varying input sizes. We add layers to the architecture to be trained on increasingly high-resolution inputs, where our model was able to converge to optimal modes at each scale (App. \ref{app:scalility}). Fig.~\ref{fig:convergence} demonstrates our faster and stable \emph{convergence}. 

\subsection{Denoising of 3D objects in spectral space}
\label{sec:denoising}
 Spectral moments of 3D objects provide a compact representation, and help building light-weight networks \citep{ramasinghe2020blended,ramasinghe2019representation,cohen2018spherical,  esteves2018learning}. 
 However, spectral information of 3D objects has not been used before for self-supervised learning, a key reason being the difficulty of learning representations in the spectral domain due to the complex structure 
 and unbounded spectral coefficients. Here, we present an efficient pretext task that is conducted in {the} spectral domain: denoising 3D spectral maps. We use two types of spectral spaces: spherical harmonics and Zernike polynomials (App. \ref{app:spectral}).  We first convert the 3D point clouds to spherical harmonic coefficients, arrange the values as a 2D map, and mask or add noise to a map portion (App.~\ref{sec:appdenoising}). The goal is to recover the original spectral map. Fig.~\ref{fig:3ddenoise} and Table \ref{tab:3dreconstruct} depicts our qualitative and quantitative results. We perform favorably well against other methods. To evaluate the learned features, we use Zernike polynomials, as they are more discriminative compared to spherical harmonics \citep{ramasinghe2019volumetric}. We first train the network on the 55k ShapeNet objects by denoising spectral maps, and then apply the trained network on the ModelNet10 \& 40. The features are then extracted from the bottleneck (similar to Sec.~\ref{sec:completion}), and fed to a FC classifier (Table \ref{tab:3dssl}). We achieve the state-of-the-art results in ModelNet40 with a simple pretext task.

\section{Conclusion}

Conditional generation in multimodal domains is a challenging task due to its ill-posed nature. In this paper, we propose a novel generative framework that  minimize a family of cost functions during training. Further, it observes the convergence patterns of latent variables and applies this knowledge during inference to traverse to multiple output modes during inference. Despite using a simple and generic architecture, we show impressive results on a diverse set of tasks. The proposed approach demonstrates faster convergence, scalability, generalizability, diversity and superior representation learning capability for downstream tasks. 


\bibliographystyle{iclr2021_conference}
\bibliography{iclr2021_conference}

\begin{thebibliography}{73}
\providecommand{\natexlab}[1]{#1}
\providecommand{\url}[1]{\texttt{#1}}
\expandafter\ifx\csname urlstyle\endcsname\relax
  \providecommand{\doi}[1]{doi: #1}\else
  \providecommand{\doi}{doi: \begingroup \urlstyle{rm}\Url}\fi

\bibitem[Achlioptas et~al.(2017)Achlioptas, Diamanti, Mitliagkas, and
  Guibas]{achlioptas2017representation}
Panos Achlioptas, Olga Diamanti, Ioannis Mitliagkas, and Leonidas Guibas.
\newblock Representation learning and adversarial generation of 3d point
  clouds.
\newblock \emph{arXiv preprint arXiv:1707.02392}, 2017.

\bibitem[Arora \& Zhang(2017)Arora and Zhang]{arora2017gans}
Sanjeev Arora and Yi~Zhang.
\newblock Do gans actually learn the distribution? an empirical study.
\newblock \emph{arXiv preprint arXiv:1706.08224}, 2017.

\bibitem[Arora et~al.(2018)Arora, Risteski, and Zhang]{arora2018gans}
Sanjeev Arora, Andrej Risteski, and Yi~Zhang.
\newblock Do {GAN}s learn the distribution? some theory and empirics.
\newblock In \emph{International Conference on Learning Representations}, 2018.

\bibitem[Bansal et~al.(2017{\natexlab{a}})Bansal, Chen, Russell, Gupta, and
  Ramanan]{bansal2017pixelnet}
Aayush Bansal, Xinlei Chen, Bryan Russell, Abhinav Gupta, and Deva Ramanan.
\newblock Pixelnet: Representation of the pixels, by the pixels, and for the
  pixels.
\newblock \emph{arXiv preprint arXiv:1702.06506}, 2017{\natexlab{a}}.

\bibitem[Bansal et~al.(2017{\natexlab{b}})Bansal, Sheikh, and
  Ramanan]{bansal2017pixelnn}
Aayush Bansal, Yaser Sheikh, and Deva Ramanan.
\newblock Pixelnn: Example-based image synthesis.
\newblock \emph{arXiv preprint arXiv:1708.05349}, 2017{\natexlab{b}}.

\bibitem[Bao et~al.(2017)Bao, Chen, Wen, Li, and Hua]{cvaegan}
Jianmin Bao, Dong Chen, Fang Wen, Houqiang Li, and Gang Hua.
\newblock Cvae-gan: Fine-grained image generation through asymmetric training.
\newblock In \emph{The IEEE International Conference on Computer Vision
  (ICCV)}, Oct 2017.

\bibitem[Barnett(2018)]{barnett2018convergence}
Samuel~A Barnett.
\newblock Convergence problems with generative adversarial networks (gans).
\newblock \emph{arXiv preprint arXiv:1806.11382}, 2018.

\bibitem[Chen \& Koltun(2017{\natexlab{a}})Chen and
  Koltun]{chen2017photographic}
Qifeng Chen and Vladlen Koltun.
\newblock Photographic image synthesis with cascaded refinement networks.
\newblock In \emph{Proceedings of the IEEE international conference on computer
  vision}, pp.\  1511--1520, 2017{\natexlab{a}}.

\bibitem[Chen \& Koltun(2017{\natexlab{b}})Chen and Koltun]{min_loss}
Qifeng Chen and Vladlen Koltun.
\newblock Photographic image synthesis with cascaded refinement networks.
\newblock In \emph{The IEEE International Conference on Computer Vision
  (ICCV)}, Oct 2017{\natexlab{b}}.

\bibitem[Chu et~al.(2020)Chu, Minami, and Fukumizu]{chu2020smoothness}
Casey Chu, Kentaro Minami, and Kenji Fukumizu.
\newblock Smoothness and stability in gans.
\newblock \emph{arXiv preprint arXiv:2002.04185}, 2020.

\bibitem[Cohen et~al.(2018)Cohen, Geiger, K{\"o}hler, and
  Welling]{cohen2018spherical}
Taco~S Cohen, Mario Geiger, Jonas K{\"o}hler, and Max Welling.
\newblock Spherical cnns.
\newblock \emph{arXiv preprint arXiv:1801.10130}, 2018.

\bibitem[Deshpande et~al.(2017)Deshpande, Lu, Yeh, Jin~Chong, and
  Forsyth]{mdn_color}
Aditya Deshpande, Jiajun Lu, Mao-Chuang Yeh, Min Jin~Chong, and David Forsyth.
\newblock Learning diverse image colorization.
\newblock In \emph{The IEEE Conference on Computer Vision and Pattern
  Recognition (CVPR)}, July 2017.

\bibitem[Driscoll \& Healy(1994)Driscoll and Healy]{driscoll1994computing}
James~R Driscoll and Dennis~M Healy.
\newblock Computing fourier transforms and convolutions on the 2-sphere.
\newblock \emph{Advances in applied mathematics}, 15\penalty0 (2):\penalty0
  202--250, 1994.

\bibitem[Du \& Mordatch(2019)Du and Mordatch]{NIPS2019_8619}
Yilun Du and Igor Mordatch.
\newblock Implicit generation and modeling with energy based models.
\newblock In H.~Wallach, H.~Larochelle, A.~Beygelzimer, F.~d~Alch\'{e}-Buc,
  E.~Fox, and R.~Garnett (eds.), \emph{Advances in Neural Information
  Processing Systems 32}, pp.\  3608--3618. Curran Associates, Inc., 2019.

\bibitem[Esteves et~al.(2018)Esteves, Allen-Blanchette, Makadia, and
  Daniilidis]{esteves2018learning}
Carlos Esteves, Christine Allen-Blanchette, Ameesh Makadia, and Kostas
  Daniilidis.
\newblock Learning so (3) equivariant representations with spherical cnns.
\newblock In \emph{Proceedings of the European Conference on Computer Vision
  (ECCV)}, pp.\  52--68, 2018.

\bibitem[Fetaya et~al.(2020)Fetaya, Jacobsen, Grathwohl, and
  Zemel]{fetaya2020understanding}
Ethan Fetaya, J{\"o}rn-Henrik Jacobsen, Will Grathwohl, and Richard Zemel.
\newblock Understanding the limitations of conditional generative models.
\newblock In \emph{International Conference on Learning Representations}, 2020.

\bibitem[Ghosh et~al.(2018)Ghosh, Kulharia, Namboodiri, Torr, and
  Dokania]{multiGAN}
Arnab Ghosh, Viveka Kulharia, Vinay~P. Namboodiri, Philip~H.S. Torr, and
  Puneet~K. Dokania.
\newblock Multi-agent diverse generative adversarial networks.
\newblock In \emph{The IEEE Conference on Computer Vision and Pattern
  Recognition (CVPR)}, June 2018.

\bibitem[Goodfellow et~al.(2014{\natexlab{a}})Goodfellow, Pouget-Abadie, Mirza,
  Xu, Warde-Farley, Ozair, Courville, and Bengio]{ganPaper}
Ian Goodfellow, Jean Pouget-Abadie, Mehdi Mirza, Bing Xu, David Warde-Farley,
  Sherjil Ozair, Aaron Courville, and Yoshua Bengio.
\newblock Generative adversarial nets.
\newblock In \emph{Advances in Neural Information Processing Systems 27}, pp.\
  2672--2680. 2014{\natexlab{a}}.

\bibitem[Goodfellow et~al.(2014{\natexlab{b}})Goodfellow, Pouget-Abadie, Mirza,
  Xu, Warde-Farley, Ozair, Courville, and Bengio]{goodfellow2014generative}
Ian Goodfellow, Jean Pouget-Abadie, Mehdi Mirza, Bing Xu, David Warde-Farley,
  Sherjil Ozair, Aaron Courville, and Yoshua Bengio.
\newblock Generative adversarial nets.
\newblock In \emph{Advances in neural information processing systems}, pp.\
  2672--2680, 2014{\natexlab{b}}.

\bibitem[Grathwohl et~al.(2020)Grathwohl, Wang, Jacobsen, Duvenaud, Norouzi,
  and Swersky]{secretlyEBM}
Will Grathwohl, Kuan-Chieh Wang, Joern-Henrik Jacobsen, David Duvenaud,
  Mohammad Norouzi, and Kevin Swersky.
\newblock Your classifier is secretly an energy based model and you should
  treat it like one.
\newblock In \emph{International Conference on Learning Representations}, 2020.

\bibitem[Han et~al.(2019)Han, Shang, Liu, and Zwicker]{han2019view}
Zhizhong Han, Mingyang Shang, Yu-Shen Liu, and Matthias Zwicker.
\newblock View inter-prediction gan: Unsupervised representation learning for
  3d shapes by learning global shape memories to support local view
  predictions.
\newblock In \emph{Proceedings of the AAAI Conference on Artificial
  Intelligence}, volume~33, pp.\  8376--8384, 2019.

\bibitem[He et~al.(2018)He, Schiele, and Fritz]{he2018diverse}
Yang He, Bernt Schiele, and Mario Fritz.
\newblock Diverse conditional image generation by stochastic regression with
  latent drop-out codes.
\newblock In \emph{Proceedings of the European Conference on Computer Vision
  (ECCV)}, pp.\  406--421, 2018.

\bibitem[Huang et~al.(2018)Huang, Liu, Belongie, and
  Kautz]{disentangle_multimodal}
Xun Huang, Ming-Yu Liu, Serge Belongie, and Jan Kautz.
\newblock Multimodal unsupervised image-to-image translation.
\newblock In \emph{The European Conference on Computer Vision (ECCV)},
  September 2018.

\bibitem[Iizuka et~al.(2016)Iizuka, Simo-Serra, and Ishikawa]{iizuka2016let}
Satoshi Iizuka, Edgar Simo-Serra, and Hiroshi Ishikawa.
\newblock Let there be color! joint end-to-end learning of global and local
  image priors for automatic image colorization with simultaneous
  classification.
\newblock \emph{ACM Transactions on Graphics (ToG)}, 35\penalty0 (4):\penalty0
  1--11, 2016.

\bibitem[Isola et~al.(2017)Isola, Zhu, Zhou, and Efros]{isola2017image}
Phillip Isola, Jun-Yan Zhu, Tinghui Zhou, and Alexei~A Efros.
\newblock Image-to-image translation with conditional adversarial networks.
\newblock In \emph{Proceedings of the IEEE conference on computer vision and
  pattern recognition}, pp.\  1125--1134, 2017.

\bibitem[Jing \& Tian(2020)Jing and Tian]{ssl_survey}
Longlong Jing and Yingli Tian.
\newblock Self-supervised visual feature learning with deep neural networks: A
  survey.
\newblock \emph{IEEE transactions on pattern analysis and machine
  intelligence}, 2020.

\bibitem[Khan et~al.(2019)Khan, Guo, Hayat, and Barnes]{khan2019unsupervised}
Salman~H Khan, Yulan Guo, Munawar Hayat, and Nick Barnes.
\newblock Unsupervised primitive discovery for improved 3d generative modeling.
\newblock In \emph{Proceedings of the IEEE Conference on Computer Vision and
  Pattern Recognition}, pp.\  9739--9748, 2019.

\bibitem[Kingma \& Welling(2014)Kingma and Welling]{vaePaper}
Diederik~P. Kingma and Max Welling.
\newblock Auto-encoding variational bayes.
\newblock \emph{CoRR}, abs/1312.6114, 2014.

\bibitem[Kodali et~al.(2017)Kodali, Abernethy, Hays, and
  Kira]{kodali2017convergence}
Naveen Kodali, Jacob Abernethy, James Hays, and Zsolt Kira.
\newblock On convergence and stability of gans.
\newblock \emph{arXiv preprint arXiv:1705.07215}, 2017.

\bibitem[Krizhevsky et~al.(2009)Krizhevsky, Nair, and Hinton]{cifar}
Alex Krizhevsky, Vinod Nair, and Geoffrey Hinton.
\newblock Cifar-10 (canadian institute for advanced research).
\newblock 2009.
\newblock URL \url{http://www.cs.toronto.edu/~kriz/cifar.html}.

\bibitem[Lee et~al.(2018)Lee, Tseng, Huang, Singh, and Yang]{disentangle_im2im}
Hsin-Ying Lee, Hung-Yu Tseng, Jia-Bin Huang, Maneesh Singh, and Ming-Hsuan
  Yang.
\newblock Diverse image-to-image translation via disentangled representations.
\newblock In \emph{The European Conference on Computer Vision (ECCV)},
  September 2018.

\bibitem[Lee et~al.(2019)Lee, Ha, and Kim]{harmonizing}
Soochan Lee, Junsoo Ha, and Gunhee Kim.
\newblock Harmonizing maximum likelihood with {GAN}s for multimodal conditional
  generation.
\newblock In \emph{International Conference on Learning Representations}, 2019.

\bibitem[Liu et~al.(2018)Liu, Reda, Shih, Wang, Tao, and
  Catanzaro]{liu2018image}
Guilin Liu, Fitsum~A Reda, Kevin~J Shih, Ting-Chun Wang, Andrew Tao, and Bryan
  Catanzaro.
\newblock Image inpainting for irregular holes using partial convolutions.
\newblock In \emph{Proceedings of the European Conference on Computer Vision
  (ECCV)}, pp.\  85--100, 2018.

\bibitem[Loquercio et~al.(2019)Loquercio, Seg{\`u}, and
  Scaramuzza]{loquercio2019general}
Antonio Loquercio, Mattia Seg{\`u}, and Davide Scaramuzza.
\newblock A general framework for uncertainty estimation in deep learning.
\newblock \emph{arXiv preprint arXiv:1907.06890}, 2019.

\bibitem[Maal{\o}e et~al.(2019)Maal{\o}e, Fraccaro, Li{\'e}vin, and
  Winther]{maaloe2019biva}
Lars Maal{\o}e, Marco Fraccaro, Valentin Li{\'e}vin, and Ole Winther.
\newblock Biva: A very deep hierarchy of latent variables for generative
  modeling.
\newblock In \emph{Advances in neural information processing systems}, pp.\
  6548--6558, 2019.

\bibitem[Mao et~al.(2019)Mao, Lee, Tseng, Ma, and Yang]{mao2019mode}
Qi~Mao, Hsin-Ying Lee, Hung-Yu Tseng, Siwei Ma, and Ming-Hsuan Yang.
\newblock Mode seeking generative adversarial networks for diverse image
  synthesis.
\newblock In \emph{Proceedings of the IEEE Conference on Computer Vision and
  Pattern Recognition}, pp.\  1429--1437, 2019.

\bibitem[Mathieu et~al.(2015)Mathieu, Couprie, and LeCun]{DeepMV}
Micha{\"e}l Mathieu, Camille Couprie, and Yann LeCun.
\newblock Deep multi-scale video prediction beyond mean square error.
\newblock \emph{CoRR}, abs/1511.05440, 2015.

\bibitem[Mirza \& Osindero(2014)Mirza and Osindero]{cGAN}
Mehdi Mirza and Simon Osindero.
\newblock Conditional generative adversarial nets.
\newblock \emph{ArXiv}, abs/1411.1784, 2014.

\bibitem[Nalisnick et~al.(2019)Nalisnick, Matsukawa, Teh, Gorur, and
  Lakshminarayanan]{nalisnick2019deep}
Eric Nalisnick, Akihiro Matsukawa, Yee~Whye Teh, Dilan Gorur, and Balaji
  Lakshminarayanan.
\newblock Do deep generative models know what they don't know?
\newblock \emph{International Conference on Learning Representations}, 2019.

\bibitem[Parkhi et~al.(2012)Parkhi, Vedaldi, Zisserman, and Jawahar]{parkhi12a}
Omkar~M. Parkhi, Andrea Vedaldi, Andrew Zisserman, and C.~V. Jawahar.
\newblock Cats and dogs.
\newblock In \emph{IEEE Conference on Computer Vision and Pattern Recognition},
  2012.

\bibitem[Pathak et~al.(2016)Pathak, Kr\"ahenb\"uhl, Donahue, Darrell, and
  Efros]{contextEncoder}
Deepak Pathak, Philipp Kr\"ahenb\"uhl, Jeff Donahue, Trevor Darrell, and Alexei
  Efros.
\newblock Context encoders: Feature learning by inpainting.
\newblock 2016.

\bibitem[Perraudin et~al.(2019)Perraudin, Defferrard, Kacprzak, and
  Sgier]{perraudin2019deepsphere}
Nathana{\"e}l Perraudin, Micha{\"e}l Defferrard, Tomasz Kacprzak, and Raphael
  Sgier.
\newblock Deepsphere: Efficient spherical convolutional neural network with
  healpix sampling for cosmological applications.
\newblock \emph{Astronomy and Computing}, 27:\penalty0 130--146, 2019.

\bibitem[Prashnani et~al.(2018)Prashnani, Cai, Mostofi, and
  Sen]{prashnani2018pieapp}
Ekta Prashnani, Hong Cai, Yasamin Mostofi, and Pradeep Sen.
\newblock Pieapp: Perceptual image-error assessment through pairwise
  preference.
\newblock In \emph{Proceedings of the IEEE Conference on Computer Vision and
  Pattern Recognition}, pp.\  1808--1817, 2018.

\bibitem[Ramasinghe et~al.(2019{\natexlab{a}})Ramasinghe, Khan, and
  Barnes]{ramasinghe2019volumetric}
Sameera Ramasinghe, Salman Khan, and Nick Barnes.
\newblock Volumetric convolution: Automatic representation learning in unit
  ball.
\newblock \emph{arXiv preprint arXiv:1901.00616}, 2019{\natexlab{a}}.

\bibitem[Ramasinghe et~al.(2019{\natexlab{b}})Ramasinghe, Khan, Barnes, and
  Gould]{ramasinghe2019representation}
Sameera Ramasinghe, Salman Khan, Nick Barnes, and Stephen Gould.
\newblock Representation learning on unit ball with 3d roto-translational
  equivariance.
\newblock \emph{International Journal of Computer Vision}, pp.\  1--23,
  2019{\natexlab{b}}.

\bibitem[Ramasinghe et~al.(2019{\natexlab{c}})Ramasinghe, Khan, Barnes, and
  Gould]{ramasinghe2019spectral}
Sameera Ramasinghe, Salman Khan, Nick Barnes, and Stephen Gould.
\newblock Spectral-gans for high-resolution 3d point-cloud generation.
\newblock \emph{arXiv preprint arXiv:1912.01800}, 2019{\natexlab{c}}.

\bibitem[Ramasinghe et~al.(2020)Ramasinghe, Khan, Barnes, and
  Gould]{ramasinghe2020blended}
Sameera Ramasinghe, Salman Khan, Nick Barnes, and Stephen Gould.
\newblock Blended convolution and synthesis for efficient discrimination of 3d
  shapes.
\newblock In \emph{The IEEE Winter Conference on Applications of Computer
  Vision}, pp.\  21--31, 2020.

\bibitem[Robbins(2007)]{Robbins2007ASA}
Herbert~E. Robbins.
\newblock A stochastic approximation method.
\newblock \emph{Annals of Mathematical Statistics}, 22:\penalty0 400--407,
  2007.

\bibitem[Sagong et~al.(2019)Sagong, Shin, Kim, Park, and Ko]{inpainting_03}
Min-cheol Sagong, Yong-goo Shin, Seung-wook Kim, Seung Park, and Sung-jea Ko.
\newblock Pepsi : Fast image inpainting with parallel decoding network.
\newblock In \emph{The IEEE Conference on Computer Vision and Pattern
  Recognition (CVPR)}, June 2019.

\bibitem[Salimans et~al.(2016)Salimans, Goodfellow, Zaremba, Cheung, Radford,
  and Chen]{salimans2016improved}
Tim Salimans, Ian Goodfellow, Wojciech Zaremba, Vicki Cheung, Alec Radford, and
  Xi~Chen.
\newblock Improved techniques for training gans.
\newblock In \emph{Advances in neural information processing systems}, pp.\
  2234--2242, 2016.

\bibitem[Sauder \& Sievers(2019)Sauder and Sievers]{sauder2019self}
Jonathan Sauder and Bjarne Sievers.
\newblock Self-supervised deep learning on point clouds by reconstructing
  space.
\newblock In \emph{Advances in Neural Information Processing Systems}, pp.\
  12942--12952, 2019.

\bibitem[Sharma et~al.(2016)Sharma, Grau, and Fritz]{sharma2016vconv}
Abhishek Sharma, Oliver Grau, and Mario Fritz.
\newblock Vconv-dae: Deep volumetric shape learning without object labels.
\newblock In \emph{European Conference on Computer Vision}, pp.\  236--250.
  Springer, 2016.

\bibitem[Sohn et~al.(2015)Sohn, Lee, and Yan]{cVAE}
Kihyuk Sohn, Honglak Lee, and Xinchen Yan.
\newblock Learning structured output representation using deep conditional
  generative models.
\newblock In \emph{Advances in Neural Information Processing Systems 28}. 2015.

\bibitem[Thanh-Tung et~al.(2019)Thanh-Tung, Tran, and
  Venkatesh]{thanh2019improving}
Hoang Thanh-Tung, Truyen Tran, and Svetha Venkatesh.
\newblock Improving generalization and stability of generative adversarial
  networks.
\newblock \emph{arXiv preprint arXiv:1902.03984}, 2019.

\bibitem[Tyle{\v c}ek \& {\v S}{\' a}ra(2013)Tyle{\v c}ek and {\v S}{\'
  a}ra]{facade_dataset}
Radim Tyle{\v c}ek and Radim {\v S}{\' a}ra.
\newblock Spatial pattern templates for recognition of objects with regular
  structure.
\newblock In \emph{Proc. GCPR}, Saarbrucken, Germany, 2013.

\bibitem[Vitoria et~al.(2020)Vitoria, Raad, and
  Ballester]{vitoria2020chromagan}
Patricia Vitoria, Lara Raad, and Coloma Ballester.
\newblock Chromagan: Adversarial picture colorization with semantic class
  distribution.
\newblock In \emph{The IEEE Winter Conference on Applications of Computer
  Vision}, pp.\  2445--2454, 2020.

\bibitem[Wang et~al.(2018)Wang, Tao, Qi, Shen, and Jia]{inpainting_nips}
Yi~Wang, Xin Tao, Xiaojuan Qi, Xiaoyong Shen, and Jiaya Jia.
\newblock Image inpainting via generative multi-column convolutional neural
  networks.
\newblock In \emph{Advances in Neural Information Processing Systems 31}. 2018.

\bibitem[Wang et~al.(2003)Wang, Simoncelli, and Bovik]{wang2003multiscale}
Zhou Wang, Eero~P Simoncelli, and Alan~C Bovik.
\newblock Multiscale structural similarity for image quality assessment.
\newblock In \emph{The Thrity-Seventh Asilomar Conference on Signals, Systems
  \& Computers, 2003}, volume~2, pp.\  1398--1402. Ieee, 2003.

\bibitem[Wang et~al.(2004)Wang, Bovik, Sheikh, and Simoncelli]{wang2004image}
Zhou Wang, Alan~C Bovik, Hamid~R Sheikh, and Eero~P Simoncelli.
\newblock Image quality assessment: from error visibility to structural
  similarity.
\newblock \emph{IEEE transactions on image processing}, 13\penalty0
  (4):\penalty0 600--612, 2004.

\bibitem[Xie et~al.(2018)Xie, Franz, Chu, and Thuerey]{tempoGAN}
You Xie, Erik Franz, Mengyu Chu, and Nils Thuerey.
\newblock {tempoGAN: A Temporally Coherent, Volumetric GAN for Super-resolution
  Fluid Flow}.
\newblock \emph{ACM Transactions on Graphics (TOG)}, 37\penalty0 (4):\penalty0
  95, 2018.

\bibitem[Yang et~al.(2019)Yang, Hong, Jang, Zhao, and Lee]{yang2019diversity}
Dingdong Yang, Seunghoon Hong, Yunseok Jang, Tianchen Zhao, and Honglak Lee.
\newblock Diversity-sensitive conditional generative adversarial networks.
\newblock \emph{arXiv preprint arXiv:1901.09024}, 2019.

\bibitem[Yang et~al.(2018)Yang, Feng, Shen, and Tian]{yang2018foldingnet}
Yaoqing Yang, Chen Feng, Yiru Shen, and Dong Tian.
\newblock Foldingnet: Point cloud auto-encoder via deep grid deformation.
\newblock In \emph{Proceedings of the IEEE Conference on Computer Vision and
  Pattern Recognition}, pp.\  206--215, 2018.

\bibitem[Yu et~al.(2018{\natexlab{a}})Yu, Lin, Yang, Shen, Lu, and
  Huang]{inpainting_02}
Jiahui Yu, Zhe Lin, Jimei Yang, Xiaohui Shen, Xin Lu, and Thomas~S. Huang.
\newblock Generative image inpainting with contextual attention.
\newblock In \emph{The IEEE Conference on Computer Vision and Pattern
  Recognition (CVPR)}, June 2018{\natexlab{a}}.

\bibitem[Yu et~al.(2018{\natexlab{b}})Yu, Lin, Yang, Shen, Lu, and
  Huang]{yu2018generative}
Jiahui Yu, Zhe Lin, Jimei Yang, Xiaohui Shen, Xin Lu, and Thomas~S Huang.
\newblock Generative image inpainting with contextual attention.
\newblock In \emph{Proceedings of the IEEE conference on computer vision and
  pattern recognition}, pp.\  5505--5514, 2018{\natexlab{b}}.

\bibitem[Zeng et~al.(2019)Zeng, Fu, Chao, and Guo]{pennet}
Yanhong Zeng, Jianlong Fu, Hongyang Chao, and Baining Guo.
\newblock Learning pyramid-context encoder network for high-quality image
  inpainting.
\newblock In \emph{The IEEE Conference on Computer Vision and Pattern
  Recognition (CVPR)}, pp.\  1486--1494, 2019.

\bibitem[Zhang et~al.(2011)Zhang, Zhang, Mou, and Zhang]{zhang2011fsim}
Lin Zhang, Lei Zhang, Xuanqin Mou, and David Zhang.
\newblock Fsim: A feature similarity index for image quality assessment.
\newblock \emph{IEEE transactions on Image Processing}, 20\penalty0
  (8):\penalty0 2378--2386, 2011.

\bibitem[Zhang et~al.(2016)Zhang, Isola, and Efros]{zhang2016colorful}
Richard Zhang, Phillip Isola, and Alexei~A Efros.
\newblock Colorful image colorization.
\newblock In \emph{European conference on computer vision}, pp.\  649--666.
  Springer, 2016.

\bibitem[Zhang et~al.(2018)Zhang, Isola, Efros, Shechtman, and
  Wang]{zhang2018perceptual}
Richard Zhang, Phillip Isola, Alexei~A Efros, Eli Shechtman, and Oliver Wang.
\newblock The unreasonable effectiveness of deep features as a perceptual
  metric.
\newblock In \emph{CVPR}, 2018.

\bibitem[Zhang \& Qi(2017)Zhang and Qi]{zhifei2017cvpr}
Song-Yang Zhang, Zhifei and Hairong Qi.
\newblock Age progression/regression by conditional adversarial autoencoder.
\newblock In \emph{IEEE Conference on Computer Vision and Pattern Recognition
  (CVPR)}. IEEE, 2017.

\bibitem[Zhang et~al.(2020)Zhang, Wang, Kong, Yin, Song, Lyu, Lv, Shi, and
  Li]{inpainting_01}
Xian Zhang, Xin Wang, Bin Kong, Youbing Yin, Qi~Song, Siwei Lyu, Jiancheng Lv,
  Canghong Shi, and Xiaojie Li.
\newblock Domain embedded multi-model generative adversarial networks for
  image-based face inpainting.
\newblock \emph{ArXiv}, abs/2002.02909, 2020.

\bibitem[Zhu et~al.(2016)Zhu, Kr{\"a}henb{\"u}hl, Shechtman, and Efros]{iGAN}
Jun-Yan Zhu, Philipp Kr{\"a}henb{\"u}hl, Eli Shechtman, and Alexei~A. Efros.
\newblock Generative visual manipulation on the natural image manifold.
\newblock In \emph{Proceedings of European Conference on Computer Vision
  (ECCV)}, 2016.

\bibitem[Zhu et~al.(2017{\natexlab{a}})Zhu, Zhang, Pathak, Darrell, Efros,
  Wang, and Shechtman]{bicycleGAN}
Jun-Yan Zhu, Richard Zhang, Deepak Pathak, Trevor Darrell, Alexei~A Efros,
  Oliver Wang, and Eli Shechtman.
\newblock Toward multimodal image-to-image translation.
\newblock In \emph{Advances in Neural Information Processing Systems 30}.
  2017{\natexlab{a}}.

\bibitem[Zhu et~al.(2017{\natexlab{b}})Zhu, Zhang, Pathak, Darrell, Efros,
  Wang, and Shechtman]{zhu2017toward}
Jun-Yan Zhu, Richard Zhang, Deepak Pathak, Trevor Darrell, Alexei~A Efros,
  Oliver Wang, and Eli Shechtman.
\newblock Toward multimodal image-to-image translation.
\newblock In \emph{Advances in neural information processing systems}, pp.\
  465--476, 2017{\natexlab{b}}.

\end{thebibliography}

\newpage
\section*{Appendix}
\setcounter{section}{0}

 \section{Related work}

\textbf{Conditional Generative Modeling.}
Conditional generation involves modeling the data distribution given a set of conditioning variables that control of modes of the generated samples. 
With the success of \textit{VAEs} \citep{vaePaper} and \textit{GANs} \citep{ganPaper} in standard generative modeling tasks, their conditioned counterparts \citep{cVAE, cGAN} have dominated conditional generative tasks recently \citep{vitoria2020chromagan, zhang2016colorful, isola2017image, contextEncoder, harmonizing, bicycleGAN, cvaegan, disentangle_im2im, pennet}.
While probabilistic latent variable models such as VAEs generate relatively low quality samples and poor likelihood estimates at inference \citep{maaloe2019biva}, GAN based models 
perform significantly better at high dimensional distributions like natural images but demonstrate unstable training behaviour. A distinct feature of GANs is its mapping of points from a random noise distribution to the various modes of the output distribution. However, in the conditional case where an additional loss is incorporated to enforce the conditioning on the input, the significantly better performance of GANs is achieved at the expense of multimodality; the conditioning loss pushes the GAN to learn to mostly ignore its noise distribution. In fact, some works intentionally ignore the noise input in order to achieve more stable training \citep{isola2017image, contextEncoder, DeepMV, tempoGAN}. 

\textbf{Multimodality.} \textit{Conditional VAE-GANs} are one popular approach for generating multimodal outputs \citep{cvaegan, bicycleGAN} using the VAE's ability to enforce diversity through its latent variable representation and the GAN's ability to enforce output fidelity through its learnt discrimanator model. \textit{Mixture models} \citep{min_loss, multiGAN, mdn_color} that discretize the output space are another approach. Domain specific \textit{disentangled representations} \citep{disentangle_im2im, disentangle_multimodal} and explicit encoding of multiple modes as inputs \cite{iGAN, isola2017image} have also been successful in generating diverse outputs. \textit{Sampling-based loss} functions enforcing similarity at a distribution level \citep{harmonizing} have also been successful in multimoal generative tasks. Further, the use of additional \textit{specialized reconstruction losses} (often using higher-level features extracted from the data distribution) and attention mechanisms also achieves multimodality through intricate model architectures in domain specific cases \citep{pennet, min_loss, vitoria2020chromagan, zhang2016colorful, iizuka2016let, inpainting_01, inpainting_02, inpainting_03, inpainting_nips, iizuka2016let}. 

We propose a simpler direction through our domain-independent energy function based approach that is also capable of learning generic representations that better support downstream tasks. Notably, our work contrasts from energy based models previously investigated for likelihood modeling due to their simplicity, however, such models are notoriously difficult to train especially on high-dimensional spaces \citep{NIPS2019_8619}.

 \section{Theoretical results}
 \label{app:propositions}
 
 \subsection{Proof for Eq.~\ref{equ:bound}}
 \label{app:gradnorm}
 \begin{equation}
    ||\mathcal{G}(x_j,w^*,z_{i,j}^*) - \mathcal{G}(x_j,w^*,z_0)|| = ||\int_{z_0}^{z_{i,j}^*} \nabla_z \mathcal{G}(x_j,w^*,z)dz||
\end{equation}

Let $\gamma(t)$ be a straight path from $z_0$ to $z^*_{i,j}$, where $\gamma(0) = z_0$ and $\gamma(1) = z_{i,j}^*$. Then,

\begin{equation}
     = ||\int_0^1 \nabla_z \mathcal{G}(x_j,w^*,\gamma (t))\frac{d\gamma}{dt}dt||
\end{equation}
\begin{equation}
    = ||\int_0^1 \nabla_z \mathcal{G}(x_j,w^*,\gamma (t))(z^*_{i,j} - z^*_0)dt||
\end{equation}
\begin{equation}
    = ||(z^*_{i,j} - z^*_0)\int_0^1 \nabla_z \mathcal{G}(x_j,w^*,\gamma (t))dt||
\end{equation}
\begin{equation}
\label{eq:norm1}
    \leq \norm{(z^*_{i,j} - z^*_0)}\norm{\int_0^1 \nabla_z \mathcal{G}(x_j,w^*,\gamma (t))dt}
\end{equation}

On the other hand the Lipschitz constraint ensures,

\begin{equation}
\label{eq:norm2}
  \norm{\nabla_z \mathcal{G}(x_j,w^*,\gamma (t))} \leq \lim_{\epsilon \to  0} \frac{\norm{\mathcal{G}(x_j,w^*,\gamma (t)) - \mathcal{G}(x_j,w^*,\gamma (t+\epsilon))}}{\norm{z_t - z_{t+\epsilon}}} \leq C,
\end{equation}

where $C$ is a constant. Combining Eq.~\ref{eq:norm1} and \ref{eq:norm2} we get,

\begin{equation}
    \frac{\norm{\mathcal{G}(x_j,w^*,z_{i,j}^*) - \mathcal{G}(x_j,w^*,z_0)}}{\norm{z_{i,j} ^* - z_{0}}} \leq \int_0^1 \norm{\nabla_z \mathcal{G}(x_j,w^*,\gamma (t))}dt \leq C.
\end{equation}

  \subsection{Convergence of the training algorithm.}
  \label{app:convergence}


\noindent \textit{Proof: } Let us consider a particular input $x_j$ and an associated ground truth $y^{g}_{i,j}$. Then, for this particular case, we denote our cost function to be $\hat{E}_{i,j} = d(w,z)$. Further,  a family of cost functions can be defined as,
\begin{equation}
    f_{w}(z) = d(w,z),
\end{equation}
for each $w \sim \omega$. Further, let us consider an arbitrary initial setting ($z_{init},w_{init}$). Then, with enough iterations, gradient descent by $\nabla_{z}f_{w}(z)$ converges $z_{init}$ to,
\begin{equation}
\label{zopt}
    \bar{z} = {\arg}\inf_{z \in \zeta} f_w.
\end{equation}
Next, with enough iterations, gradient descent by $\nabla_{w}f_{w}(\bar{z})$ converges $w$ to,
\begin{equation}
\label{wopt}
    \bar{w} = {\arg}\inf_{w \in \omega} f_w(\bar{z}).
\end{equation}

Observe that $f_{\bar{w}}(\bar{z}) \leq f_{w_{init}}$, where the equality occurs when $\nabla_{z}f_{w}(z) = \nabla_{w}f_{w}(\bar{z}) = 0$. If $f_w(z)$ has a unique global minima, repeating Equation \ref{zopt} and \ref{wopt} converges to that global minima, giving $\{z^*_{i,j}, w^*_{i,j}\}$. It is straight forward to see that using a small number of iterations (usually one in our case) for each sample set for Equation \ref{wopt}, i.e., stochastic gradient descent, gives us,
\begin{equation}
    \{z_{i,j}^*, w^*\}= \underaccent{z_{i,j} \in \zeta , w \in \omega}{\arg\min} \mathbb{E}_{i \in I,j\in J}[\hat{E}_{i,j}],
\end{equation}
 where $w^*$ is fixed for all samples and modes \citep{Robbins2007ASA}. Note that the proof is valid only for the convex case, and we rely on stochastic gradient descent to converge to at least a good local minima, as commonly done in many deep learning settings.

 \subsection{Proof for Remark}
 \label{app:remark}
 \paragraph{Remark:} \textit{Consider a generator ${G}(x,z)$ and a discriminator $D(x,z)$ with a finite capacity, where $x$ and $z$ are input and the noise vector, respectively. Then, consider an arbitrary input $x_j$ and the corresponding set of ground truths $\{y^g_{i,j}\}, i=1,2,..N$. Further, let us define the optimal generator $ G^*(x_j, z) = \hat{y}, \hat{y} \in \{y^g_{i,j}\}$, $L_{GAN} = \mathbb{E}_{i} [\log D(y^g_{i,j})] + \mathbb{E}_{z}[\log(1-D(G(x_j,z))]$ and  $L_{\ell} = \mathbb{E}_{i,z}[|y^g_{i,j} - G(x_j,z)|]$. Then, $G^* \neq \hat{G}^*$ where $\hat{G}^* = \arg\underaccent{G}{\min}\underaccent{D}{\max} L_{GAN} + \lambda L_{\ell}$,  $\forall \lambda \neq 0$.} 
 \textit{Proof.}
 
 It is straightforward to derive the equilibrium point of  $\arg\underaccent{G}{\min}\underaccent{D}{\max} L_{GAN}$ from the original GAN formulation. However, for clarity, we show some steps here.
 
 Let,
 \begin{equation}
     V(G,D) = \arg\underaccent{G}{\min}\underaccent{D}{\max}  \mathbb{E}_{i} [\log D(y^g_{i,j})] + \mathbb{E}_{z}[\log(1-D(G(x_j,z))] 
 \end{equation}
 
 Let $p(\cdot)$ denote the probability distribution. Then,
 \begin{equation}
     V(G,D) = \arg\underaccent{G}{\min}\underaccent{D}{\max} \int_{\Upsilon} p(y^g_{\cdot,j})\log D(y^g_{\cdot,j}) + p(\bar{y}_{\cdot,j})(\log(1-D(G(x_j,z))dy
 \end{equation}
 
  \begin{equation}
     V(G,D) = \arg\underaccent{G}{\min}\underaccent{D}{\max} \mathbb{E}_{y \sim y^g_{\cdot,j}} [ \log D(y)] + \mathbb{E}_{ y \sim \bar{y}_{\cdot,j}} [\log(1-D(y)]
 \end{equation}
 
 Consider the inner loop. It is straightforward to see that $V(G,D)$ is maximized w.r.t. $D$ when $D(y) = \frac{ p(y^g_{\cdot,j})}{p(y^g_{\cdot,j}) + p(\bar{y}_{\cdot,j})}$. Then, 
 
 \begin{equation}
   C(G) =  V(G,D) = \arg\underaccent{G}{\min}  \mathbb{E}_{y \sim y^g_{\cdot,j}}[\log \frac{ p(y^g_{\cdot,j})}{p(y^g_{\cdot,j}) + p(\bar{y}_{\cdot,j})}] +  \mathbb{E}_{y \sim \bar{y}_{\cdot,j}}[\log \frac{ p(\bar{y}_{\cdot,j})}{p(y^g_{\cdot,j}) + p(\bar{y}_{\cdot,j})} ]
 \end{equation}
 
 Then, following the  \textbf{Theorem 1} from \citet{goodfellow2014generative}, it can be shown that the global minimum of the virtual training criterion $C(G)$ is achieved if and only if $p(y^g_{\cdot,j}) = p(\bar{y}_{\cdot,j})$. 
  
  


Next, consider the $L_1$ loss for $x_j$,

\begin{equation}
    L_1 = \frac{1}{N} \sum_{i} \abs{y^g_{i,j} - G(x_j, z, w)} 
\end{equation}
\begin{equation}
    \nabla_{w}L_1  = -\frac{1}{N} \sum_{i} \text{sgn}(y^g_{i,j} - G(x_j, z, w))\nabla_{w}(G(x_j, z, w))
\end{equation}

For $L_1$ to approach to a minima, $\nabla_{w}L_1 \rightarrow 0$. Since $\{y^g_{i,j}\}$ is not a singleton, when $L_1 \rightarrow 0$, $G(x_j, z, w) \neq \hat{y} \in \{y^g_{i,j}\}$.

Now, let us consider the $L_2$ loss,
\begin{equation}
    L_2 = \frac{1}{N} \sum_{i} \norm{y^g_{i,j} - G(x_j, z, w)}^2 
\end{equation}
\begin{equation}
    \nabla_{w}L_2  = -\frac{2}{N} \sum_{i}(y^g_{i,j} - G(x_j, z, w))\nabla_{w}(G(x_j, z, w))
\end{equation}

For $\nabla_{w}L_2 \rightarrow 0$, $G(x_j, z, w) \rightarrow \frac{1}{N} \sum_{i}y^g_{i,j}.$ However, omitting the very specific case where $(\frac{1}{N} \sum_{i}y^g_{i,j}) \in \{y^g_{i,j}\}$, which is highly unlikely in a complex distribution, as $L_2 \rightarrow 0$, $G(x_j, z, w) \neq \hat{y} \in \{y^g_{i,j}\}$. Therefore, the goals of $\arg\underaccent{G}{\min}\underaccent{D}{\max} L_{GAN}$ and $\lambda L_{\ell}$ are contradictory and $G^* \neq \hat{G}^*$. Note that we do not extend our proof to high order $L$ losses as it is intuitive.

\subsection{Lipschitz continuity and structuring of the latent space}
\label{app:lipschitz}

\begin{figure}
\centering
\captionsetup{size=small}
\begin{subfigure}{0.15\textwidth}
\includegraphics[width=\linewidth]{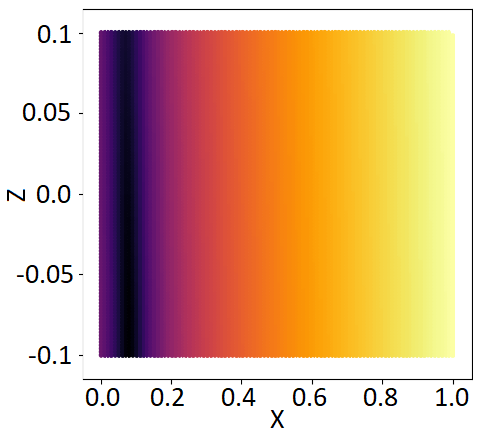} 
\end{subfigure} 
\begin{subfigure}{0.15\textwidth}
\includegraphics[width=\linewidth]{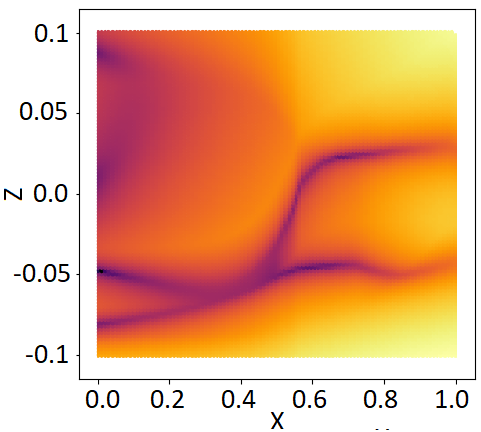}
\end{subfigure}
\begin{subfigure}{0.15\textwidth}
\includegraphics[width=\linewidth]{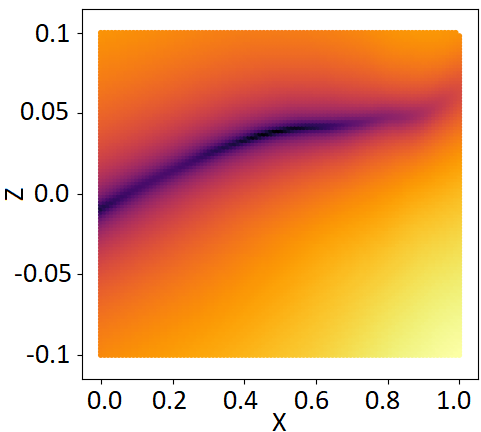}
\end{subfigure}

\begin{subfigure}{0.15\textwidth}
\includegraphics[width=\linewidth]{figures/toy/energy_surface3.png} 
\caption{0 epochs}
\end{subfigure} 
\begin{subfigure}{0.15\textwidth}
\includegraphics[width=\linewidth]{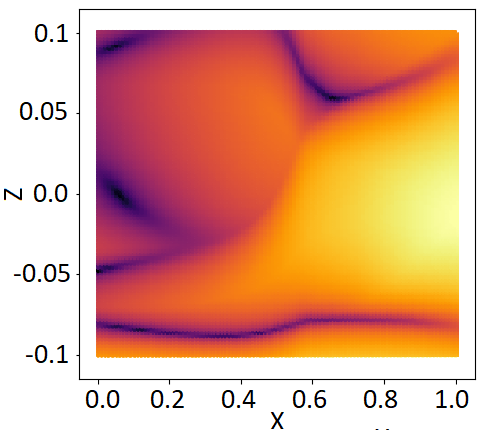}
\caption{5 epochs}
\end{subfigure}
\begin{subfigure}{0.15\textwidth}
\includegraphics[width=\linewidth]{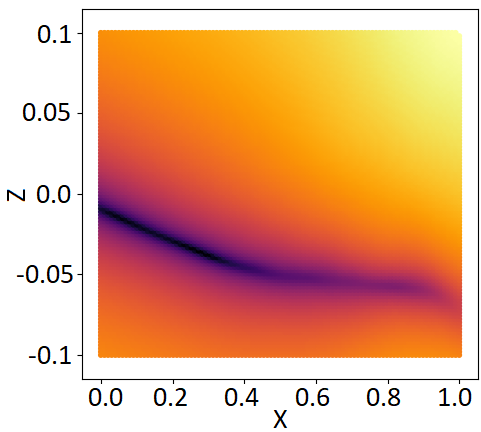}
\caption{10 epochs}
\end{subfigure}
\caption{The behaviour of cost heatmaps $\hat{E}$ against $(x,z)$ as the training progresses (toy example). The latent space gets increasingly structured as $w \rightarrow{w^*}$. Also, in (c) the network intelligently puts the optimal latent codes further apart as the distance between the two  ground truth modes ($m=4$ and $m=-4$) keeps increasing.} \label{fig:toy_energy_plot}
\end{figure}

Enforcing the Lipschitz constraint encourages meaningful structuring of the latent space: suppose $z^*_{1,j}$ and $z^*_{2,j}$ are two optimal codes corresponding to two ground truth modes for a particular input. Since $\norm{z^*_{2,j} - z^*_{1,j}}$ is lower bounded by $\frac{\norm{\mathcal{G}(x_j,w^*,z^*_{2,j}) - \mathcal{G}(x_j,w^*,z^*_{1,j})}}{L}$, where $L$ is the Lipschitz constant, the minimum distance between the two latent codes is proportional to the difference between the corresponding ground truth modes. Also, in practice, we observed that this encourages the optimum latent codes to be placed sparsely. Fig.~\ref{fig:toy_energy_plot} illustrates a visualization from the toy example.  As the training progresses, the optimal $\{z^*\}$ corresponding to minimas of $\hat{E}$ are identified and placed sparsely. Note that as expected, at the $10^{th}$ epoch the distance between the two optimum $z^*$ increases as $x$ goes from $0$ to $1$, in other words, as the $\norm{4(x, x^2, x^3) - (-4(x, x^2, x^3))}$ increases.

\begin{figure}
\centering
\includegraphics[width=0.8\linewidth]{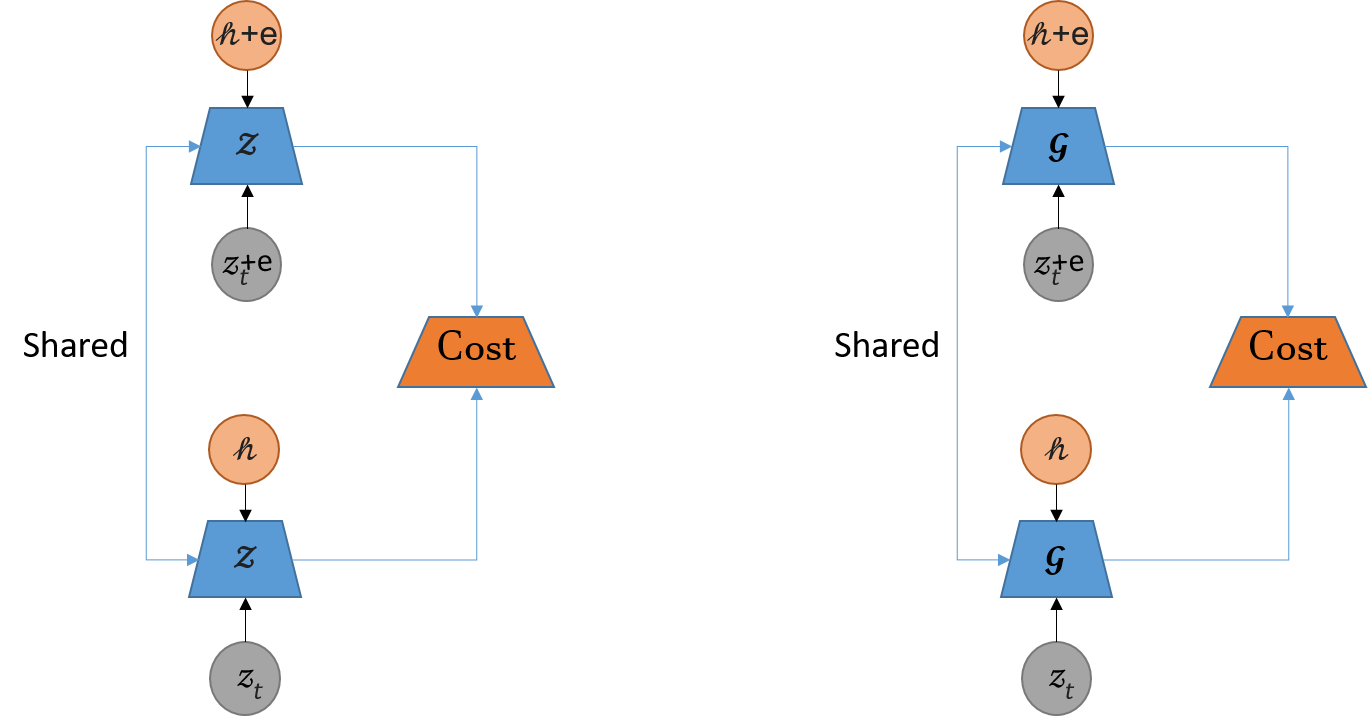} 
\caption{We enforce the Lipschitz continuity on both $\mathcal{G}$ and $\mathcal{Z}$.}
\label{app:fig_lip}
\end{figure}

Practical implementation is done as follows: during the training phase, a small noise $e$ is injected to the inputs of $\mathcal{Z}$ and $\mathcal{G}$, and the networks are penalized for any difference in output. More formally, $L_{\mathcal{Z}}$  and $\hat{E}$ now become, $L_1 [ z_{t+1},\mathcal{Z}(z_{t}, h) ] + \alpha L_1 [ \mathcal{Z}(z_{t}+e, h+e),\mathcal{Z}(z_{t}, h) ]$ and $L_1 [y^g,\mathcal{G}(h,z)] + \alpha L_1 [\mathcal{G}(h+e,z+e),\mathcal{G}(h,z) ]$, respectively. Fig.~\ref{app:fig_lip} illustrates the procedure.

 \subsection{Towards a measurement of uncertainty}

 In Bayesian approaches, the uncertainty is represented using the distribution of the network parameters $\omega$. Since a network output is unique for fixed $\bar{w} \sim \omega$, sampling from the output is equivalent to sampling from $\omega$. Often, $\omega$ is modeled as a parametric distribution or obtained through sampling, and at inference, the model uncertainty can be estimated as $\mathbb{VAR}_{p(y|x)}(y)$. One intuition behind this is that for more confident inputs, $p(y|x,w)$ will showcase less variance over the distribution of $\omega$---hence lower $\mathbb{VAR}_{p(y|x)}(y)$---as the network parameters have learned redundant information \citep{loquercio2019general}.

As opposed to sampling from the distribution of network parameters, we model the optimal $z^*$ for a particular input as a probability distribution $p(z^*)$, and measure $\mathbb{VAR}_{p(y|x)}(y)$ where  $p(y|x) = \int p(y|x,z^*)p(z^*|x)dz$. Our intuition is that in the vicinity of well observed data $\mathbb{VAR}_{p(y|x)}(y)$ is lower, since for training data 1) we enforce the Lipschitz constraint on $\mathcal{G}(x,z)$ over $(x,z)$ and 2) $\hat{E}(y^g,\mathcal{G}(x,z))$ resides in a relatively stable local minima against $z^*$ for observed data, as in practice, $z^* = \mathbb{E}_{epochs}[z^*] + \epsilon$ for a given $x$, where $\epsilon$ is some random noise which is susceptible to change over each epoch. Further, Let $(x, z^*)$ and $y^g$ be the inputs to a network $\mathcal{G}$ and the corresponding ground truth label, respectively. 

Formally, let $p(y^g|x,z^*) = \mathcal{N}(y^g;\mathcal{G}(x,z^*), \alpha \mathbb{I})$ and $z^* \sim \mathcal{U}(|z^*-\mathbb{E}(z^*)| < \delta)$, where $\alpha$ is some variable  describing the noise in the input $x$ and $\delta$ is a small positive scalar. Then,

\begin{equation}
\label{uncertainty}
   \mathbb{COV}_{p(y^g|x)}(y^g) \approx =\frac{1}{K}\sum_{k=1}^{K} [\alpha_k \mathbb{I}] + \overline{\mathbb{COV}}({\mathcal{G}( x,z^*)}).
\end{equation}

where $\overline{\mathbb{COV}}$ is the sample covariance.

\textit{proof:} $\mathbb{E}_{p(y^g|x)}(y^{g})  = \int y^gp(y^g|x)dy^g$.

= $\int y^g [ \int \mathcal{N}(y^g;\mathcal{G}(x,z^*), \alpha \mathbb{I})p(z^*|x)dz^*]dy^g $ 

= $\int  [\int y^g \mathcal{N}(y^g;\mathcal{G}(x,z^*), \alpha \mathbb{I})p(z^*|x)dy^g]dz^* $ 

= $\int  [\int y^g \mathcal{N}(y^g;\mathcal{G}(x,z^*), \alpha \mathbb{I})dy^g]p(z^*|x)dz^* $ 

= $\int \mathcal{G}(x,z^*)p(z^*|x)dz^*$ 

Let $\pi \delta^2 = A$, and $p(z^*|x) \approx \frac{1}{A}$. Then, by Monte-Carlo approximation,

$\approx \frac{1}{K} \sum_{k=1}^{K}\mathcal{G}(x, z^*_k)$

Next, consider,

$\mathbb{COV}_{p(y^g|x)}(y^g) = \mathbb{E}_{p(y^g|x)}((y^g)(y^g)^T) - \mathbb{E}_{p(y^g|x)}(y^{g})\mathbb{E}_{p(y^g|x)}(y^{g})^T$

$= \int \int (y^g) (y^g)^Tp(y^g|x,z^*)p(z^*|x)dz^*dy^g - \mathbb{E}_{p(y^g|x)}(y^{g})\mathbb{E}_{p(y^g|x)}(y^{g})^T$

$= \int [\mathbb{COV}_{p(y^g|x,z^*)} + \mathbb{E}_{p(y^g|x,z^*)}\mathbb{E}_{p(y^g|x,z^*)}^Tp(z^*|x)dz - \mathbb{E}_{p(y^g|x)}(y^{g})\mathbb{E}_{p(y^g|x)}(y^{g})^T$

$\approx \frac{1}{K}\sum_{k=1}^{K} [\alpha_k \mathbb{I} + G(x,z^*_k)G(x,z^*_k)^T] - \frac{1}{K^2} [(\sum_{k=1}^{K}G(x,z^*_k))(\sum_{k=1}^{K}G(x,z^*_k))^T]$.

$=\frac{1}{K}\sum_{k=1}^{K} [\alpha_k \mathbb{I}] + \overline{\mathbb{COV}}({\mathcal{G}( x,z^*)})$

\begin{figure}
\centering
\includegraphics[width=0.5\linewidth]{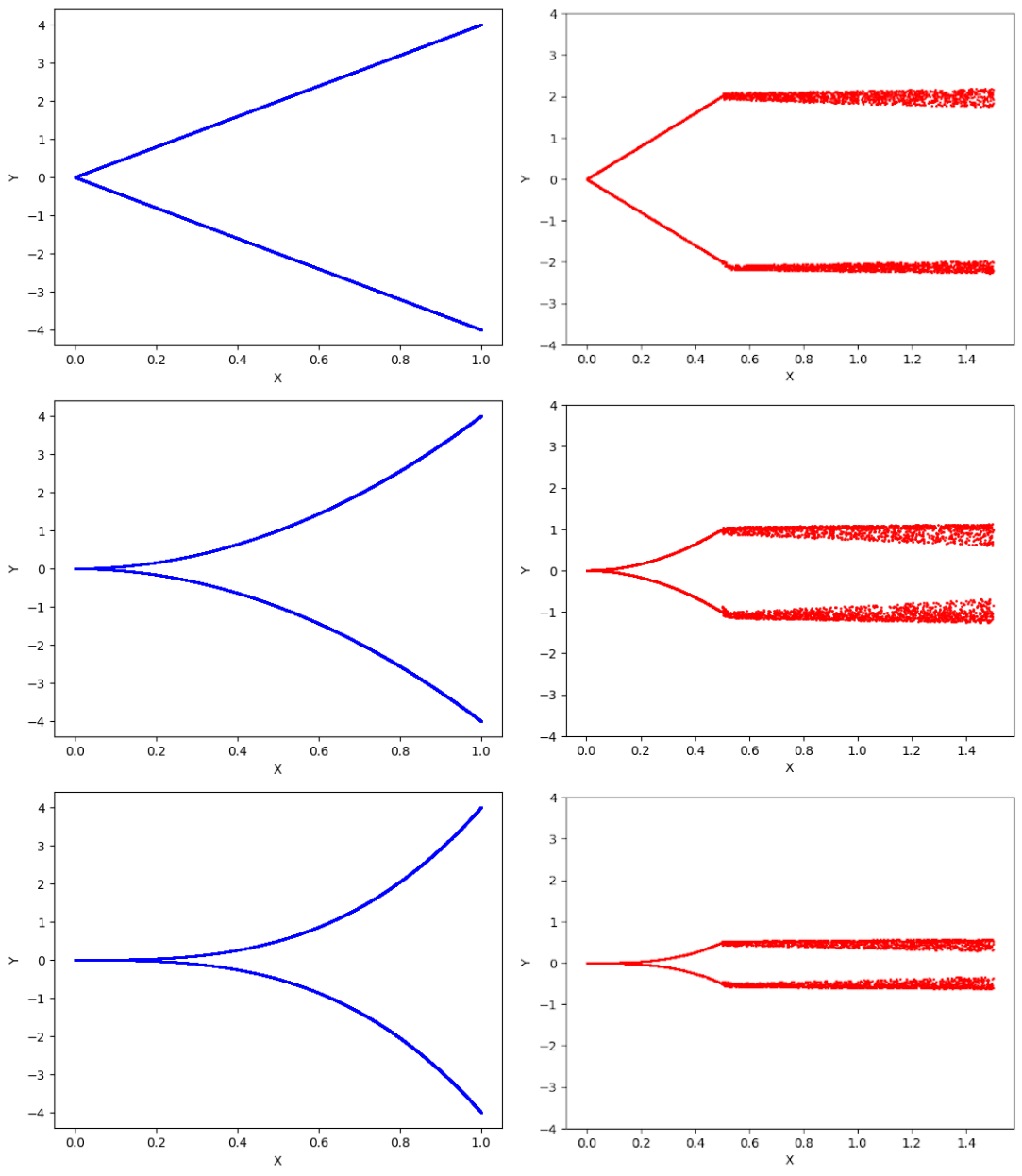} 
\caption{The uncertainty measurement illustration with the toy example. \emph{(left-column: ground truth, right-column: prediction)}. We train the model with $x \in [0,0.5]$ and test with $x \in [0,1.5]$. During the testing, we add a small Gaussian noise to $z^*$ at each $x$ and get stochastic outputs. As illustrated, the sample variance (the uncertainty measurement) increases as $x$ deviates from the observed data portion.}
\label{app:uncertainty_toy}
\end{figure}

\begin{figure}
\centering
\includegraphics[width=1.0\linewidth]{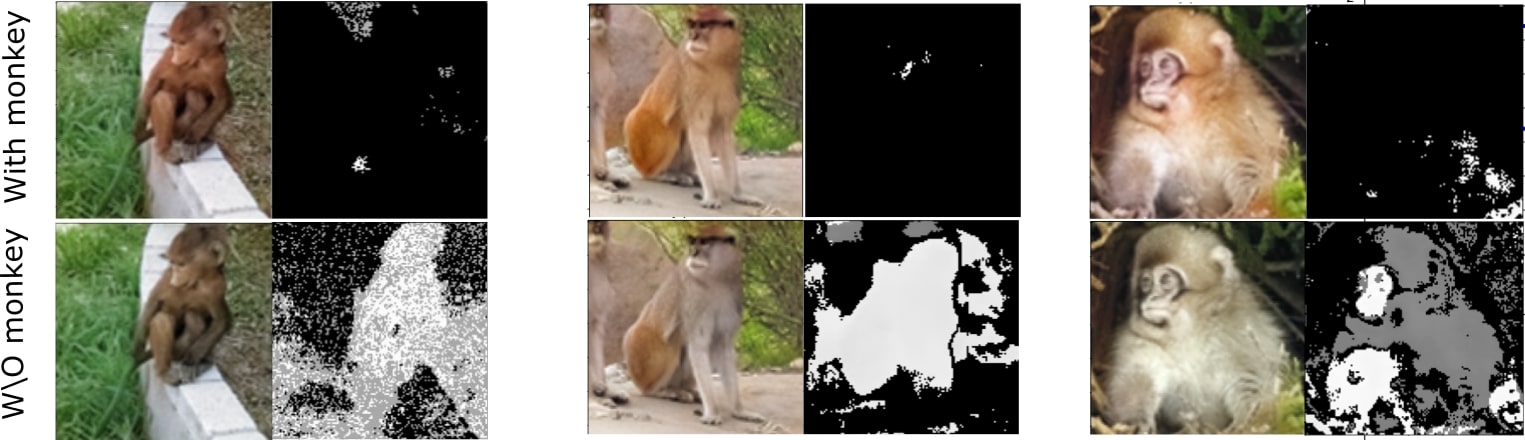}
\caption{Colorization predictions for models trained with and without monkey class. Output images are shown side by side with corresponding uncertainty maps. For models trained without monkey data, high uncertainty is predicted for pixels belonging to the monkey portion (intensity is higher for high uncertainty).}
\label{app:monkey}
\end{figure}

 Note that in similar to Bayesian uncertainty estimations, where an approximate distribution $q(w)$ is used to estimate $p(w|D)$, where $D$ is data, our model sample from the an empirical distribution $p(z^*|x)$. In practice, we treat $\alpha_k$ as a constant over all the samples--hence omit from the calculation---and use stochastic forward passes to obtain Eq.~\ref{uncertainty}. Then, the diagonal entries are used to calculate the uncertainty in the each dimension of the output. We test this hypothesis on the toy example and the colorization task, as shown in Fig.~\ref{app:uncertainty_toy} and Fig.~\ref{app:monkey}, respectively.
 
 \section{Experiments}
 \subsection{Experimental architectures}
 \label{sec:architecture}
 
\begin{figure}
\centering
\includegraphics[width=0.8\linewidth]{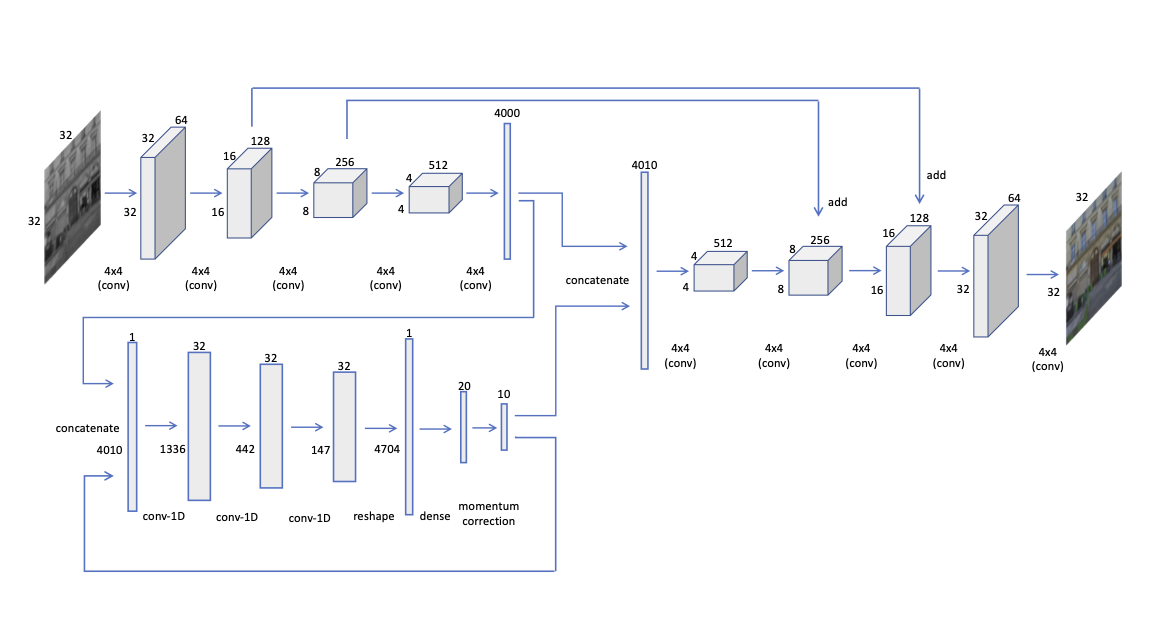} 
\includegraphics[width=0.8\linewidth]{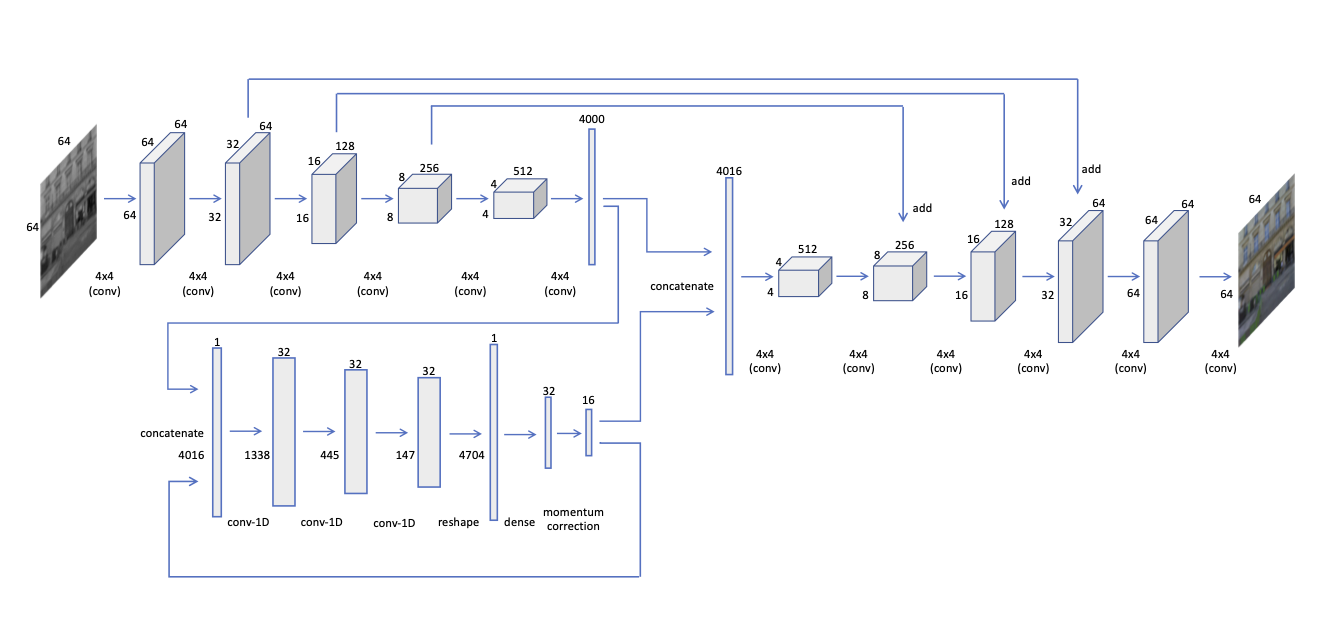} 
\includegraphics[width=0.8\linewidth]{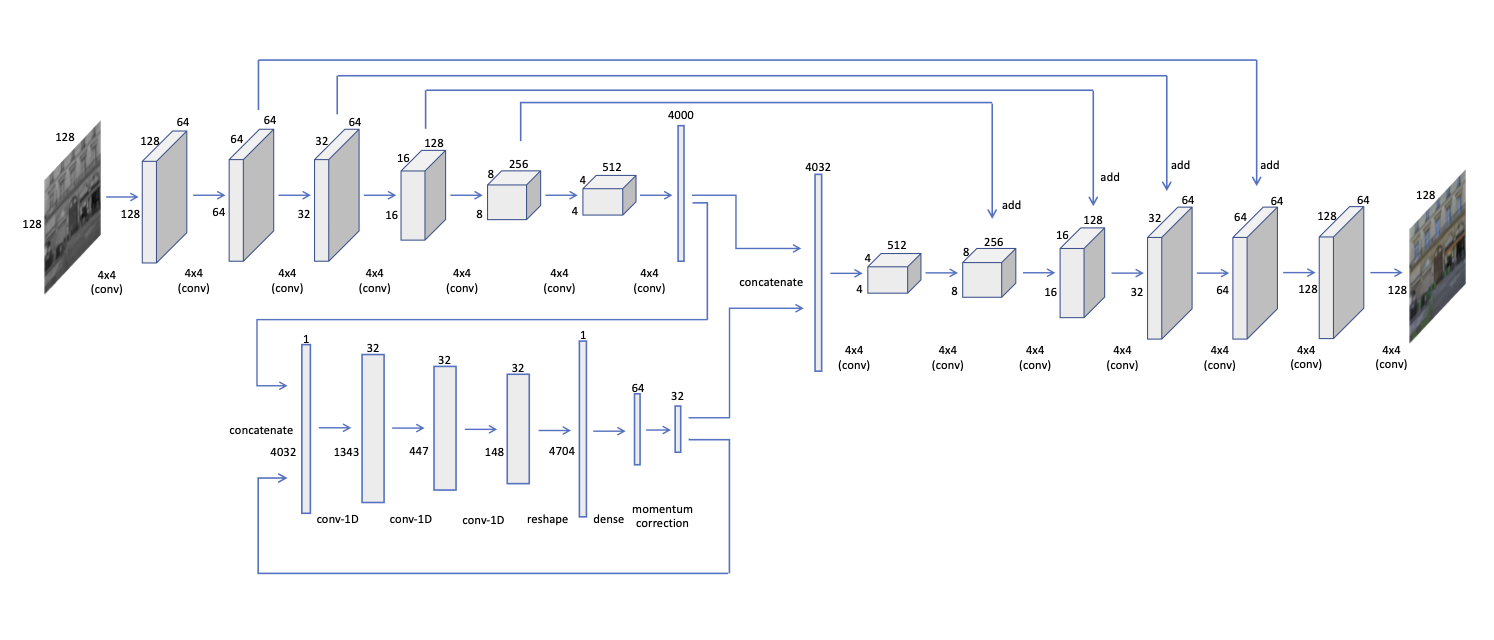} 
\includegraphics[width=0.8\linewidth]{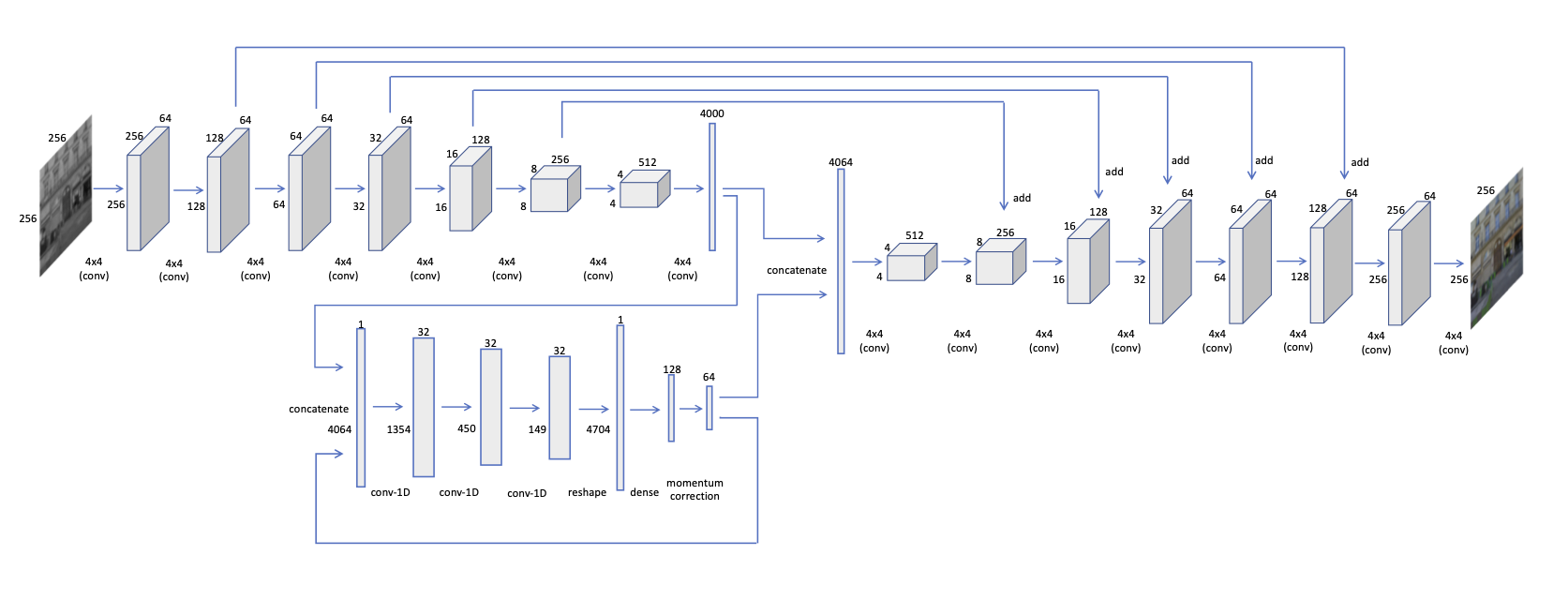} 
\caption{The model architecture for various input sizes. The same general structure in maintained with minimal changes to accomodate for the changing input size.}.
\label{app:arch_multi}
\end{figure}

 For the experiments on images, we mainly use  $128 \times 128$ size inputs. However, to demonstrate the scalability, we use several different architectures and show that the proposed framework is capable of converging irrespective of the architecture. Fig.~\ref{app:arch_multi} shows the architectures for different input sizes.
 
For training, we use the Adam optimizer with hyper-parameters $\beta_1 = 0.9, \beta_2 = 0.999, \epsilon = 1\times 10^{-8}$, and a learning rate $lr = 1 \times 10^{-5}$. We use batch normalization after each convolution layer, and leaky ReLu as the activation, except the last layer where we use $tanh$. All the weights are initialized using a random normal distribution with $0$ mean and $0.5$ standard deviation. Furthermore, we use a batch size of 20 for training, though we did not observe much change in performance for different batch sizes. We choose the dimensions of $z$ to be $10,16,32,64$ for $32 \times 32$, $64 \times 64$, $128 \times 128$, $256 \times 256$ input sizes, respectively. An important aspect to note here is that the dimension of $z$ should not be increased too much, as it would increase the search space for $z$ unnecessarily. While training, $z$ is updated $20$ times for a single $\mathcal{G}, \mathcal{H}$ update. Similarly, at inference, we use $20$ update steps for $z$ , in order to converge to the optimal solution. All the values are chosen empirically.

 \subsection{Evaluation metrics}
 \label{app:evaluation}
 Although heavily used in the literature, per pixel metrics such as PSNR do\sout{es} not effectively capture the perceptual quality of an image. To overcome this shortcoming, more perceptually motivated metrics have
 been proposed such as SSIM \cite{wang2004image}, MSSIM \cite{wang2003multiscale}, and FSIM \cite{zhang2011fsim}. However the similarity of two images is
 largely context dependant, and may not be captured by the aforementioned metrics. As a solution, recently, two deep feature based perceptual metrics--LPIP \cite{zhang2018perceptual} and PieAPP \cite{prashnani2018pieapp}--were proposed, which coincide well with the human judgement. To cover all these aspects, we evaluate our experiments against four metrics: LPIP, PieAPP, PSNR and SSIM. 
 
\subsection{Unbalanced color distributions}
\label{app:colordist}

The color distribution of a natural dataset in $a$ and $b$ planes (LAB space) are strongly biased towards low values. If not taken into account, the loss function can be dominated by these desaturated values. Richard \textit{et al.} \cite{zhang2016colorful} addressed this problem by rebalancing class weights according to the probability of color occurrence. However, this is only possible in a case where the output domain is discretized. To tackle this problem in the continuous domain, we push the output color distribution towards a uniform distribution as explained in Sec.~\ref{sec:colorization} in the main paper. 
 
 \subsection{Multimodality}
 \label{sec:multimodality}
\begin{figure}
\centering
\captionsetup{size=small}
\includegraphics[width=1.0\linewidth]{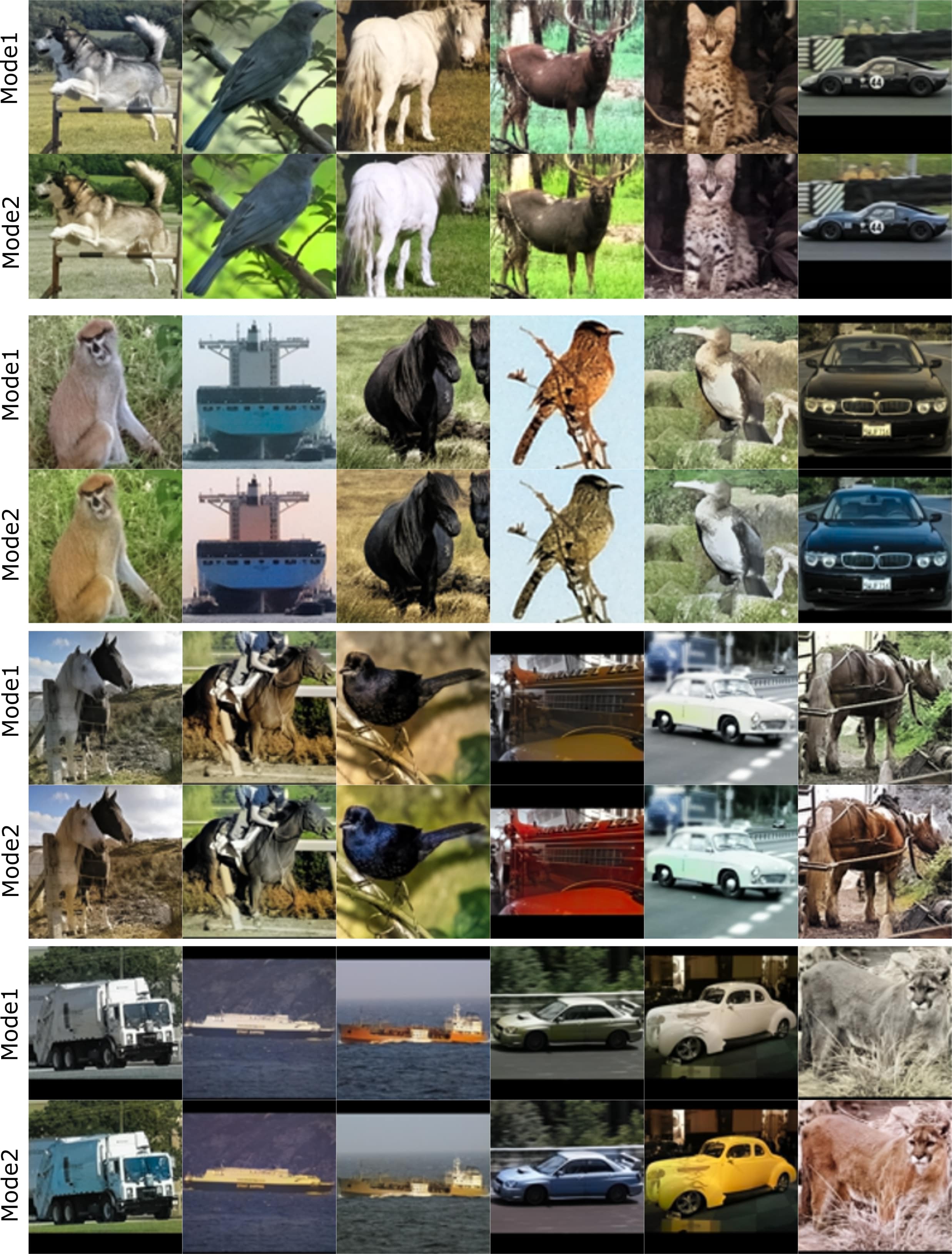} 
\caption{Multimodel predictions of our model in colorization}
\label{app:fig_colormulty2}
\end{figure}

\begin{figure}
\centering
\captionsetup{size=small}
\includegraphics[width=1.0\linewidth]{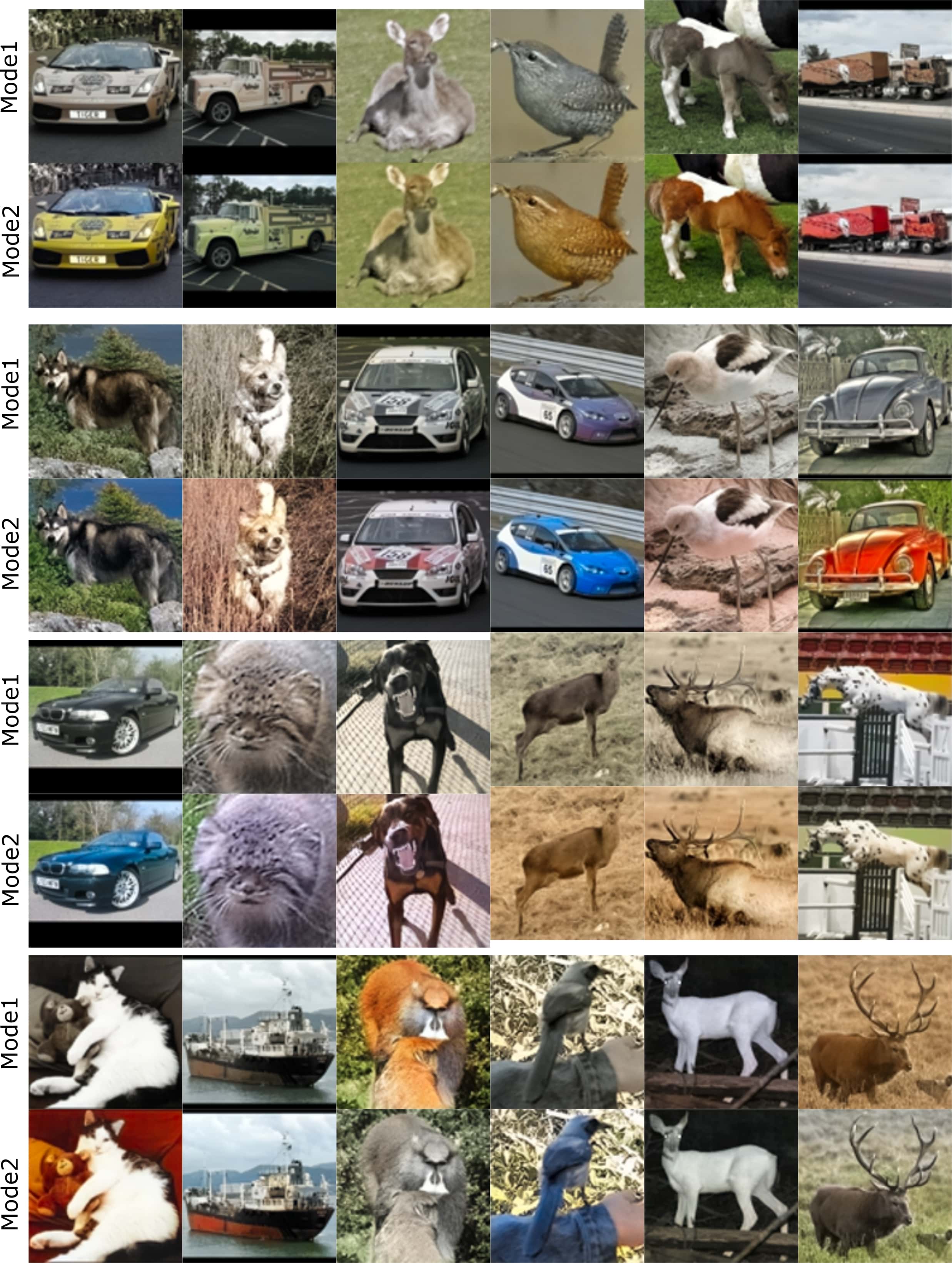} 
\caption{Multimodel predictions of our model in colorization}
\label{app:fig_colormulty1}
\end{figure}

\begin{figure}
\centering
\captionsetup{size=small}
\includegraphics[width=0.7\linewidth]{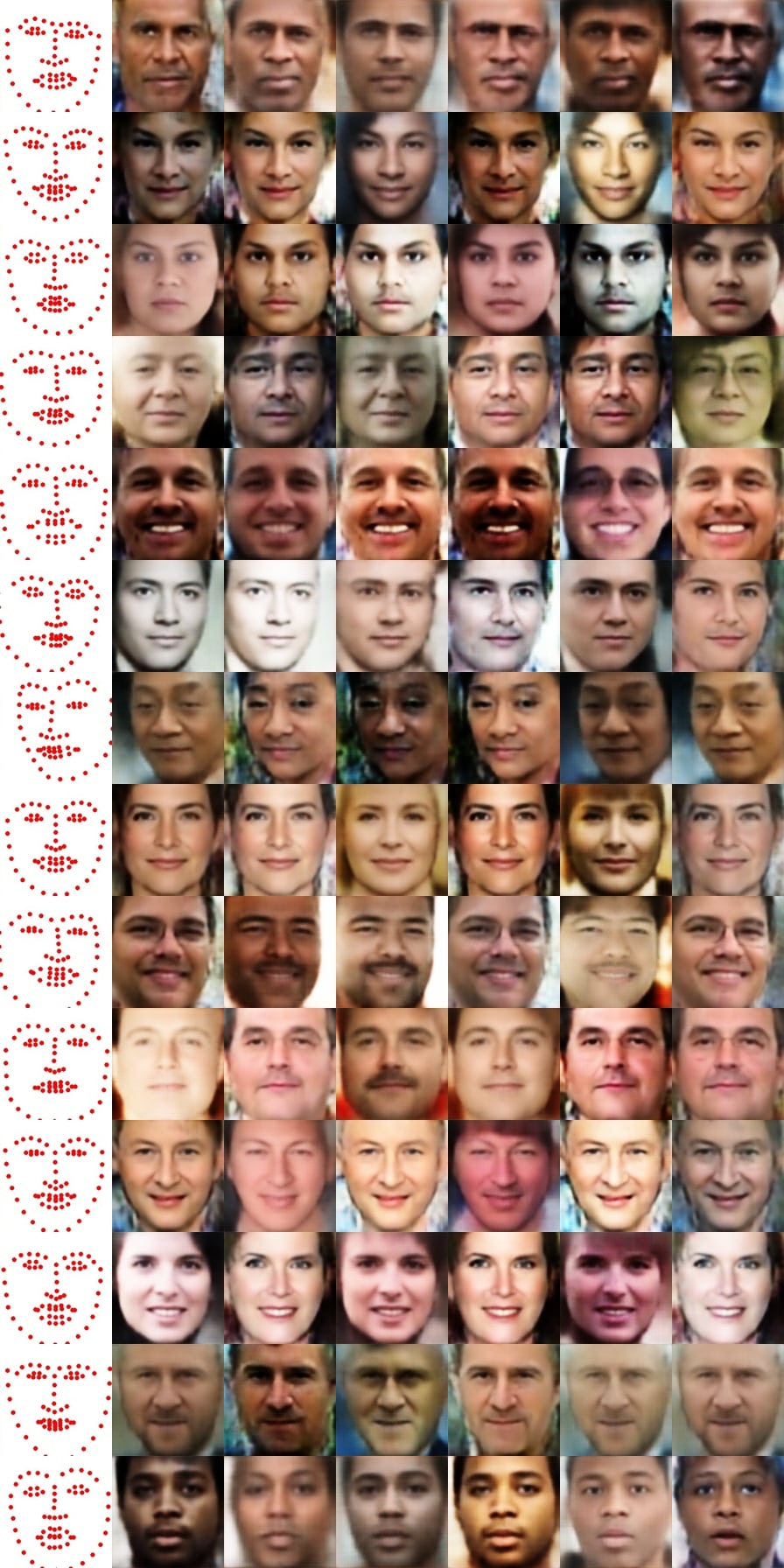} 
\caption{Multimodel predictions of our model in landmarks-to-faces.}
\label{app:fig_landmarks}
\end{figure}

\begin{figure}
\centering
\captionsetup{size=small}
\includegraphics[width=0.7\linewidth]{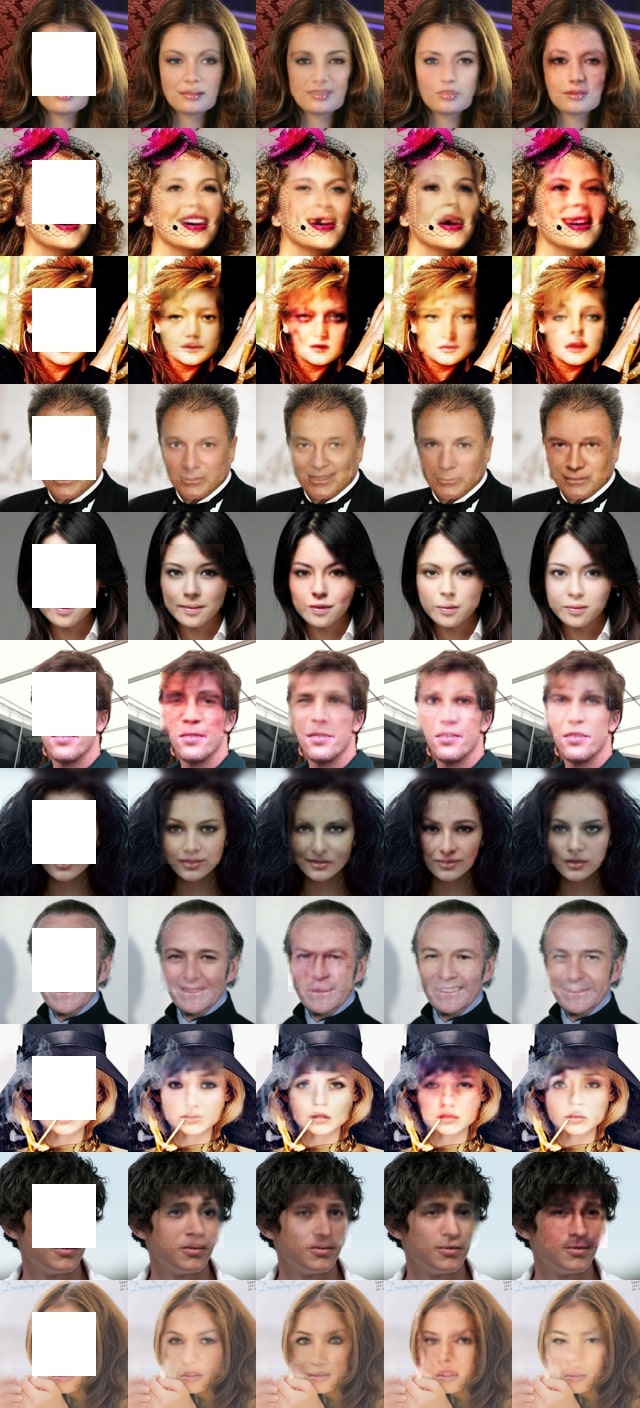} 
\caption{Multimodel predictions of our model in face inpainting.}
\label{app:fig_faces_multy}
\end{figure}

\begin{figure}
\centering
\captionsetup{size=small}
\includegraphics[width=0.7\linewidth]{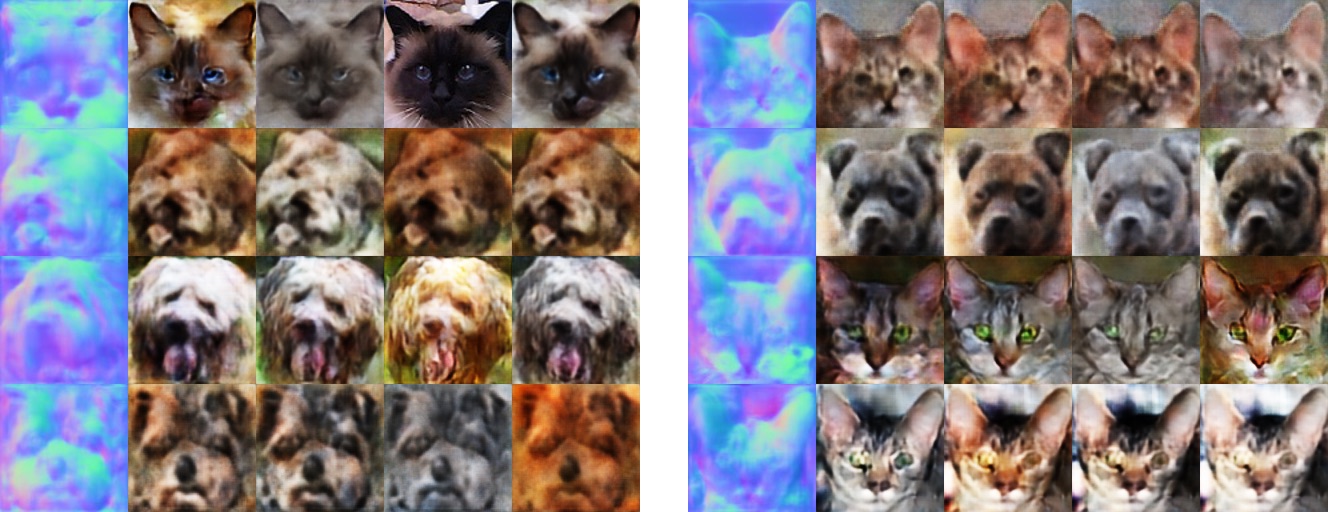} 
\caption{Multimodel predictions of our model in surface-normals-to-pet-faces. Note that this is generally a difficult task due to the diverse texture.}
\label{app:fig_pets_multy}
\end{figure}

\begin{figure}
\centering
\captionsetup{size=small}
\includegraphics[width=0.5\linewidth]{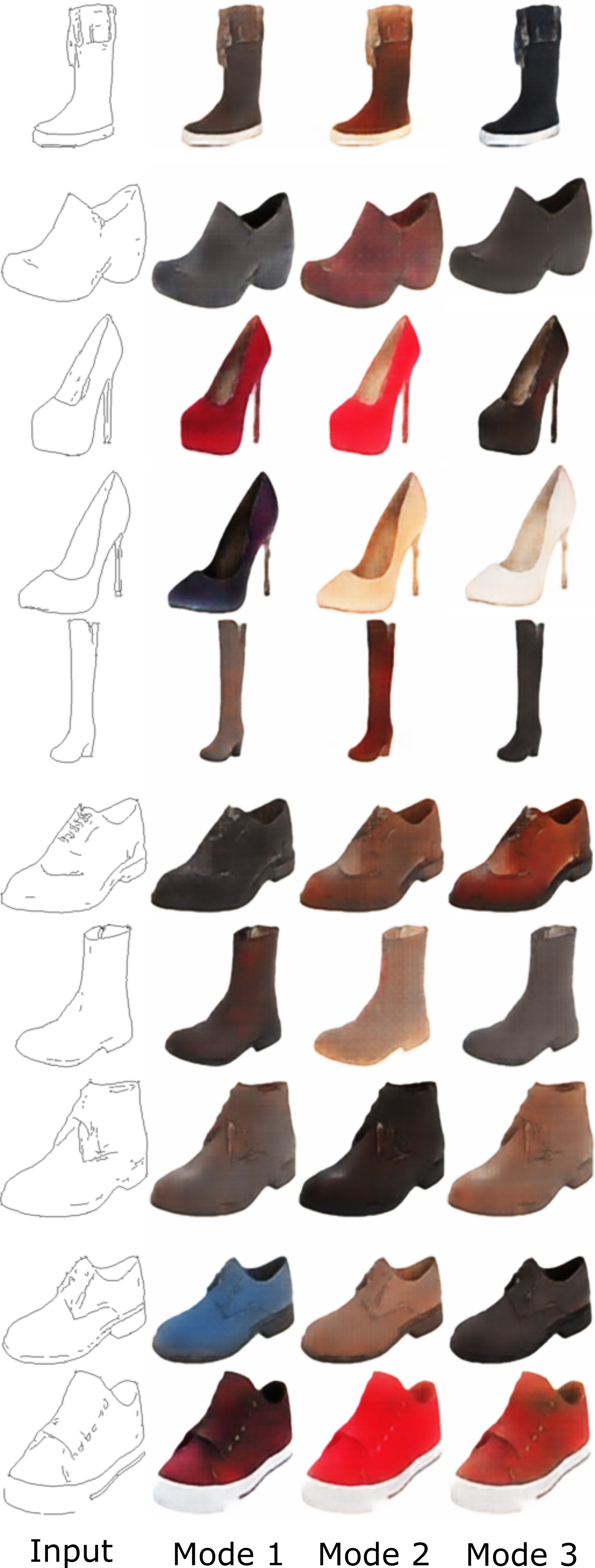} 
\caption{Multimodel predictions of our model in sketch-to-shoes translation.}
\label{app:fig_shoesmulty}
\end{figure}

\begin{figure}
\centering
\captionsetup{size=small}
\includegraphics[width=0.5\linewidth]{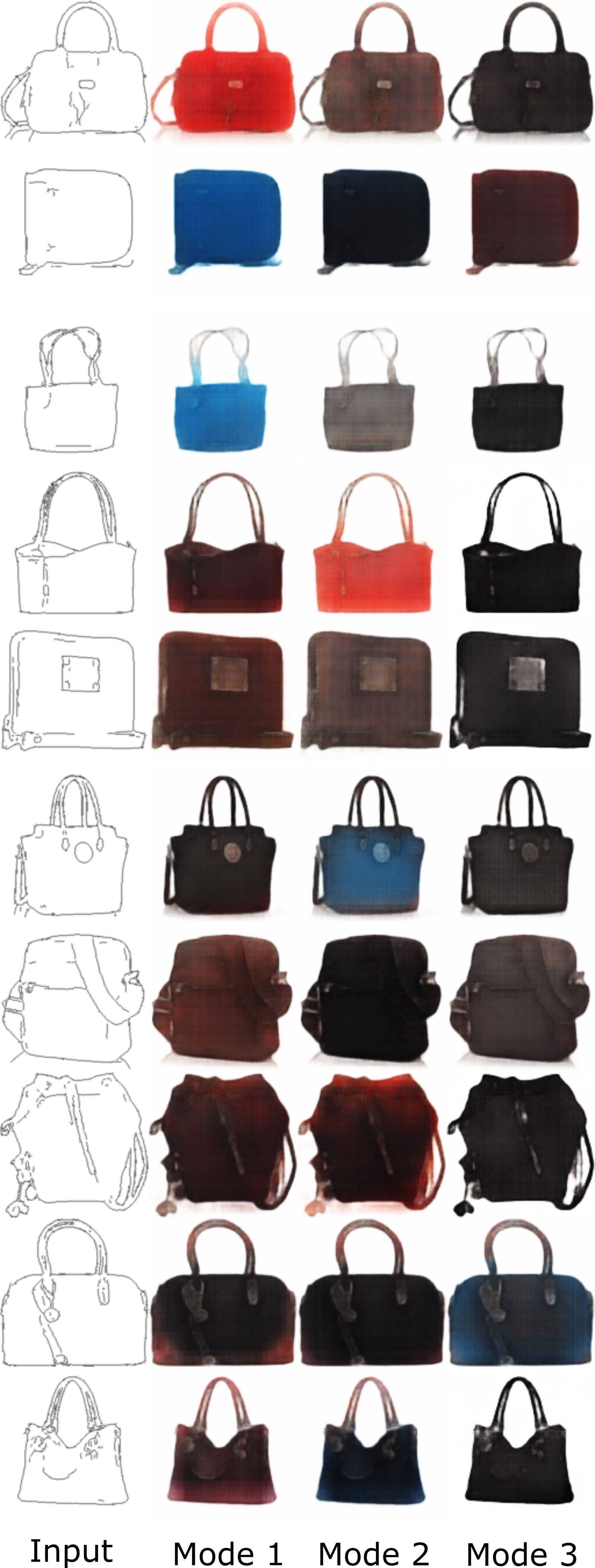} 
\caption{Multimodel predictions of our model in sketch-to-bag translation.}
\label{app:fig_bagsmulty}
\end{figure}

 An appealing attribute of our network is its ability to converge to multiple optimal modes at inference. A few
 such examples are shown in Fig.~\ref{app:fig_colormulty1},  Fig.~\ref{app:fig_colormulty2}, Fig.~\ref{app:fig_shoesmulty} Fig.~\ref{app:fig_bagsmulty}, Fig.~\ref{app:fig_faces_multy}, Fig.~\ref{app:fig_pets_multy} and Fig.~\ref{app:fig_landmarks}. For the \textit{facial-land-marks-to-faces} experiment, we used the UTKFace dataset \citep{zhifei2017cvpr}. For the \textit{surface-normals-to-pets} experiment, we used the Oxford Pet dataset \citep{parkhi12a}. In order to get the surface normal images, we follow Bansal \citet{bansal2017pixelnn}. First, we crop the bounding boxes of pet faces and then apply PixelNet \citep{bansal2017pixelnet} to extract surface normals. For \textit{maps-to-ariel} and \textit{edges-to-photos} experiments, we used the datasets provided by \citet{isola2017image}. 
 
 For measuring the diversity, we adapt the following procedure: 1) we generate 20 random samples from the model. 2) calculate the mean pixel value $\mu_{i}$ of each sample. 3) pick the closest sample $s_m$ to the average of all the mean pixels $\lambda = \frac{1}{20}\sum_{i=1}^{20} \mu_{i}$. 4) pick the $10$ samples which have
 maximum mean pixel distance from 
 $s_m$. 5) calculate the mean standard deviation of the $10$ samples picked in step 4. 6) repeat the experiment $5$ times for each model and get the expected standard deviation.


 \subsection{Colorization on STL dataset}
 \label{app:colorization}
\begin{figure}
\centering
\captionsetup{size=small}
\includegraphics[width=1.0\linewidth]{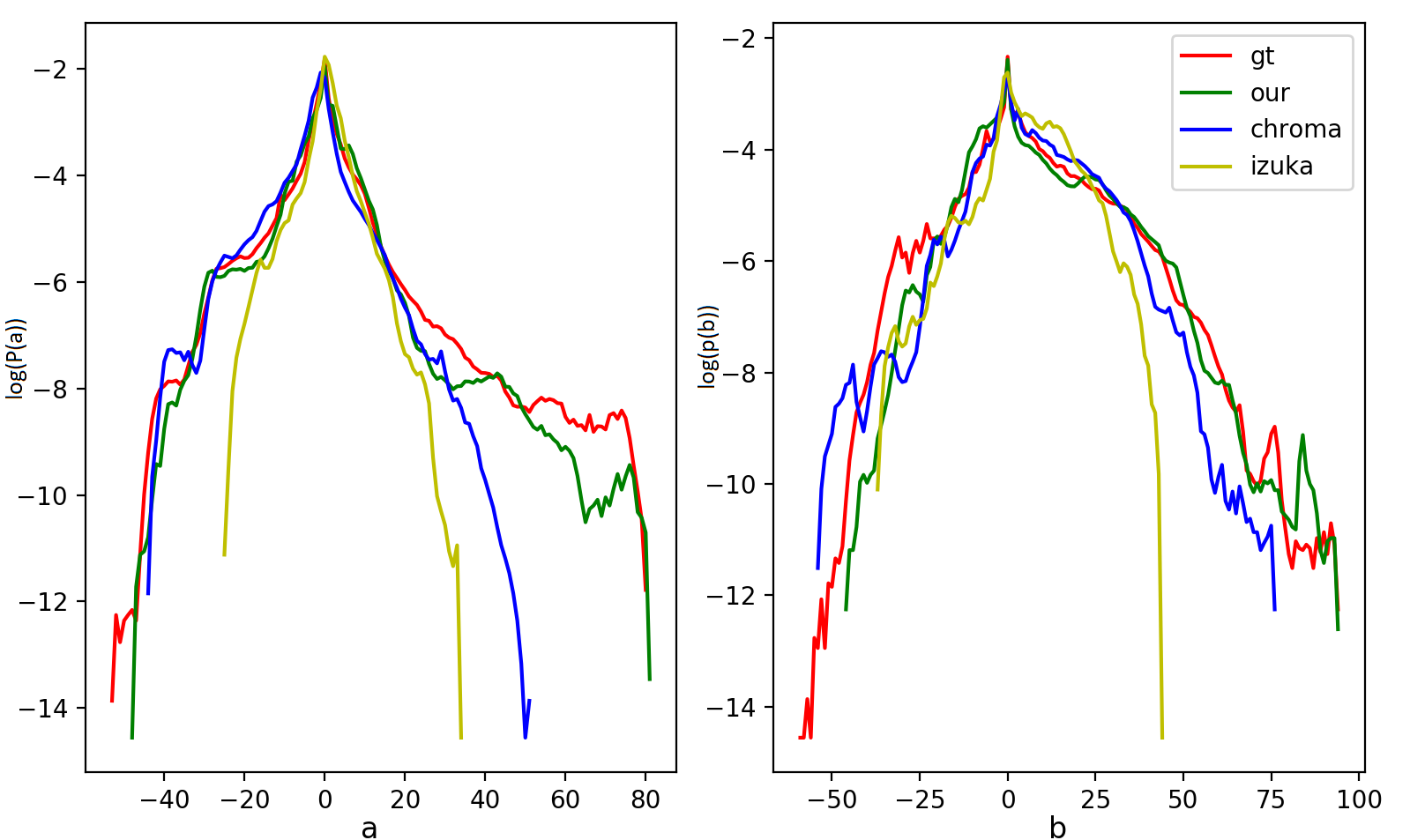} 
\caption{Color distribution comparison of $a,b$ planes. Our method produces the closest distribution to the ground truth.}
\label{app:fig_line}
\end{figure}

\begin{figure}
\centering
\captionsetup{size=small}
\includegraphics[width=1.0\linewidth]{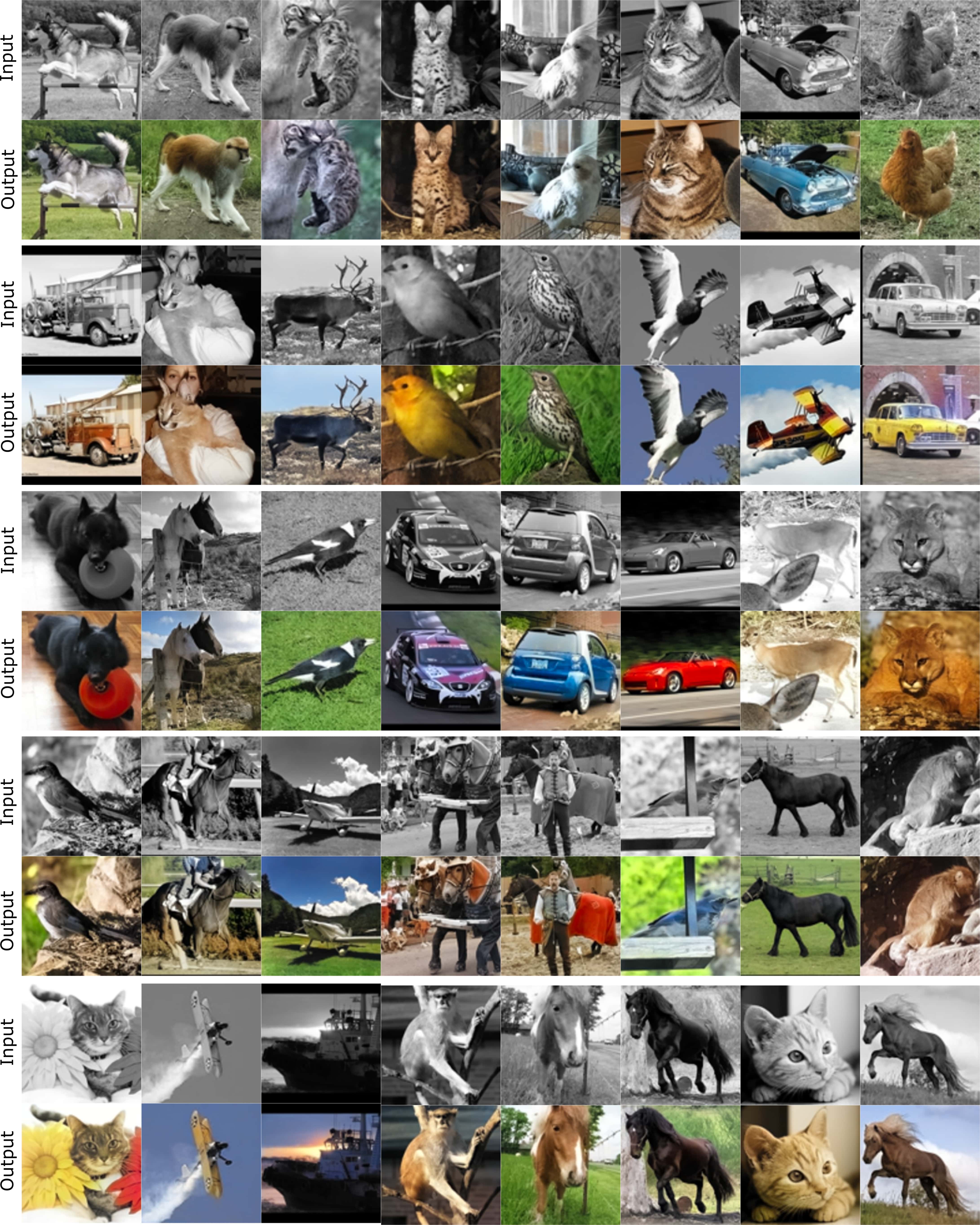} 
\caption{Qualitative results of our model in the colorization task on STL dataset.}
\label{app:fig_colorstl}
\end{figure}

 Additional colorization examples on the STL dataset are shown in Fig.~\ref{app:fig_colorstl}. We also compare the color distributions of the predicted $a,b$ planes with state-of-the-art. The results are shown in Fig.~\ref{app:fig_line} and Table \ref{tbl:distribution}. As evident, our method predicts the closest color distribution to the ground truth.

 \begin{table}[]
\centering
\small
\begin{tabular}{|l|l|l|}
\hline
Method & a   & b  \\ \hline
Chroma      & 0.71 & 0.78 \\ 
Izuka  & 0.68 & 0.63 \\
\hline
Ours    & 0.82\% & 0.80\%  \\ \hline
\end{tabular}
\caption{{IOU of the predicted color distributions against the ground truth. Our method shows better results.}}
\label{tbl:distribution}
\end{table}
 

 \subsection{Colorization on ImageNet dataset}
 \label{app:colorization}
\begin{figure}
\centering
\captionsetup{size=small}
\includegraphics[width=1.0\linewidth]{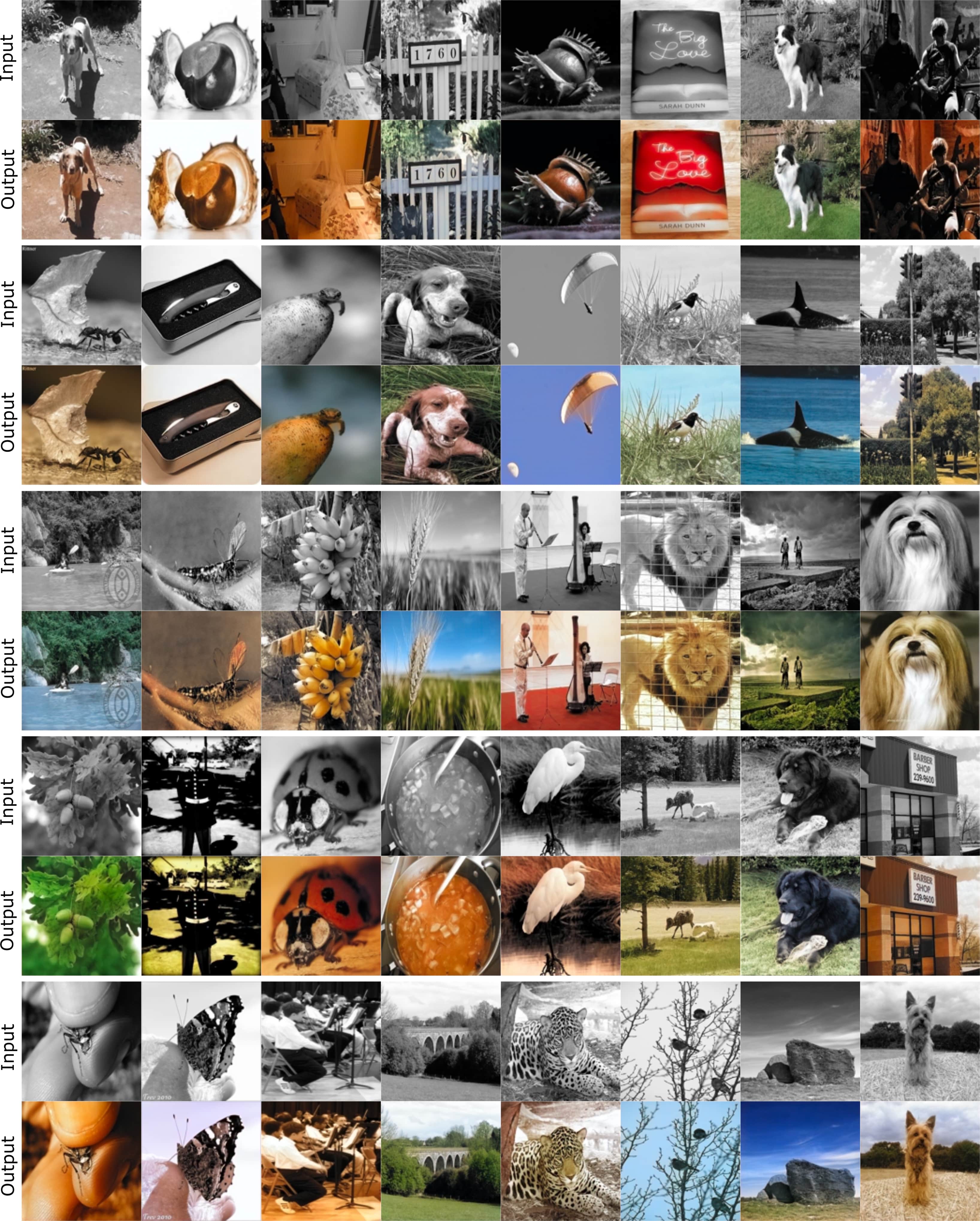} 
\caption{Qualitative results of our model in the colorization task on ImageNet dataset.}
\label{app:fig_colorimg}
\end{figure}

  Additional colorization examples on the ImageNet dataset are shown in Fig.~\ref{app:fig_colorimg}.


\subsection{Self-supervised learning setup}
\label{app:ssl}
Here we evaluate the performance of our model on down-stream tasks, using three distinct setups involving bottleneck features of trained models. The bottleneck layer features (of models trained on some dataset) are fed to a fully-connected layer and trained on a different dataset.

The baseline experiment uses the output of the penultimate layer in a Resnet-50 trained on ImageNet for classification as the bottleneck features. The comparison to state-of-the-art experiment involves \cite{pennet} where the five outputs of its multi-scale decoder are max-pooled and concatenated to use as the bottleneck features. The outputs of layers before this were also experimented with, and the highest performance was obtained for these selected features. In our network, the output of the encoder network was used  as the bottleneck features.

\subsection{Scalability}
\label{app:scalility}

One promising attribute of the proposed method compared to the state-of-the-art is its scalability. In other words, we propose a generic framework which is not bound to the architecture, hence, the model can be scaled to different input sizes without affecting the convergence behaviour. To demonstrate this, we use 4 different architectures and train them on 4 different input sizes ($32 \times 32$, $64 \times 64$, $128 \times 128$, $256 \times 256$) on the same tasks: image completion and colorization. The different architectures we use are shown in  Fig.~\ref{app:arch_multi}.

\subsection{Ablation study on the $z$ dimension}

To demonstrate the effect of dimension of $z$ on the model accuracy, we conduct an ablation study for the colorization task for the input size $128 \times 128$. Table~\ref{tbl:ablation} shows the results. The quality of the outputs increases to a maximum when $dim(z) = 32$, and then decreases. This is intuitive because when the search space of $z$ gets unnecessarily high, it becomes difficult for $\mathcal{Z}$ to learn the paths to optimum modes, due to limited capacity.

\begin{table}[]
\centering
\small
\begin{tabular}{|l|l|l|l|}
\hline
Dimensionality & LPIP  & PieAPP  & Diversity \\ \hline
5     & 1.05 & 3.40 & 0.01\\ \hline
10    & 0.58 & 2.91 & 0.018\\ \hline
16    & 0.14 & 1.89 & 0.021\\ \hline
32    & \textbf{0.12}  & \textbf{1.47} &  0.043 \\ \hline
64    & 0.27  & 1.71 & \textbf{0.048}\\ \hline
128    & 0.69 & 2.12 & 0.043\\ \hline
\end{tabular}
\caption{Ablation study against the dimension of $z$ for the colorization task ($128 \times 128$ inputs).}
\label{tbl:ablation}
\end{table}

\subsection{User studies}
\label{app:psycological}
Evaluation of synthesized images is an open problem \citep{salimans2016improved}. Although recent metrics such as LPIP \citep{zhang2018perceptual} and PieAPP \citep{prashnani2018pieapp} have
been proposed, which coincide closely with
human judgement, perceptual user studies remain the preferred method. Therefore, to evaluate the quality of our synthesized images in the colorization task, we conduct two types of user studies: a Turing test and a psychophysical study. In the Turing test, we show the users a series of  paired images, ground truth and our predictions, and ask the users to pick the most realistic image. Here, following \citet{zhang2016colorful}, we display each image for 1 second, and then give the users an unlimited amount of time to make the choice. For the psychophysical study, we choose the two best performing methods according to the LPIP metric: \citet{vitoria2020chromagan} and  \citet{iizuka2016let}. We create a series of batches of three images, \citet{vitoria2020chromagan},  \citet{iizuka2016let} and ours, and ask the users to pick the best quality image. In this case, each batch is shown to the users for 5 seconds, and the users have to make this decision during that time. We conduct the Turing test on ImageNet, and the psychophysical study on both ImageNet and STL datasets. For each test, we use $500$ randomly sampled batches and $\sim 15$ users.

We also conduct Turing tests to evaluate the image completion tasks on Facades and Celeb-HQ datasets. The results are shown in Table~\ref{tbl:turing_inpainting}.

\begin{table}[]
\centering
\small
\begin{tabular}{|l|l|l|}
\hline
Dataset & Celeb-HQ   & Facades  \\ \hline
GT      & 59.11\% & 55.75\% \\ \hline
Ours    & 40.89\% & 44.25\%  \\ \hline
\end{tabular}
\caption{ {Turing Test for GT vs ours on popular image datasets Celeb-HQ and Facades.}}
\label{tbl:turing_inpainting}
\end{table}

\subsection{Image completion}
\begin{figure}
\centering
\captionsetup{size=small}
\includegraphics[width=\linewidth]{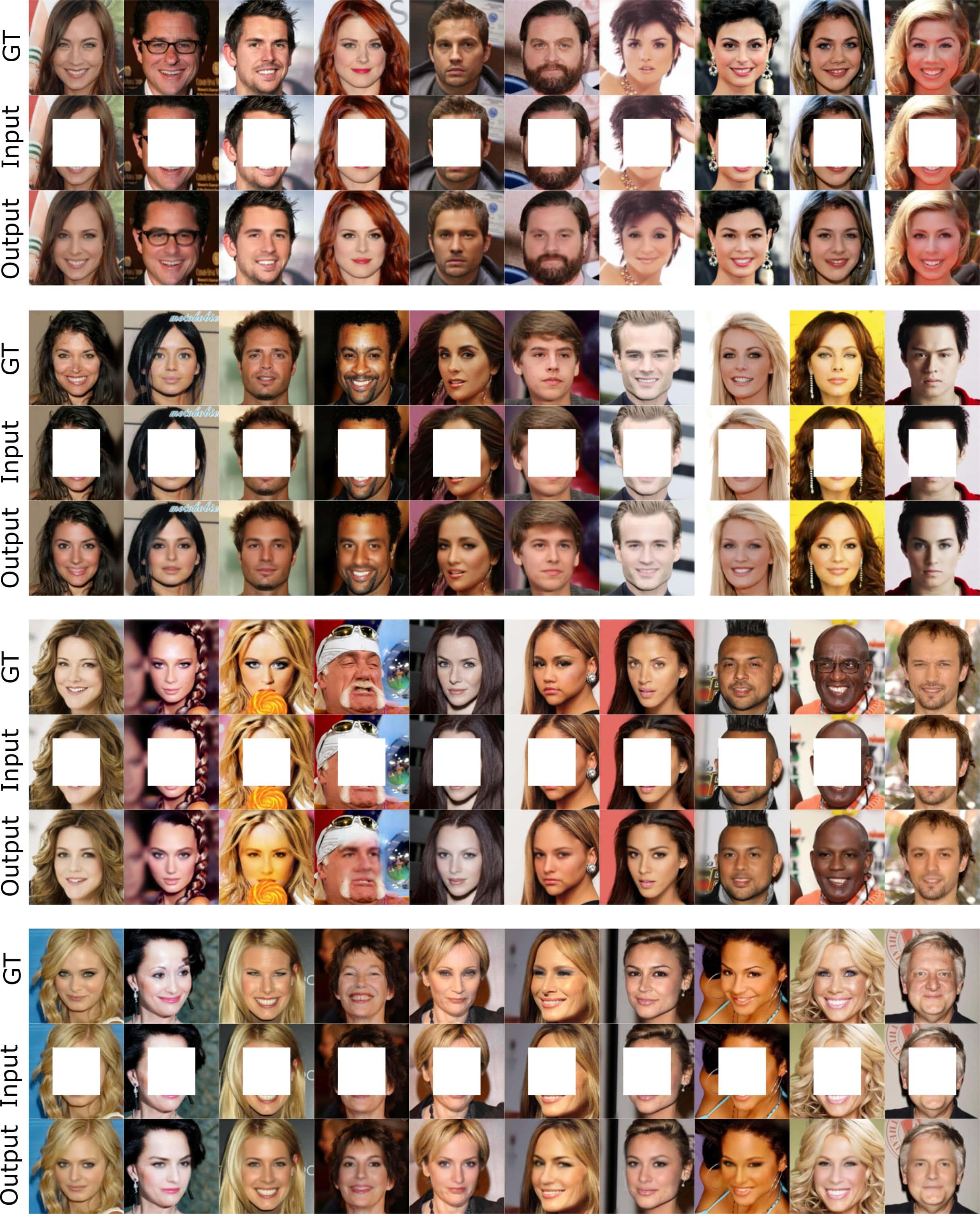} 
\caption{Qualitative results of our model in the image completion task on Celeb-HQ dataset.}\label{fig:compl_supp1}
\end{figure}

\begin{figure}
\centering
\captionsetup{size=small}
\includegraphics[width=\linewidth]{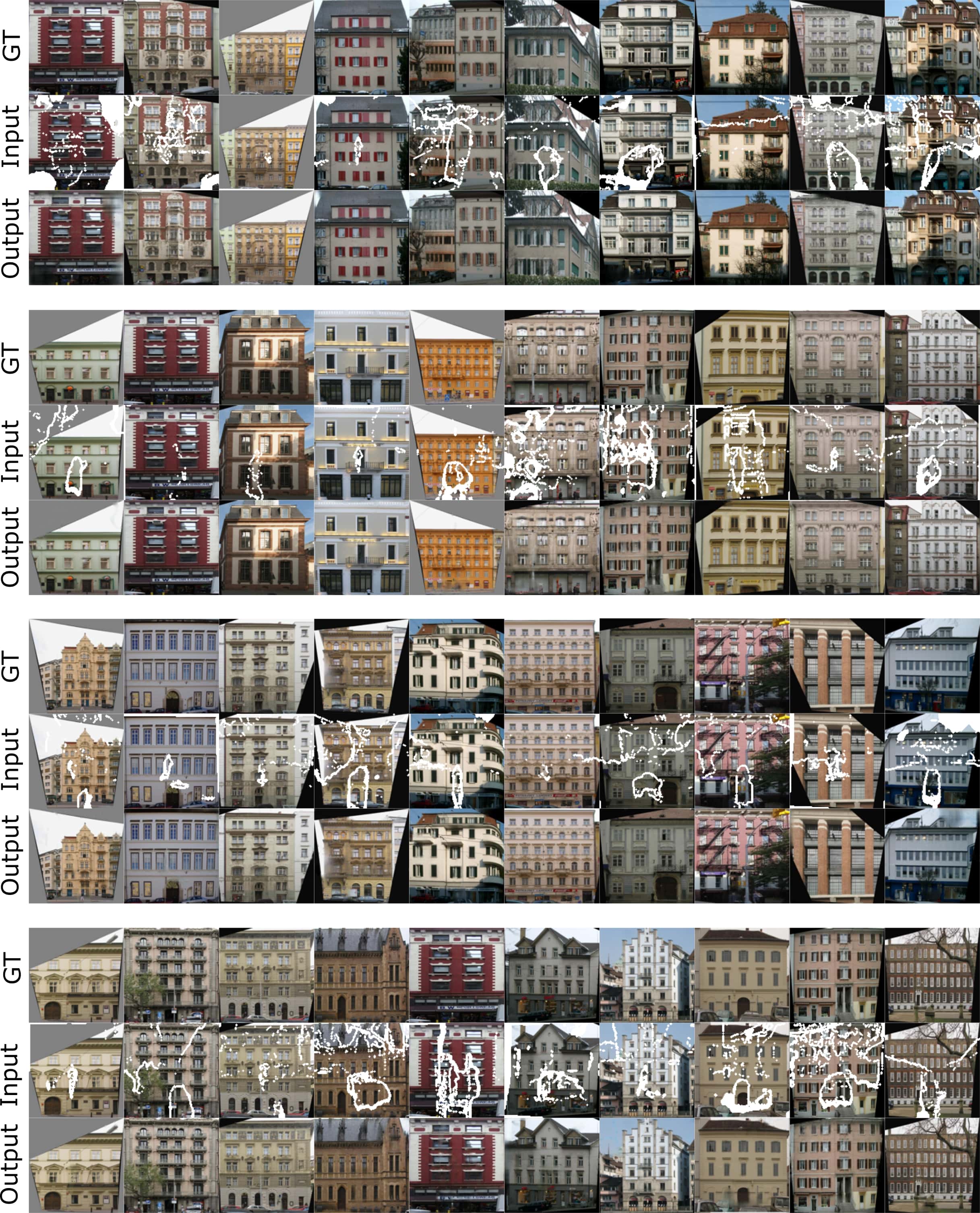} 
\caption{Qualitative results of our model in the image completion task on Facades dataset.}\label{fig:compl_supp2}
\end{figure}

The additional image completion examples are provided in Figs.~\ref{fig:compl_supp1} and \ref{fig:compl_supp2}. Our turing test results on Celeb-HQ and Facades are shown in Table~\ref{tbl:turing_inpainting}.

\label{app:imagecomplete}

\subsection{3D spectral map denoising}
\label{sec:appdenoising}
In this experiment, we use two types of spectral moments: spherical harmonics and Zernike polynomials (see App. \ref{app:spectral}). The minimum number of sample points required to accurately represent a finite energy function in a particular function space depends on the used sampling theorem. According to Driscoll and Healy’s theorem \cite{driscoll1994computing},  $4N^2$ equiangular sampled points are needed to represent a function on $\mathbb{S}^2$ using spherical  moments at a maximum degree $N$. Therefore, we compute the first $16384$ spherical moments of 3D objects where $l \leq 128$ by sampling $256 \times 256$  equiangular points in $\theta$ and $\phi$ directions, where $0 \leq \theta \leq \pi$ and $0 \leq \phi \leq 2\pi$. Afterwards, we arrange the spherical moments as a $128 \times 128$ feature map, and convolve with a $2 \times 2$ kernel with stride size $2$ to downsample the feature map to $64 \times 64$ size. The output is then fed to $64$-size architecture. We add Gaussian noise and mask portions of the spectral map to corrupt it. Afterwards, the model is trained to de-noise the input.

For Zernike polynomials, we compute the first $100$ moments for each 3D object where $n \leq 9$, and arrange the moments as a $10 \times 10$ feature map. Then, the feature map is upsampled using transposed convolution by using a $5 \times 5$ kernel and with a stride size $3$. The upsamapled feature map is fed to a $32$-size network and trained end-to-end to denoise. We first train the network on 55k objects in ShapeNet, and then apply the trained network on the Modelnet10 and Modelnet40 to extract the bottleneck features. These features are then fed to a single fully connected layer for classification.

\section{Spectral domain representation of 3D objects}
\label{app:spectral}

Spherical harmonics and Zernike polynomials are orthogonal and complete functions in $\mathbb{S}^2$ and $\mathbb{B}^3$, respectively, hence, 3D point clouds can be represented by a set of coefficients corresponding to a linear combination of these functions \cite{perraudin2019deepsphere,ramasinghe2019volumetric, ramasinghe2019spectral}.

\subsection{Spherical harmonics}
Spherical harmonics are complete and orthogonal functions defined on the unit sphere ($\mathbb{S}^2$) as,

\begin{equation}
Y_{l,m} (\theta, \phi) = (-1)^m\sqrt{\frac{2l+1}{4\pi}\frac{(l-m)!}{(l+m)!}}P_l^m(\cos\phi)e^{im\theta},
\end{equation}

where $\theta \in [0, 2\pi ]$ is the azimuth angle, $\phi \in [0,\pi]$ is the polar angle, $l \in \mathbb{Z}^{+}$, $m  \in \mathbb{Z}$, and $|m| < l$. Here, $P_l^m(\cdot)$ is the associated Legendre function defined as,
\begin{equation}
P_l^m(x) = (-1)^m \frac{(1-x^2)^{m/2}}{2^ll!}\frac{d^{l+m}}{dx^{l+m}}(x^2-1)^l.
\end{equation}

Spherical harmonics demonstrate the following orthogonal property,

\begin{equation}
    \int_{0}^{2\pi} \int_{0}^{\pi} Y_l^m (\theta, \phi) Y_{l'}^{m'}(\theta, \phi)^\dagger \sin{\phi}\, d\phi d\theta  = \delta_{l,l'} \delta_{m,m'},
\end{equation}

where $^\dagger$ denotes the complex conjugate and,
\begin{equation}
\delta_{m,m'} =  
\begin{cases}
    1,& \text{if } m = m' \\
    0,              & \text{otherwise}.
\end{cases}
\end{equation} 

Since spherical harmonics are complete in $\mathbb{S}^2$, any function $f:\mathbb{S}^2 \rightarrow \mathbb{R}$ with finite energy can be rewritten as
\begin{equation}
f(\theta,\phi) = \sum_{l}\sum_{m=-l}^{l}\hat{f}(l,m)Y_{l,m}(\theta,\phi), \quad
\end{equation}
where,
\begin{equation}
    \hat{f}(l,m) = \int_0^{\pi} \int_0^{2 \pi}f(\theta, \phi) Y_{l}^{m}(\theta, \phi)^\dagger \sin\phi \, d\phi d\theta .
\end{equation}

\subsection{3D Zernike polynomials}
3D Zernike polynomials are  complete and orthogonal on $\mathbb{B}^3$ and defined as,
\begin{equation}
Z_{n,l,m}(r, \theta, \phi) = R_{n,l}(r)Y_{l,m}(\theta, \phi),
\end{equation}
where,
\begin{equation}
    R_{n,l}(r) = \sum_{v=0}^{(n-1)/2} q^v_{nl} r^{2v +l},
\end{equation}

and  $q^v_{nl}$ is a scaler defined as
\begin{equation}
    q^v_{nl} = \frac{(-1)^{\frac{(n-l)}{2}}}{2^{(n-l)}} \sqrt{\frac{2n+3}{3}}{(n-l)\choose \frac{(n-l)}{2}}(-1)^v \frac{{\frac{(n-l)}{2}\choose v} {2(\frac{(n-l)}{2} + l + v) + 1\choose (n-l)  } }{{ \frac{(n-l)}{2} + l + v \choose \frac{(n-l)}{2}}}.
\end{equation}

Here $Y_{l,m}(\theta, \phi)$ is the spherical harmonics function, $n \in \mathbb{Z}^{+}$, $l \in [0, n]$, $m \in [-l, l]$ and $n-l$ is even. 3D Zernike polynomials also show orthogonal properties as,
\begin{equation}
\begin{split}
    \int_{0}^{1} \int_{0}^{2 \pi} \int_{0}^{\pi}  Z_{n,l,m}(\theta, \phi, r) & {Z}^{\dagger}_{n',l',m'} r^2  \sin\phi \, drd\phi d\theta \\
    & = \frac{4\pi}{3} \delta_{n,n'}\delta_{l,l'}\delta_{m,m'},
\end{split}
\end{equation}

Since Zernike polynomials are complete in $\mathbb{B}^3$, any function $f:\mathbb{B}^3 \rightarrow \mathbb{R}$ with finite energy can be rewritten as,

\begin{equation}
\label{reconstruction}
f(\theta, \phi, r) = \sum\limits_{n=0}^{\infty} \sum\limits_{l = 0}^{n} \sum\limits_{m = -l}^{l} \Omega_{n,l,m}(f) Z_{n,l,m}(\theta, \phi, r)
\end{equation}
where $\Omega_{n,l,m}(f)$ can be obtained using
\begin{equation}
\label{omega}
\Omega_{n,l,m}(f) = \int_{0}^{1} \int_{0}^{2 \pi} \int_{0}^{\pi}  f(\theta, \phi, r) {Z}^{\dagger}_{n,l,m} r^2 \sin\phi \, drd\phi d\theta.
\end{equation}

\section{Image-to-Image translation}
\subsection{Sketch-to-shoes qualitative results}
\begin{figure}
\centering
\captionsetup{size=small}
\includegraphics[width=1.0\linewidth]{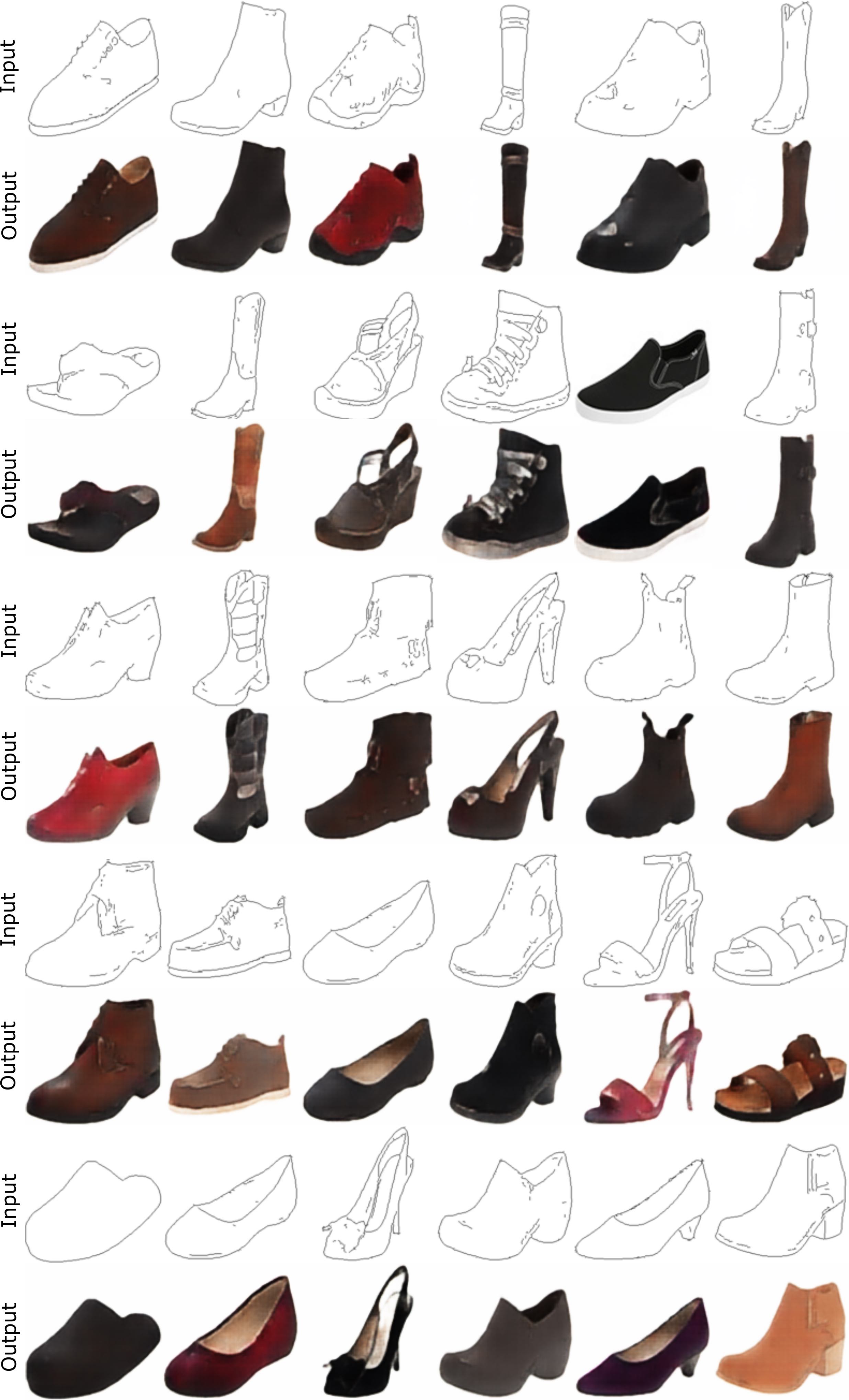} 
\caption{Qualitative results of our model in sketch-to-shoe translation.}
\label{fig:sketch-to-shoes}
\end{figure}

Additional qualitative results of the sketch-to-shoe translation task are shown in Fig.~\ref{fig:sketch-to-shoes}.

\subsection{Map-to-photo qualitative results}\label{sec:maptophoto}
\begin{figure}
\centering
\captionsetup{size=small}
\includegraphics[width=\linewidth]{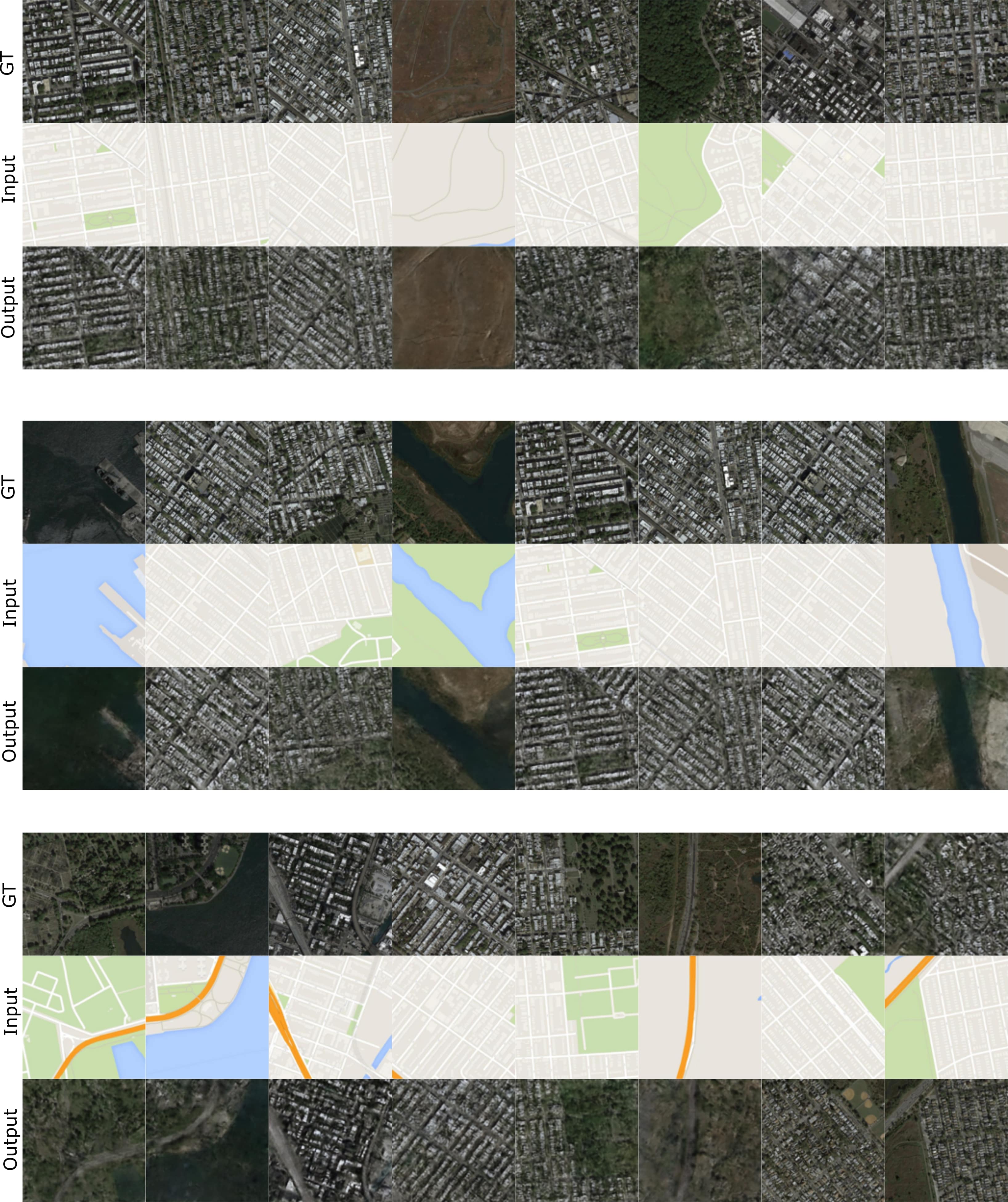} 
\caption{Qualitative results of our model in map-to-photo translation.}
\label{fig:map-to-photo}
\end{figure}

Additional qualitative results of the map-to-photo translation task are shown in Fig.~\ref{fig:map-to-photo}.
\label{app:colorization}

\section{Convergence at inference}

\begin{figure}[h]
    \centering
    \includegraphics[width=0.33\linewidth]{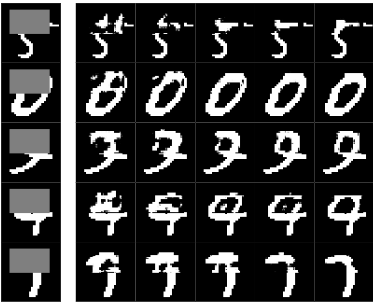}
    \caption{Output gets better as the $z$ traverse to the optimum position at inference. Left column is the input. Five right columns show outputs at iterations 2, 4, 6, 8 and 10 (from left to right).}
    \label{fig:mnist_example_finetune}
\end{figure}

\begin{figure}
\centering
\captionsetup{size=small}
\includegraphics[width=\linewidth]{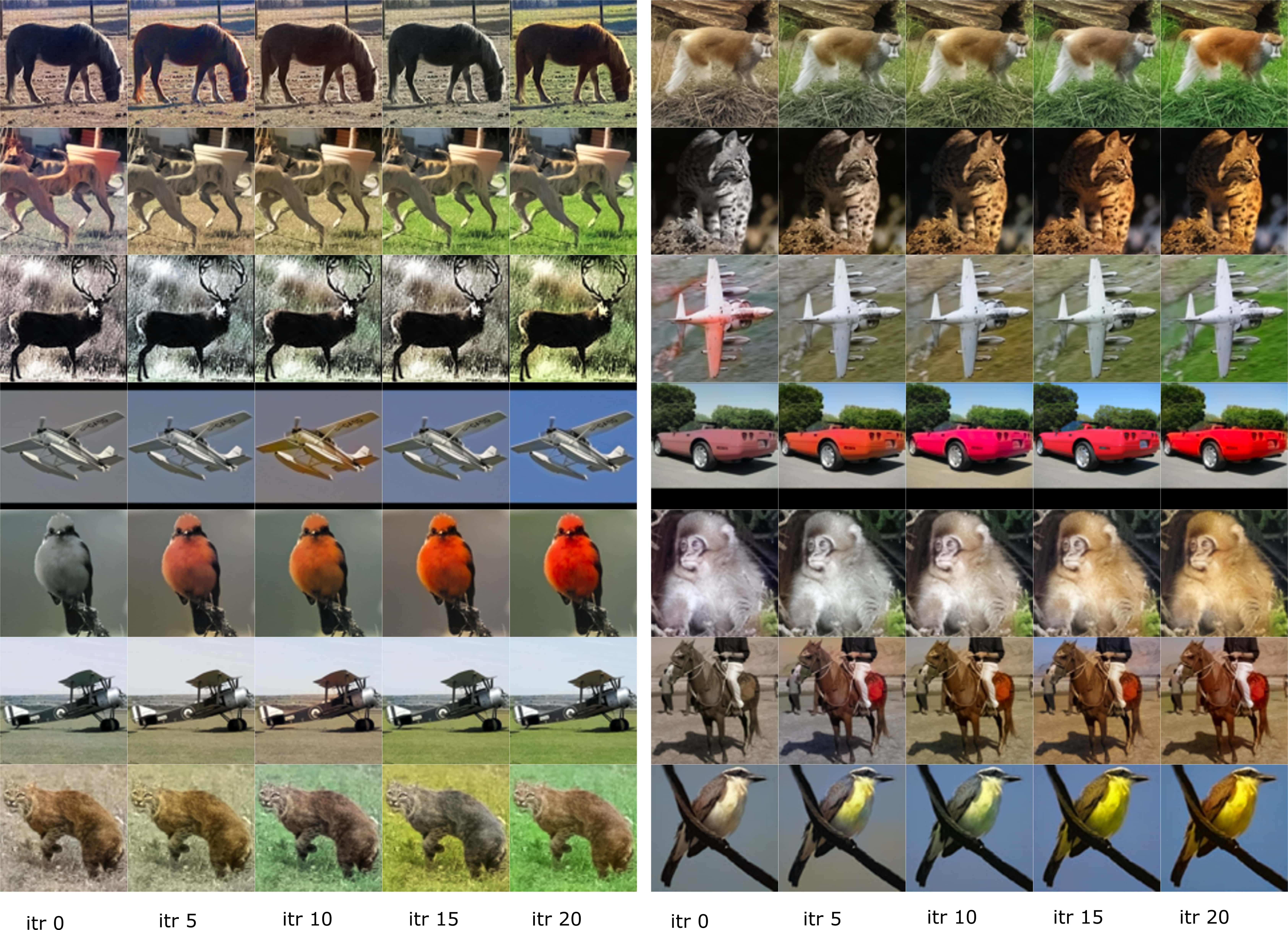} 
\caption{Output quality increases as $z \rightarrow z^*$ at inference.}
\label{fig:inference}
\end{figure}

A key aspect of our method is the optimization of the predictions at inference. Fig.~\ref{fig:mnist_example_finetune} and Fig.~\ref{fig:inference} demonstrate this behaviour on the MNIST image completion and STL colorization tasks, respectively.

\end{document}